%% file: main.tex
\documentclass[10pt,twocolumn,letterpaper]{article}

\usepackage{pgf}
\usepackage{acro}
\usepackage{amsmath}
\usepackage{multirow}
\usepackage[normalem]{ulem}
\usepackage[percent]{overpic}
\usepackage{wacv}              %

\input{preamble}
\definecolor{wacvblue}{rgb}{0.21,0.49,0.74}
\usepackage[pagebackref,breaklinks,colorlinks,allcolors=wacvblue]{hyperref}

\newcommand{\imgpsnr}[3]{
\begin{overpic}[width=#1, c]{#2}
    \put(2,3){\color{white}\fontsize{3}{6}\selectfont PSNR: #3}
\end{overpic}
}

\useunder{\uline}{\ul}{}

\DeclareAcronym{name}{short = ShadowCLR, long = Shadow Consistency Learning for Removal}

\def\wacvPaperID{1128} %
\def\confName{WACV}
\def\confYear{2027}

\title{Consistency as Regularization for Unsupervised Shadow Removal}

\author{Anh-Kiet Duong$^{1,2}$, Petra Gomez-Krämer$^{1}$, Jean-Michel Carozza$^{2}$ \\
$^{1}$ L3i Laboratory, La Rochelle University, 17042 La Rochelle Cedex 1, France \\
$^{2}$ LIENSs Laboratory, La Rochelle University, 17042 La Rochelle Cedex 1, France \\
{\tt\small \{anh.duong,petra.gomez,jean-michel.carozza\}@univ-lr.fr}
}

\begin{document}
\maketitle
\input{sec/0_abstract}    
\input{sec/1_intro}
\input{sec/2_related}
\input{sec/3_method}
\input{sec/4_experiment}

\input{sec/5_conclusion}

{
    \small
    \bibliographystyle{ieeenat_fullname}
    \bibliography{main}
}

\input{sec/X_supp}

\end{document}

%% file: sec/0_abstract.tex
\begin{abstract}
Shadow removal is an important preprocessing step for many vision tasks, yet existing supervised methods require paired shadow and shadow-free images, while unsupervised approaches often still rely on shadow masks or shadow-free references. We propose \acs{name}, an unsupervised framework that learns shadow removal directly from shadow images. Our key observation is that shadows vary across observations while the underlying scene content remains largely consistent. We therefore use consistency across shadow observations as regularization, encouraging the model to recover scene-consistent appearance while suppressing shadow-specific variations. Global and local consistency further enable us to explore visually related images, learn from imperfectly aligned observations, and focus the representation on shared scene information. Experiments on multiple benchmarks show that \acs{name} achieves competitive and often superior performance over state-of-the-art unsupervised methods, demonstrating that consistency can provide regularization for shadow removal without shadow masks or shadow-free images.
\end{abstract}

%% file: sec/1_intro.tex
\section{Introduction}
\label{sec:intro}

Shadows are common in visual data and can degrade downstream tasks such as change detection, surveillance, 3D reconstruction, and scene understanding~\cite{storey2019normalizing,dong2024review,jung2009efficient,sanin2010improved,bouguet19993d}. By disrupting illumination, texture, and color statistics, they may introduce false detections and reconstruction errors. Shadow removal is therefore an important preprocessing step in many real-world applications, yet remains difficult because of the complex interaction between illumination, scene geometry, and surface reflectance~\cite{hu2026unveiling}.

\begin{figure}[t]
    \centering
    \includegraphics[width=0.89\linewidth]{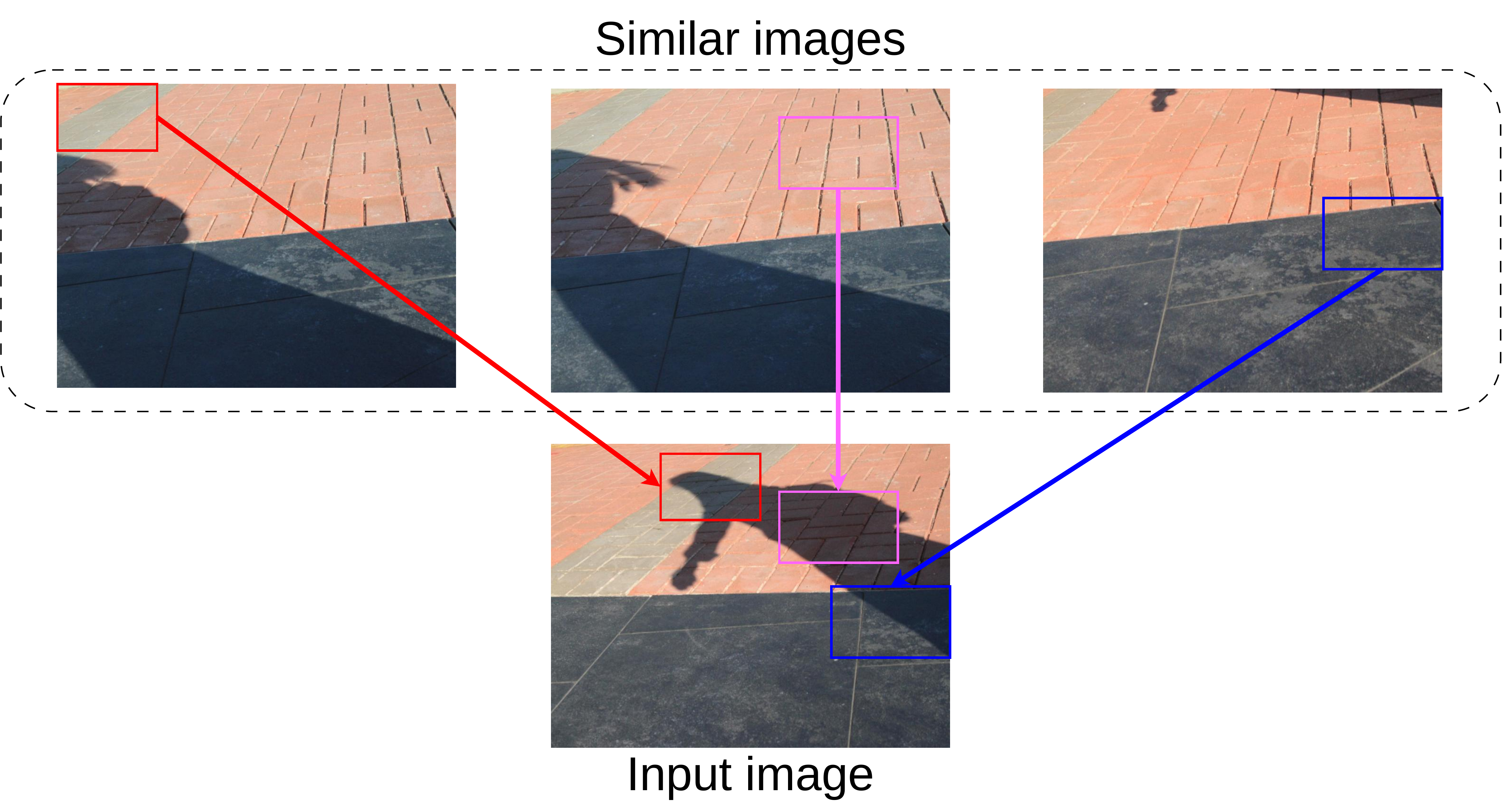}
    \vspace{-1.4ex}
    \caption{Illustration of our core idea. Shadow patterns vary across observations, but scene content remains consistent. We use this consistency to learn shadow removal.}
    \label{fig:motiv}
    \vspace{-5ex}
\end{figure}

Recent deep learning approaches~\cite{chen2021canet, le2021physics, yucel2023lra, liu2023shadow, liu2024recasting, xu2025omnisr, wang2025softshadow, hu2025shadowhack} achieve strong shadow removal performance by training on paired shadow and shadow-free images. However, constructing such paired data requires substantial collection effort and careful alignment, making it difficult to scale to unconstrained scenarios. Unsupervised methods~\cite{hu2019mask, le2020shadow, liu2021shadow, liu2021from, jin2021dc, guo2023boundary} reduce this requirement, but typically still depend on shadow masks or unpaired shadow-free images, which remain difficult to obtain at scale.

In practice, the same scene may appear under different shadow and illumination conditions while preserving its underlying content. Motivated by this observation, we introduce \acf{name}, an unsupervised shadow removal framework that exploits consistency across shadow observations as a source of supervision. As illustrated in \cref{fig:motiv}, the model learns to retain scene-consistent information while suppressing shadow-dependent variations. Visually related images are automatically discovered when available, while a shadow generator ensures that the framework remains applicable even when such related images are limited or absent. The main contributions are:
\begin{itemize}
    \item We propose unsupervised shadow removal that requires neither target shadow masks nor shadow-free images and learns via consistency of shadow observations.
    \item Our framework can also perform unsupervised shadow segmentation as a natural byproduct of shadow removal.
    \item Our framework achieves competitive, and in some cases superior results compared with current state-of-the-art unsupervised methods.
\end{itemize}

The rest of this paper is organized as follows. \Cref{sec:related} reviews related works. \Cref{sec:method} describes the proposed method. \Cref{sec:experiments} reports experimental results and analysis. Finally, \Cref{sec:conclusion} concludes the paper.

%% file: sec/2_related.tex
\section{Related work}
\label{sec:related}

Deep shadow removal methods are generally divided into supervised and unsupervised approaches according to whether paired shadow/shadow-free supervision is required. Other settings, including semi-supervised learning and methods specialized for documents or portraits, have also been studied~\cite{guo2024single}; here, we focus on general natural-image shadow removal.

\subsection{Supervised deep shadow removal}

Early deep shadow removal approaches mainly adopted supervised learning with paired shadow and shadow-free images. DeShadowNet~\cite{qu2017deshadownet} introduced a multi-context embedding architecture, followed by ST-CGAN~\cite{wang2018stacked}, which jointly addressed shadow detection and removal using adversarial learning. Later methods improved spatial reasoning and illumination estimation, including DSC~\cite{hu2018direction}, RIS-GAN~\cite{zhang2020ris}, and DHAN~\cite{cun2020towards}, with designs based on spatial context, residual learning, and hierarchical feature aggregation. Subsequent works incorporated physical priors~\cite{le2019shadow,fu2021auto,liu2023decoupled,zhu2022efficient}, style guidance~\cite{wan2022style}, and invertible or transformer-based architectures~\cite{zhu2022bijective,guo2023shadowformer,xiao2024homoformer}.

More recently, diffusion models have been increasingly explored for shadow removal. ShadowDiffusion~\cite{guo2023shadowdiffusion}, DeS3~\cite{jin2024des3}, and latent diffusion approaches~\cite{mei2024latent} exploit generative priors to improve illumination restoration. More recent methods such as Diff-Shadow~\cite{luo2025diff} and Detail-Preserving Latent Diffusion~\cite{xu2025detail} further enhance reconstruction quality in both shadow and non-shadow regions. Despite their strong performance, supervised methods still depend on costly paired datasets, which limits their use in less controlled environments.

\subsection{Unsupervised deep shadow removal}

To alleviate the dependence on paired training data, unsupervised methods learn from unpaired or weakly supervised samples. Existing methods can mainly be grouped into two categories. The first relies on unpaired shadow and shadow-free images. Mask-ShadowGAN~\cite{hu2019mask}, LG-ShadowNet~\cite{liu2021shadow}, DC-ShadowNet~\cite{jin2021dc}, S3R-Net~\cite{kubiak2024s3r}, and FASR-Net~\cite{lin2025fasr} exploit shadow-free images through adversarial or frequency-domain learning. The second avoids shadow-free references but still requires shadow masks. Le \etal~\cite{le2020shadow} and G2R~\cite{liu2021from} use mask-guided consistency, while Guo \etal~\cite{guo2023boundary} use masks to guide a diffusion process.

Obtaining reliable shadow-free references or accurate shadow masks nevertheless remains difficult in practice. Existing approaches often use ground-truth masks provided by datasets or masks generated by supervised shadow detectors~\cite{zhu2018bidirectional}, which may limit their applicability.

\subsection{Unsupervised deep shadow detection}

Shadow detection aims to localize shadow regions and has been studied alongside shadow removal and generation~\cite{hu2026unveiling,guo2024single}. Early pre-deep-learning methods explored region-level optimization~\cite{vicente2017leave}. Other approaches ASD~\cite{freitas2017automatic}, HSV-Shadow~\cite{kar2015moving}, and Zhang~\etal~\cite{zhang2024deep} exploit handcrafted shadow cues using traditional-based strategies to localize shadow regions. Despite these efforts, deep unsupervised shadow detection has received relatively limited attention. Meanwhile, several unsupervised shadow removal methods still rely on accurate shadow masks obtained from manual annotations or supervised shadow detectors~\cite{le2020shadow,liu2021from}. This dependency limits their applicability where shadow annotations are unavailable. Therefore, developing unsupervised shadow detection or shadow removal methods that do not rely on shadow masks is a crucial step.

%% file: sec/3_method.tex
\section{Proposed method}
\label{sec:method}
\begin{figure*}[ht]
    \centering
    \includegraphics[width=\textwidth]{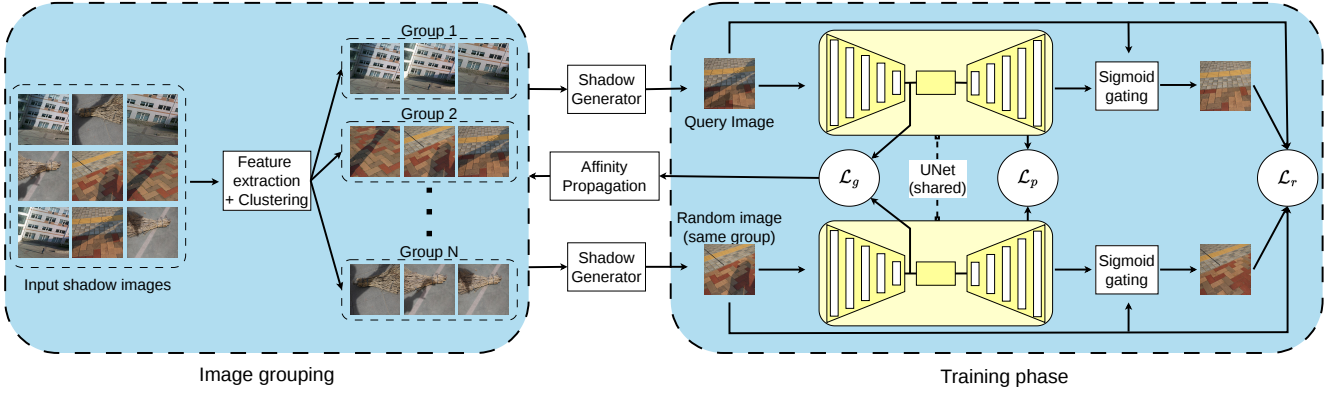}
    \vspace{-3ex}
    \caption{Overview of the proposed framework. Images are first grouped by self-refining image grouping (\cref{sec:group}). Within each group, the network learns to remove shadows by pairwise reconstruction (\cref{sec:pairwise}) and patch-wise correspondence loss (\cref{sec:patchwise}), optimized under the overall objective in \cref{sec:method_overall}.}
    \label{fig:framework}
    \vspace{-3ex}
\end{figure*}

As introduced in \cref{sec:intro}, our key idea is to leverage consistency among different shadow observations for unsupervised shadow removal. In \cref{fig:framework}, we provide an overview of the framework, and this section details each component.

\subsection{Pairwise image loss}
\label{sec:pairwise}
To reconstruct an image, we employ a UNet architecture $U$. Although diffusion-based models achieve strong performance~\cite{guo2023boundary,luo2025diff}, we choose UNet for simple single-pass inference; an iterative refinement illustrating diffusion-style is provided in the supplementary material. Assuming image groups are available, we sample two batches $I=\left\{ x_i \right\}_{i=1}^{B}$ and $\hat{I}=\left\{ \hat{x}_i \right\}_{i=1}^{B}$ such that $x_i$ and $\hat{x}_i$ are in the same group, each group appears once in $I$, and use $\hat{x}_i$ as the pseudo-target for reconstructing $x_i$. Since shadow regions may differ between $x_i$ and $\hat{x}_i$, training across multiple pairs encourages $U(x_i)$ to approach a group-consistent appearance.

However, a group-consistent appearance may correspond to either removing shadows or introducing similar shadows across images. As shadowed regions are typically darker, we introduce a task-specific module that enforces a one-way transformation toward shadow removal:
\begin{equation*}
\begin{aligned}
    g &= \text{sigmoid}\left(s \cdot \left(U(x_i) - x_i \right) \right), \\
    U'(x_i) &= g \cdot U(x_i) + (1 - g) \cdot x_i,
\end{aligned}
\end{equation*}
where $s$ controls the adjustment strength. This formulation blends $U(x_i)$ with $x_i$, ensuring the output mainly brightens shadowed regions while allowing small decreases for smooth transitions if needed. With this sigmoid-based gating and an objective function $\| U'(x_i) - \hat{x}_i \|_1$, the model $U'$ learns to remove shadows that are not consistently present across different observations of the same group.

In addition, some regions may remain shadowed across all observations within a group, making the pseudo-target term alone insufficient. We therefore introduce a reconstruction consistency term $|U'(x_i)-U'(\hat{x}_i)|_1$. Once the model produces a brighter correction for such a region in one observation, this term encourages the paired reconstruction toward the same corrected appearance. Combined with the one-way brightness bias imposed by the sigmoid gating, these corrections can be progressively reinforced across different pairs. Additionally, relying only on this consistency term may lead to trivial constant outputs. Moreover, images within the same group are not necessarily pixel-aligned due to camera shifts, illumination variations, or moving objects as illustrated in \cref{fig:different}. We therefore include a self-reconstruction term $|U'(x_i)-x_i|_1$ to preserve the input-specific content and appearance.

However, this introduces a data assumption that similar images must be available during training. To address this, we use a shadow generator $\mathcal{S}$. For simplicity, we adopt \texttt{RandomShadow}~\cite{buslaev2020albumentations}, which generates shadows by randomly selecting polygonal regions in the image and reducing their brightness. This alleviates the data assumption and allows our method to be trained even when similar images are scarce or absent ($x_i = \hat{x}_i$). Combining all terms, the final reconstruction loss becomes:
\begin{equation*}
\resizebox{\linewidth}{!}{$
\mathcal{L}_{\text{r}} = \| U'(S(x_i)) - \hat{x}_i \|_1 + \| U'(S(x_i)) - x_i \|_1 + \| U'(S(x_i)) - U'(S(\hat{x}_i)) \|_1.
$}
\end{equation*}
As all components of $\mathcal{L}_{\text{r}}$ compare images of identical resolution, their weights are set equally for simplicity. We also examine the impact of different parameters in \cref{sec:ablation}. Together, these terms promote consistent shadow removal across each group, reduce sensitivity to imperfect pixel-level alignment, and enable training even without visually related images.

\input{figures/pairwise}

\subsection{Self-refining image grouping}
\label{sec:group}
As \cref{sec:pairwise} assumes access to image groups, we next consider how such groups can be obtained. While these relationships may be available from camera metadata, we consider the more general setting where neither group labels nor the number of groups is known. We initialize pseudo labels using Affinity Propagation~\cite{frey2007clustering}, which does not require the number of groups and using the default parameters~\cite{pedregosa2011scikit} on precomputed pairwise $-\ell_1$ image distances.

During the forward pass of the paired batches $I$ and $\hat{I}$ defined in \cref{sec:pairwise}, the shadow generator $\mathcal{S}$ is first applied to each image before feeding it to the shared UNet. We collect the last $\ell_2$-normalized global UNet encoder features $z_i$ and $\hat{z}_i$ corresponding to $\mathcal{S}(x_i)$ and $\mathcal{S}(\hat{x}_i)$, respectively. Since $x_i$ and $\hat{x}_i$ belong to the same pseudo group, $(z_i,\hat{z}_i)$ forms a positive pair, while the remaining $2(B-1)$ features in the two batches are treated as negatives. We then have a global contrastive loss:
\begin{equation*}
\resizebox{\linewidth}{!}{$
\mathcal{L}_{\text{g}} = -w_i \log\frac{e^{z_i^\top \hat{z}_i / \tau}}{e^{z_i^\top \hat{z}_i / \tau} + \sum_{j \neq i} e^{z_i^\top z_j / \tau} + \sum_{j \neq i} e^{z_i^\top \hat{z}_j / \tau}}, 
w_i = \operatorname{sg}\!\left( 1 + z_i^\top \hat{z}_i \right).
$}
\end{equation*}
where the temperature hyper-parameter $\tau$ is set to 0.3 following the standard practice~\cite{duong2026image, hu2024asymmetric}. The stop-gradient $\operatorname{sg}(\cdot)$ term acts as a confidence weight: samples already close to each other receive a stronger attraction, while uncertain assignments are updated more conservatively (detailed in the supplementary material). After each epoch, we re-extract the last $\ell_2$-normalized global UNet encoder features for all training images and compute their pairwise cosine similarities. Affinity Propagation is then reapplied to these similarities to obtain updated pseudo labels for the next epoch, allowing the grouping and feature representations to progressively refine each other. As illustrated in \cref{fig:group}, the resulting pseudo groups contain visually related images despite variations in shadows and illumination.

This self-refining scheme directly uses the UNet encoder, avoiding any auxiliary model or heavy pretrained backbone. By minimizing the distance between positive pairs while separating negative pairs, the contrastive objective encourages the encoder to capture scene-level information that is consistent across different observations while suppressing shadow-specific variations. This provides cleaner and more consistent representations for the decoder, improving the quality of the restored images. Comparing images in the learned feature space also reduces sensitivity to the pixel-level variations discussed in \cref{sec:pairwise}.

\subsection{Patch-wise correspondence loss}
\label{sec:patchwise}
As discussed in \cref{sec:pairwise}, images within the same group may not be perfectly pixel-aligned due to camera shifts, illumination variations, or moving objects as illustrated in \cref{fig:different}. Although the global contrastive learning in \cref{sec:group} reduces the influence of these variations at the scene level, global representations alone may not sufficiently constrain local structures. We therefore introduce a patch-wise correspondence mechanism, illustrated in \cref{fig:patchwise}.

Given input images $x_i, \hat{x}_i$, we extract multi-scale feature maps $v^{l}_i$ and $\hat{v}^{l}_i \in \mathbb{R}^{C_l \times H_l \times W_l}$ from the UNet encoder, where $C_l$ denotes the feature dimension and $H_l\times W_l$ its spatial resolution of the $l$-th layer among $L$ UNet layers. Each spatial location $(m,n)$ is treated as a patch-wise feature $v_i^l(m,n)\in\mathbb{R}^{C_l}$ and is $\ell_2$-normalized along the feature dimension. We also obtain a global representation $z_i^l$ for each feature map by spatial mean pooling followed by $\ell_2$ normalization. The correspondence loss is defined as:
\begin{equation*}
\resizebox{\linewidth}{!}{$
\begin{aligned}
    & \mathcal{L}_{\text{p}} = \frac{1}{\sum_l H_lW_l}\sum_{(l,m,n)} -w_i(m, n) \log\frac{e^{v_i^{l}(m, n)^\top \hat{v}_i^{l}(m', n') / \tau}}{e^{v_i^{l}(m, n)^\top \hat{v}_i^{l}(m', n') / \tau} + \sum_{j \neq i} e^{v_i^{l}(m, n)^\top z^l_j / \tau} + \sum_{j \neq i} e^{v_i^{l}(m, n)^\top \hat{z}^l_j / \tau}}, \\
    & w_i(m, n) = \operatorname{sg}\!\left( 1 + v_i^{l}(m, n)^\top \hat{v}_i^{l}(m', n') \right), (m', n') = \arg\max_{(m', n')} v_i^{l}(m, n)^\top \hat{v}_i^{l}(m', n').
\end{aligned}
$}
\end{equation*}
Similar to the global contrastive loss $\mathcal{L}_{\mathrm{g}}$ in \cref{sec:group}, the stop-gradient weight assigns stronger supervision to more confident correspondences. Here, the positive pair is formed between corresponding local patches, whereas each anchor patch is contrasted with the mean feature of each image from other groups. Since this mean aggregates information from all patches of the negative image, it allows each patch-wise feature to be compared against the overall image content without directly treating every patch from another group as a negative. This reduces false-negative constraints when similar objects also appear in different groups. Unlike LPIPS~\cite{zhang2018unreasonable} feature matching at identical spatial locations, we search for the most similar patch in the paired image, relaxing the spatial alignment assumption and improving robustness to camera shifts and moving objects.

\subsection{Overall framework}
\label{sec:method_overall}
The pairwise reconstruction in \cref{sec:pairwise} learns shadow removal by enforcing consistency either between images from the same group or between different shadow generations of the same image. Visually related images are automatically discovered by the self-refining grouping in \cref{sec:group}. The global and patch-wise contrastive objectives further encourage the model to preserve scene-consistent information while suppressing shadow variations, and relax the need for pixel-level alignment under camera shifts or moving objects. The overall objective is a combination of three terms:

\begin{equation*}
    \mathcal{L} = \mathcal{L}_{\text{r}} + \lambda_{\text{g}}\mathcal{L}_{\text{g}} + \lambda_{\text{p}} \, \mathcal{L}_{\text{p}},
\end{equation*}
where $\lambda_{\text{g}}, \lambda_{\text{p}}$ balance the grouping and patch-wise correspondence terms, respectively. Overall the regularization is derived solely from consistency constraints together with the prior that shadow regions are darker, encoded by the sigmoid gating and shadow generation, without requiring shadow masks or shadow-free images.

%% file: figures/pairwise.tex
\begin{figure}[ht]
\centering

\subfloat[Examples of image groups discovered.\label{fig:group}]{
    \begin{minipage}[b]{0.42\linewidth}
        \centering
        \includegraphics[width=0.31\linewidth]{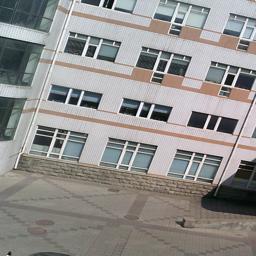}
        \includegraphics[width=0.31\linewidth]{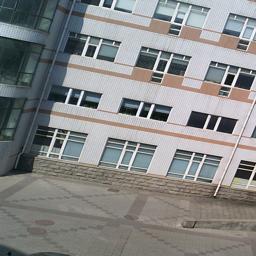}
        \includegraphics[width=0.31\linewidth]{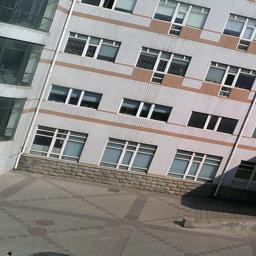}\\[0.3em]
        \includegraphics[width=0.31\linewidth]{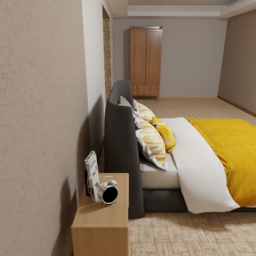}
        \includegraphics[width=0.31\linewidth]{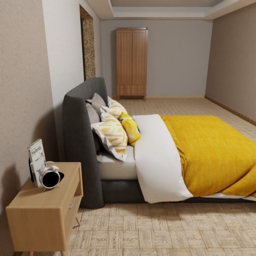}
        \includegraphics[width=0.31\linewidth]{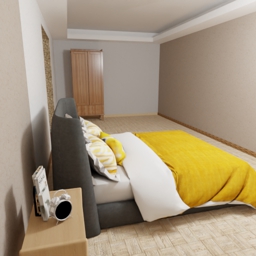}
    \end{minipage}
}
\hspace{0.02\linewidth}
\subfloat[Examples of configuration variations across similar images.\label{fig:different}]{
    \begin{minipage}[b]{0.42\linewidth}
        \centering
        \includegraphics[width=0.31\linewidth]{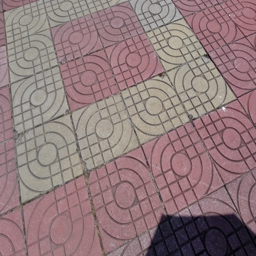}
        \includegraphics[width=0.31\linewidth]{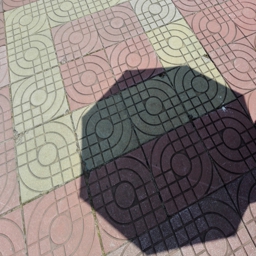}
        \includegraphics[width=0.31\linewidth]{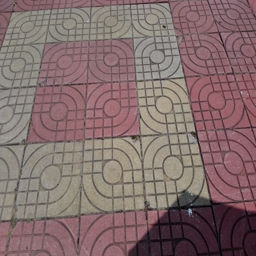}\\[0.3em]
        \includegraphics[width=0.31\linewidth]{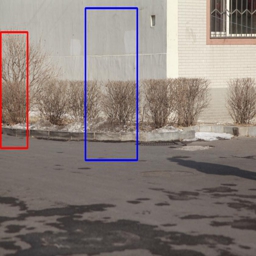}
        \includegraphics[width=0.31\linewidth]{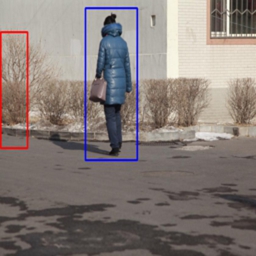}
        \includegraphics[width=0.31\linewidth]{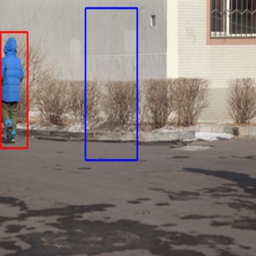}
    \end{minipage}
}

\subfloat[Illustration of patch-wise correspondence loss.\label{fig:patchwise}]{
    \includegraphics[width=0.65\linewidth]{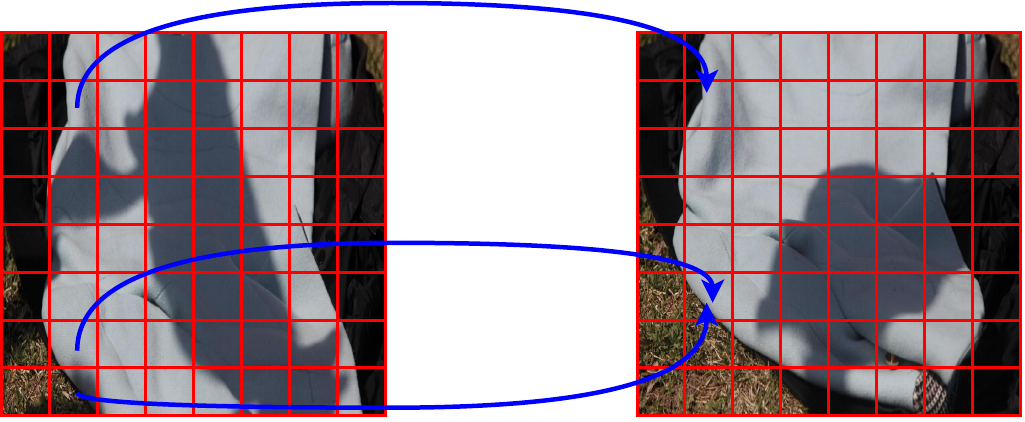}
}
\vspace{-1ex}
\caption{
Visualization of the proposed grouping and correspondence strategy: discovered pseudo groups, variations among similar images, and patch-wise matching for misaligned regions. 
}
\vspace{-2ex}
\label{fig:pairwise}
\end{figure}

%% file: sec/4_experiment.tex
\section{Experiments}
\label{sec:experiments}

In this section, we evaluate the proposed method in \cref{sec:method}, compare it with existing approaches in \cref{sec:sota}, show qualitative results in \cref{sec:qualitative}, and analyze its components through ablation studies in \cref{sec:ablation}.

\subsection{Experimental settings}

\paragraph{Setups.} We implemented our method in PyTorch~\cite{paszke2019pytorch}, running all experiments on a single NVIDIA H100 GPU. Following prior works~\cite{guo2023boundary,jin2021dc,liu2021from}, all images are resized to $256\times256$ for both training and testing. For image restoration, we use a lightweight U-Net trained from scratch, following~\cite{liu2021from, hu2019mask}, to isolate the effect of our framework without relying on pretrained restoration or diffusion models. At inference, the model costs $0.05$~TFLOPs per image. As reported in \cref{tab:aistd,tab:merge,tab:complexity}, our parameters and FLOPs remain comparable to other non-diffusion-based unsupervised methods, whereas diffusion-based methods such as Guo~\etal~\cite{guo2023boundary} require $61$~TFLOPs. The model is trained for $100$ epochs with batch size $8$ using the Adam optimizer~\cite{kingma2014adam}, a learning rate of $1\times10^{-5}$, and hyperparameters $s=128$, $\lambda_g=\lambda_p=1.0$. We also analyze in \cref{sec:ablation,sec:more_results} the effect of $s, \lambda_g$ and $\lambda_p$.

\paragraph{Datasets.} We evaluate our method on AISTD~\cite{le2019shadow}, INS~\cite{xu2025omnisr},  WSRD+~\cite{vasluianu2023wsrd, vasluianu2024ntire}, LRSS~\cite{gryka2015learning}, SRD~\cite{qu2017deshadownet}, and the video shadow removal dataset~\cite{le2020shadow}. AISTD (also referred to as ISTD+) is an adjusted version of the ISTD dataset~\cite{wang2018stacked} with reduced illumination discrepancies in non-shadow areas, which is important since unsupervised methods have no access to shadow-free images. AISTD includes $1,330$ shadow images for training and $540$ for testing. SRD contains $2,680$ shadow images for training and $408$ for testing. INS contains $30,000$ training images, and $2,000$ testing images. The video dataset contains $8$ videos of static scenes with consistent backgrounds. 

\paragraph{Evaluation metrics.} We follow prior works~\cite{guo2023boundary, liu2021shadow} and use MATLAB to compute Root Mean Square Error (RMSE), Peak Signal-to-Noise Ratio (PSNR), and Structural Similarity Index (SSIM)~\cite{wang2004image} for evaluation. RMSE is computed over the three \textit{Lab} channels, while PSNR and SSIM are calculated in RGB to assess visual fidelity and structural consistency. A lower RMSE indicates a smaller reconstruction error, while a higher PSNR and SSIM values correspond to better restoration quality.

\subsection{Comparison with the state-of-the-art}
\label{sec:sota}

In this section, we compare \acs{name} with state-of-the-art methods for shadow removal. The comparison includes a classic method: Guo \etal~\cite{guo2012paired}; supervised methods: DHAN~\cite{cun2020towards}, AEF~\cite{fu2021auto}, BMNet~\cite{zhu2022bijective}, ShadowFormer~\cite{guo2023shadowformer}, ShadowDiffusion~\cite{guo2023shadowdiffusion}, HomoFormer~\cite{xiao2024homoformer}, and Omnisr~\cite{xu2025omnisr}; and unsupervised methods: Mask-ShadowGAN~\cite{hu2019mask}, LG-ShadowNet~\cite{liu2021shadow}, DC-ShadowNet~\cite{jin2021dc}, Le \etal~\cite{le2020shadow}, G2R~\cite{liu2021from}, Guo \etal~\cite{guo2023boundary}, S3R-Net~\cite{kubiak2024s3r}, and FASR-Net~\cite{lin2025fasr}. All results are taken from the original papers while results produced by official implementations are shown in \textit{italic}.

\input{tables/aistd.tex}

In \cref{tab:aistd}, we report the quantitative comparison on the AISTD dataset. Supervised methods achieve strong performance by training on paired shadow and shadow-free images. Unsupervised methods also achieve competitive results, in some cases approaching the performance of supervised ones; however, they typically depend on shadow-free references and/or shadow masks for guidance. On the other hand, \acs{name} achieves competitive results compared with state-of-the-art unsupervised methods. Specifically, our method surpasses early supervised approaches such as DHAN~\cite{cun2020towards} and AEF~\cite{fu2021auto}.

\input{tables/merge}

In \cref{tab:merge}, we report the quantitative comparison on the SRD, LRSS, WSRD+ and INS datasets. Following~\cite{jin2021dc}, the LRSS results are obtained by fine-tuning a model pretrained on SRD, while other results are obtained by training on the respective dataset. Unlike AISTD, which provides ground-truth shadow masks, these datasets do not include shadow mask annotations. As many unsupervised methods rely on shadow mask supervision, we have to use the generated masks from other shadow detection methods to evaluate the performance of these methods. Compared with the other evaluated unsupervised baselines, \acs{name} achieves higher performance across all metrics and datasets. Note that INS is a synthetic dataset representing scenarios where similar images are scarce or absent.

\subsection{Qualitative results}
\label{sec:qualitative}
\input{figures/demo_AISTD}
\input{figures/demo_SRD}

The qualitative results on the AISTD and SRD datasets are shown in \cref{fig:demo_AISTD,fig:demo_SRD}. In both cases, \acs{name} effectively removes shadows while maintaining color consistency and preserving scene details. Compared to prior unsupervised methods such as Mask-ShadowGAN, LG-ShadowNet, and DC-ShadowNet, our method generates results that are visually cleaner and closer to the ground truth. Notably, \acs{name} better restores surface textures and illumination transitions, especially in challenging regions such as partially overlapping with dark objects. These visual outcomes further confirm the quantitative improvements in \cref{tab:aistd,tab:merge}. Note that Guo \etal~\cite{guo2023boundary} did not release code or outputs; therefore, their method is excluded from the qualitative comparison.

\subsection{Grouping quality}
\label{sec:group_quality}

Only AISTD provides scene group information through image filenames. We therefore use these filename-derived groups as ground truth to evaluate grouping quality. Grouping with pairwise $\ell_1$ produces 115 groups with an average of 11.57 images per group ranging from 2 to 111 images and an ARI of $0.588$. Using DINOv3 features gives 114 groups with 11.67 images per group on average ranging from 4 to 27 images and an ARI of $0.700$. After training, our method produces 119 groups with an average of 11.18 images per group ranging from 3 to 27 images, increasing the ARI to $0.918$. The ground truth contains 89 groups with an average of 14.94 images per group, ranging from 1 to 29 images. On average, $98.80\%$ of the shadow pixels have at least one non-shadow observation within their ground truth group, $66.43\%$ of the candidate images provide a non-shadow observation for the anchor's shadow pixels. Nevertheless, recovering the ground-truth grouping is difficult while the presence of singleton groups makes selecting a fixed representative image as a pseudo-target unreliable.

More specifically, with $\ell_1$ and DINOv3 grouping, only $50.55\%$ and $53.40\%$ of the candidate images, respectively, provide a correct non-shadow observation for the anchor's shadow pixels (non-shadow may differ from the ground truth). Selecting higher-ranked candidates does not fully solve this issue: the 70th percentile provides correct observations for only $66.21\%$ and $70.29\%$ of the anchor's shadow pixels, while the 90th percentile reaches $69.86\%$ and $75.44\%$. After self-refinement, these values improve to $63.87\%$ overall and $86.54\%$/$92.99\%$ at the 70th/90th percentiles, respectively, but still does not provide a reliable valid pseudo-target. This becomes particularly problematic when a group contains only one image or when images from different scenes are incorrectly grouped together. These observations motivate learning from pairwise consistency rather than relying on a median or high-percentile group representative as a fixed pseudo-target.

\subsection{Ablation studies}
\label{sec:ablation}

Following~\cite{guo2023boundary,liu2021from}, we conduct ablations on AISTD, which provides scene-level information and the most reported unsupervised shadow removal results. ARI gain is the difference before and after training; all variants use $-\ell_1$ initialization except DINOv3-based variants, which use pairwise DINOv3 feature similarity. As shown in \Cref{tab:ablation}, each component contributes to performance. Replacing the patch-wise loss with LPIPS or fine-tuning DINOv3~\cite{simeoni2025dinov3} with the global contrastive loss instead of directly optimizing the UNet reduces performance, showing that contrastive learning is more effective when jointly improving the restoration model. Removing the shadow generator has little effect when similar images are available. For gating, $s=64$ better preserves non-shadow regions, while $s=256$ better restores shadows. Using only $\mathcal{L}_r$ with the shadow generator ($\lambda_g=\lambda_p=0$ without grouping) performs worse than DINOv3 grouping and one-image-per-group, especially for shadow detection and restoration. This suggests that feature-level consistency helps learn more general appearance representations even across images from different GT groups. One image per GT group still performs comparably to Mask-ShadowGAN despite fewer training images. The iterative diffusion-style variant and further ablations are given in the supplementary. Using a constant confidence weight also reduces ARI gain and restoration quality, especially PSNR, indicating less accurate RGB color recovery.

\input{tables/ablation}

\subsection{Unsupervised shadow segmentation}

Shadow removal inherently provides cues for locating shadow regions, as it involves estimating where illumination should be compensated. We extend \acs{name} to perform unsupervised shadow segmentation by directly using the sigmoid-based gating map $g$ introduced in \cref{sec:pairwise} as a proxy soft shadow matte. The gating map is converted to grayscale using standard luminance weights. Following~\cite{chen2020multi,cun2020towards}, we train the model on the combined SRD and AISTD datasets and evaluate the performance on AISTD. We further train and test on the SBU dataset~\cite{vicente2016large}, which contains $638$ test images. The qualitative and quantitative results are shown in \cref{fig:segmentation,tab:segmentation}. Compared with AISTD, whose shadows mainly appear on relatively planar surfaces, SBU contains more complex shadows across diverse surfaces, leading to lower performance; the same trend is also observed across both supervised and unsupervised methods.

\input{figures/segmentation}

\input{tables/segmentation}

\subsection{Video shadow removal}

We further evaluate \acs{name} on the public video shadow removal dataset~\cite{le2020shadow}, which contains static scenes with consistent backgrounds. Each sequence provides a pseudo shadow-free reference frame $V_{\text{max}}$, obtained by taking the maximum pixel intensity across all frames. Following prior works~\cite{le2020shadow, guo2023boundary, liu2021from, liu2021shadow}, we directly apply the model trained on AISTD to these sequences and additionally report results for a variant fine-tuned for one epoch. 

\input{tables/video}

\input{figures/video}

The evaluation focuses on moving-shadow regions, defined as pixels that alternate between shadowed and non-shadowed states throughout the video. The moving-shadow mask is generated according to the official protocol in~\cite{le2020shadow} using a threshold of $80$. All quantitative metrics are computed within the moving-shadow region. Following prior works, RMSE$^{*}$ is additionally reported with a threshold of $40$, while the remaining metrics use the standard threshold of $80$. As shown in \cref{tab:video} and \cref{fig:video}, \acs{name} achieves superior performance among unsupervised methods.

\subsection{Limitation and discussion}

A main limitation of our method lies in preserving very fine details and small shadow structures. Although feature-level consistency and patch-wise correspondence alleviate misalignment caused by camera shifts or moving objects, subtle changes such as grass moving between observations may not be sufficiently represented in the learned features, leading to slight detail degradation as shown in \cref{fig:failure}. Similar discrepancies may also occur in supervised learning when pairs are captured at different times. The failure cases further show that smaller $s$ better preserves complex scene details, whereas larger $s$ is more effective for complex shadows. More advanced refinement strategies could potentially alleviate this issue; for example, Hu \etal~\cite{hu2025shadow} fine-tune SAM~\cite{kirillov2023segment} to obtain shadow-invariant segmentation, which could be used to enforce texture consistency across materials and boundaries. The shadow generator could also be improved to produce more diverse and realistic shadow variations. Such refinement could be incorporated in future work. In this work, we intentionally retain a lightweight formulation to emphasize the core idea.

\input{figures/failure}

%% file: tables/aistd.tex
\begin{table*}[ht]
\centering
\caption{Quantitative comparison results of \acs{name} with the state-of-the-art methods on the AISTD dataset~\cite{le2019shadow, wang2018stacked}.
The best and second-best results within each setting are highlighted in \textbf{bold} and
{\ul underlined}, respectively.}
\label{tab:aistd}
\resizebox{0.99\textwidth}{!}{
\begin{tabular}{l|cc|cc|ccc|ccc|ccc}
\hline
\multirow{2}{*}{Method} & \multicolumn{2}{c|}{Setting} & \multicolumn{2}{c|}{Complexity} & \multicolumn{3}{c|}{Shadow Region (S)} & \multicolumn{3}{c|}{Non-Shadow Region (N)} & \multicolumn{3}{c}{All Image (A)} \\
 & Free & Mask & TFLOPS & Params (M) & PSNR $\uparrow$ & SSIM $\uparrow$ & RMSE $\downarrow$ & PSNR $\uparrow$ & SSIM $\uparrow$ & RMSE $\downarrow$ & PSNR $\uparrow$ & SSIM $\uparrow$ & RMSE $\downarrow$ \\ \hline
Guo \etal~\cite{guo2012paired} & \multicolumn{2}{c|}{Classic} & - & - & 26.89 & 0.960 & 20.1 & 35.48 & 0.975 & 3.1 & 25.51 & 0.924 & 6.1 \\ \hline
DHAN~\cite{cun2020towards} & \multicolumn{2}{c|}{\multirow{6}{*}{Supervised}} & 0.26 & 21.8 & 32.92 & 0.988 & 9.6 & 27.15 & 0.971 & 7.4 & 25.66 & 0.956 & 7.8 \\
AEF~\cite{fu2021auto} & \multicolumn{2}{c|}{} & - & 196.76 & 36.04 & 0.978 & 6.7 & 31.16 & 0.892 & 3.8 & 29.45 & 0.861 & 4.2 \\
BMNet~\cite{zhu2022bijective} & \multicolumn{2}{c|}{} & - & 0.58 & 37.87 & 0.991 & 5.6 & 37.51 & {\ul 0.985} & 2.5 & 33.98 & {\ul 0.972} & 3.0 \\
ShadowFormer~\cite{guo2023shadowformer} & \multicolumn{2}{c|}{} & - & 11.37 & 39.48 & {\ul 0.992} & 5.2 & {\ul 38.82} & 0.983 & 2.3 & {\ul 35.46} & 0.971 & 2.8 \\
ShadowDiffusion~\cite{guo2023shadowdiffusion} & \multicolumn{2}{c|}{} & - & 55.52 & \textbf{39.69} & {\ul 0.992} & {\ul 5.0} & \textbf{38.89} & \textbf{0.987} & {\ul 2.3} & \textbf{35.67} & \textbf{0.975} & {\ul 2.7} \\
HomoFormer~\cite{xiao2024homoformer} & \multicolumn{2}{c|}{} & - & 17.81 & {\ul 39.49} & \textbf{0.993} & \textbf{4.7} & 38.75 & 0.984 & \textbf{2.2} & 35.35 & \textbf{0.975} & \textbf{2.6} \\ \hline
Mask-ShadowGAN~\cite{hu2019mask} & \checkmark &  & 0.05 & 22.8 & 32.19 & 0.984 & 10.8 & 33.44 & 0.974 & 3.8 & 28.81 & 0.946 & 4.8 \\
LG-ShadowNet~\cite{liu2021shadow} & \checkmark &  & 0.03 & 5.7 & 32.44 & 0.982 & 9.9 & 33.68 & 0.971 & 3.4 & 29.20 & 0.945 & 4.4 \\
DC-ShadowNet~\cite{jin2021dc} & \checkmark &  & 0.05 & 10.6 & 31.06 & 0.976 & 12.2 & 27.03 & 0.961 & 6.8 & 25.03 & 0.926 & 7.8 \\
Le \etal~\cite{le2020shadow} &  & \checkmark & - & - & 33.09 & 0.983 & 10.4 & 35.26 & 0.977 & 2.9 & 30.12 & 0.950 & 4.0 \\
G2R~\cite{liu2021from} &  & \checkmark & 0.11 & 22.8 & 33.58 & 0.979 & 8.9 & 35.52 & 0.976 & 2.9 & 30.52 & 0.944 & 3.9 \\
Guo \etal~\cite{guo2023boundary} &  & \checkmark & 61 & 113.7 & {\ul 35.71} & {\ul 0.986} & {\ul 7.6} & {\ul 36.39} & {\ul 0.981} & {\ul 2.7} & {\ul 32.11} & {\ul 0.959} & {\ul 3.5} \\
S3R-Net~\cite{kubiak2024s3r} & \checkmark & \multicolumn{1}{l|}{} & - & - & - & - & 12.2 & - & - & 6.4 & - & - & 7.1 \\
FASR-Net~\cite{lin2025fasr} & \checkmark & \multicolumn{1}{l|}{} & - & - & 31.89 & 0.982 & 8.61 & 34.57 & 0.978 & 2.8 & 27.58 & 0.934 & 3.8 \\
\acs{name} (Ours) &  &  & 0.05 & 11.4 & \textbf{37.47} & \textbf{0.989} & \textbf{5.9} & \textbf{36.96} & \textbf{0.984} & \textbf{2.6} & \textbf{33.35} & \textbf{0.964} & \textbf{3.1} \\ \hline
\end{tabular}
}
\end{table*}

%% file: tables/merge.tex
\begin{table*}[ht]
\centering
\caption{Quantitative comparison results of \acs{name} with the state-of-the-art methods on the SRD~\cite{qu2017deshadownet}, INS~\cite{xu2025omnisr}, LRSS~\cite{gryka2015learning}, and WSRD+~\cite{vasluianu2024ntire} datasets. The best and second-best results within each setting are highlighted in \textbf{bold} and {\ul underlined}, respectively.}
\label{tab:merge}
\resizebox{0.99\textwidth}{!}{
\begin{tabular}{l|cc|cc|cc|cc|cc|cc}
\hline
\multirow{2}{*}{Method} & \multicolumn{2}{c|}{Setting} & \multicolumn{2}{c|}{Complexity} & \multicolumn{2}{c|}{INS} & \multicolumn{2}{c|}{WSRD+} & \multicolumn{2}{c|}{LRSS} & \multicolumn{2}{c}{SRD} \\
 & Free & Mask & TFLOPS & Params (M) & PSNR $\uparrow$ & SSIM $\uparrow$ & PSNR $\uparrow$ & SSIM $\uparrow$ & PSNR $\uparrow$ & SSIM $\uparrow$ & PSNR $\uparrow$ & SSIM $\uparrow$ \\ \hline
DHAN~\cite{cun2020towards} & \multicolumn{2}{c|}{\multirow{7}{*}{Supervised}} & 0.26 & 21.8 & 27.84 & 0.963 & - & - & 25.57 & - & 30.74 & 0.958 \\
AEF~\cite{fu2021auto} & \multicolumn{2}{c|}{} & - & 196.76 & 27.91 & 0.957 & 21.66 & 0.752 & - & - & 28.40 & 0.893 \\
BMNet~\cite{zhu2022bijective} & \multicolumn{2}{c|}{} & - & 0.58 & 27.90 & 0.958 & 24.75 & 0.816 & - & - & 31.69 & 0.956 \\
ShadowFormer~\cite{guo2023shadowformer} & \multicolumn{2}{c|}{} & - & 11.37 & 28.62 & 0.963 & {\ul 25.44} & {\ul 0.820} & 31.91 & 0.876 & 32.90 & 0.958 \\
ShadowDiffusion~\cite{guo2023shadowdiffusion} & \multicolumn{2}{c|}{} & - & 55.52 & {\ul 29.12} & {\ul 0.966} & \textbf{-} & \textbf{-} & - & - & {\ul 34.73} & {\ul 0.970} \\
HomoFormer~\cite{xiao2024homoformer} & \multicolumn{2}{c|}{} & - & 17.81 & 28.98 & 0.965 & - & - & - & - & \textbf{35.37} & \textbf{0.972} \\
Omnisr~\cite{xu2025omnisr} & \multicolumn{2}{c|}{} & - & - & \textbf{30.38} & \textbf{0.973} & \textbf{26.07} & \textbf{0.835} & \textbf{33.62} & \textbf{0.892} & 32.87 & 0.969 \\ \hline
Mask-ShadowGAN~\cite{hu2019mask} & \checkmark &  & 0.05 & 22.8 & \textit{24.91} & \textit{0.928} & \textit{20.31} & \textit{0.741} & \textit{30.83} & \textit{0.852} & \textit{30.49} & \textit{0.947} \\
LG-ShadowNet~\cite{liu2021shadow} & \checkmark &  & 0.03 & 5.7 & \textit{25.79} & \textit{0.933} & \textit{21.94} & \textit{0.752} & {\ul \textit{31.28}} & \textit{0.857} & \textit{30.85} & \textit{0.946} \\
DC-ShadowNet~\cite{jin2021dc} & \checkmark &  & 0.05 & 10.6 & \textit{26.48} & \textit{0.941} & \textit{22.62} & \textit{0.761} & 31.01 & {\ul 0.861} & {\ul 31.53} & {\ul 0.955} \\
G2R~\cite{liu2021from} + DHAN~\cite{cun2020towards} &  & \checkmark & 0.11+0.26 & 22.8+21.8 & \textit{24.21} & \textit{0.932} & \textit{18.04} & \textit{0.726} & \textit{29.23} & \textit{0.845} & \textit{29.12} & \textit{0.945} \\
G2R~\cite{liu2021from} + SDDNet~\cite{cong2023sddnet} &  & \checkmark & 0.11+0.01 & 22.8+46.5 & \textit{25.06} & \textit{0.925} & \textit{19.10} & \textit{0.738} & \textit{28.47} & \textit{0.839} & \textit{29.38} & \textit{0.948} \\
\acs{name} (Ours) &  &  & 0.05 & 11.4 & \textbf{28.07} & \textbf{0.960} & \textbf{23.81} & \textbf{0.793} & \textbf{31.54} & \textbf{0.868} & \textbf{31.93} & \textbf{0.958} \\ \hline
\end{tabular}
}
\vspace{-2.5ex}
\end{table*}

%% file: figures/demo_AISTD.tex
\begin{figure*}[ht]
\centering
\newcommand{\ncols}{10}
\setlength{\tabcolsep}{0.1em}

\newlength{\imgwidthAISTD}
\setlength{\imgwidthAISTD}{\dimexpr(0.98\linewidth - \ncols\tabcolsep)/\ncols\relax}

\begin{tabular}{*{\ncols}{c}}
\centering
\includegraphics[width=\imgwidthAISTD]{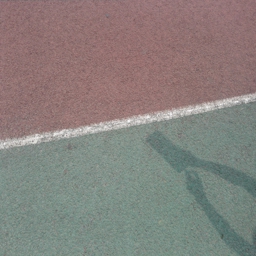} &
\includegraphics[width=\imgwidthAISTD]{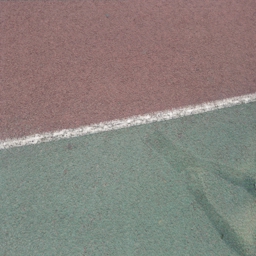} &
\includegraphics[width=\imgwidthAISTD]{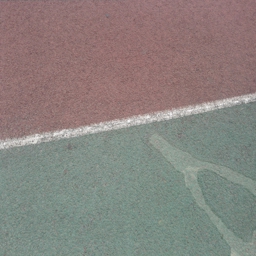} &
\includegraphics[width=\imgwidthAISTD]{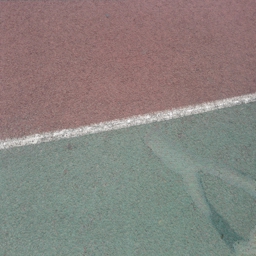} &
\includegraphics[width=\imgwidthAISTD]{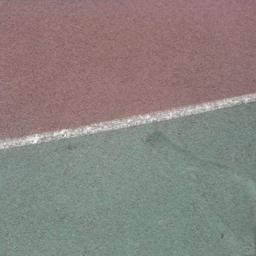} &
\includegraphics[width=\imgwidthAISTD]{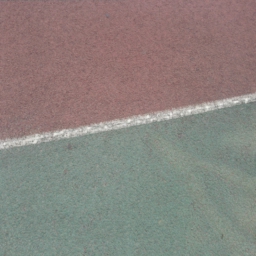} &
\includegraphics[width=\imgwidthAISTD]{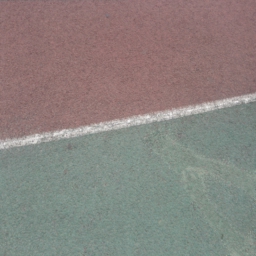} &
\includegraphics[width=\imgwidthAISTD]{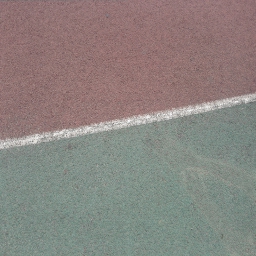} &
\includegraphics[width=\imgwidthAISTD]{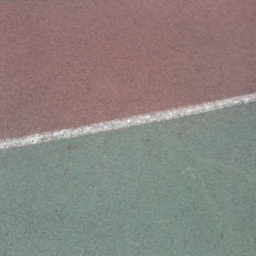} &
\includegraphics[width=\imgwidthAISTD]{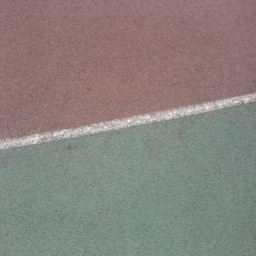} \\

\includegraphics[width=\imgwidthAISTD]{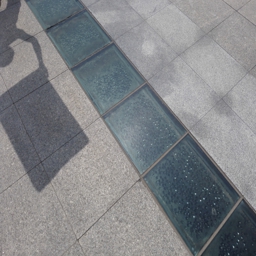} &
\includegraphics[width=\imgwidthAISTD]{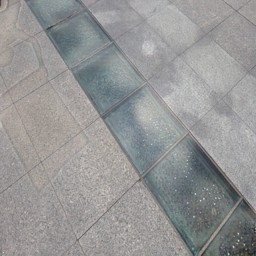} &
\includegraphics[width=\imgwidthAISTD]{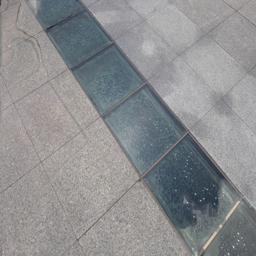} &
\includegraphics[width=\imgwidthAISTD]{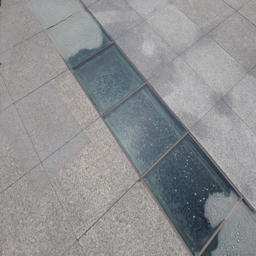} &
\includegraphics[width=\imgwidthAISTD]{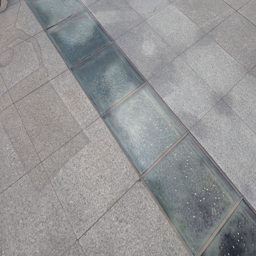} &
\includegraphics[width=\imgwidthAISTD]{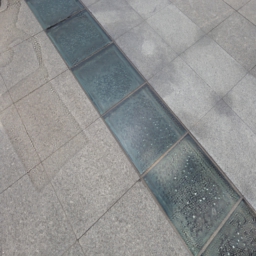} &
\includegraphics[width=\imgwidthAISTD]{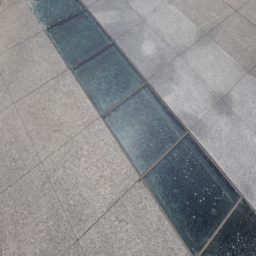} &
\includegraphics[width=\imgwidthAISTD]{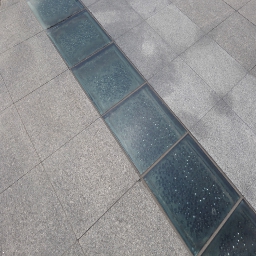} &
\includegraphics[width=\imgwidthAISTD]{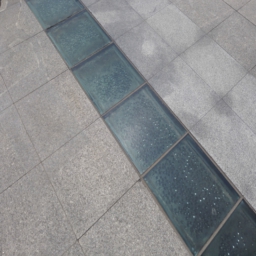} &
\includegraphics[width=\imgwidthAISTD]{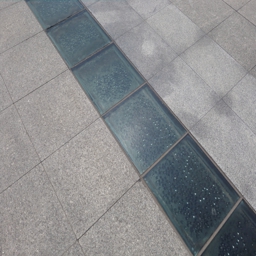} \\

\includegraphics[width=\imgwidthAISTD]{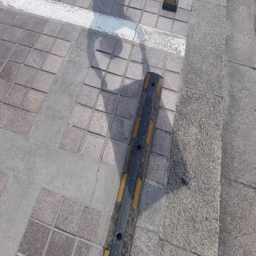} &
\includegraphics[width=\imgwidthAISTD]{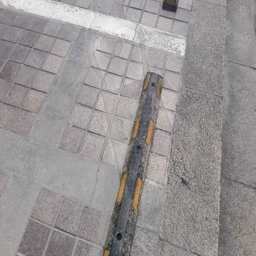} &
\includegraphics[width=\imgwidthAISTD]{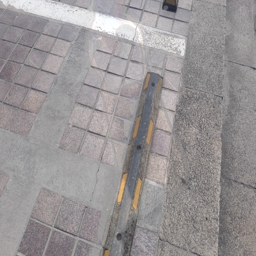} &
\includegraphics[width=\imgwidthAISTD]{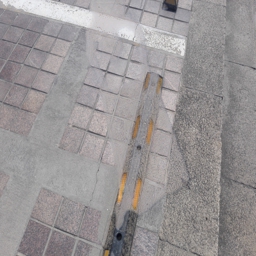} &
\includegraphics[width=\imgwidthAISTD]{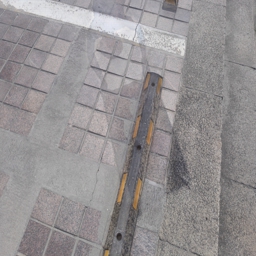} &
\includegraphics[width=\imgwidthAISTD]{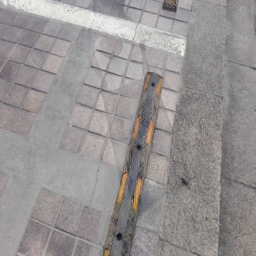} &
\includegraphics[width=\imgwidthAISTD]{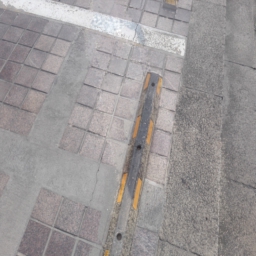} &
\includegraphics[width=\imgwidthAISTD]{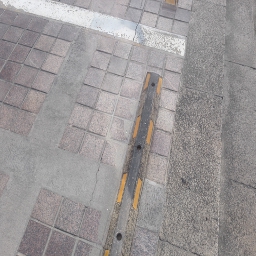} &
\includegraphics[width=\imgwidthAISTD]{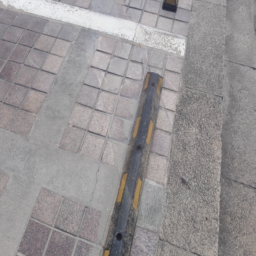} &
\includegraphics[width=\imgwidthAISTD]{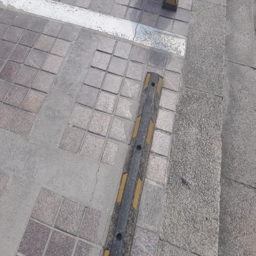} \\

\includegraphics[width=\imgwidthAISTD]{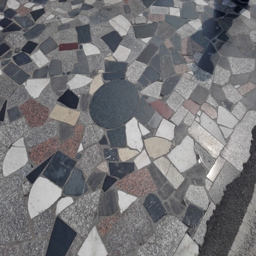} &
\includegraphics[width=\imgwidthAISTD]{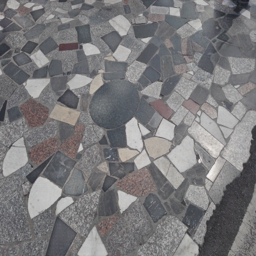} &
\includegraphics[width=\imgwidthAISTD]{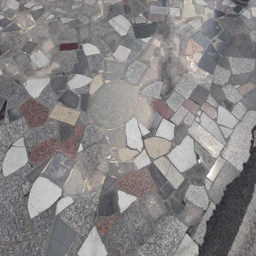} &
\includegraphics[width=\imgwidthAISTD]{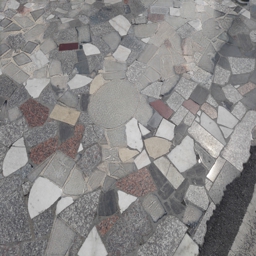} &
\includegraphics[width=\imgwidthAISTD]{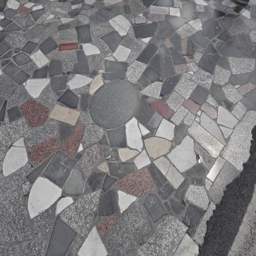} &
\includegraphics[width=\imgwidthAISTD]{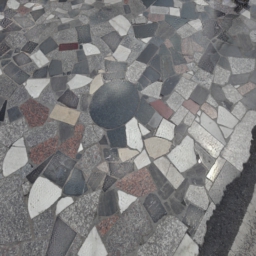} &
\includegraphics[width=\imgwidthAISTD]{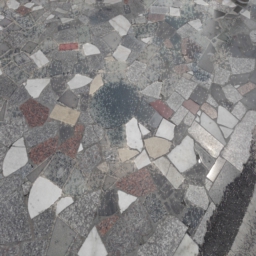} &
\includegraphics[width=\imgwidthAISTD]{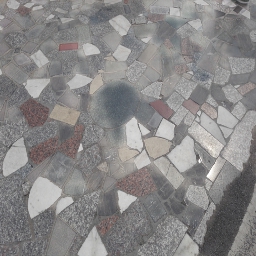} &
\includegraphics[width=\imgwidthAISTD]{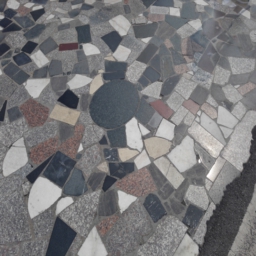} &
\includegraphics[width=\imgwidthAISTD]{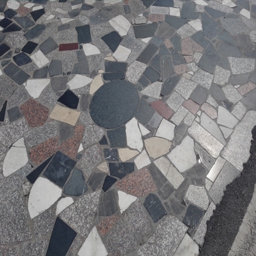} \\[-0.2em]

\scriptsize Input & 
\resizebox{\imgwidthAISTD}{!}{Mask-ShadowGAN} & 
\scriptsize Le \etal & 
\scriptsize G2R & 
\scriptsize LG-ShadowNet & 
\scriptsize DC-ShadowNet & 
\scriptsize ShadowDiffusion & 
\scriptsize HomoFormer & 
\scriptsize Ours & 
\scriptsize GT \\
\end{tabular}
\vspace{-1.2ex}
\caption{Qualitative comparison on the AISTD dataset \cite{le2019shadow, wang2018stacked}. 
\acs{name} produces cleaner and more natural results compared to other unsupervised methods, effectively restoring illumination and surface texture.}
\label{fig:demo_AISTD}
\vspace{-2.5ex}
\end{figure*}

%% file: figures/demo_SRD.tex
\begin{figure}[!ht]
\centering
\newcommand{\ncols}{8}
\setlength{\tabcolsep}{0.1em}

\newlength{\imgwidthSRD}
\setlength{\imgwidthSRD}{\dimexpr(0.98\linewidth - \ncols\tabcolsep)/\ncols\relax}

\begin{tabular}{*{\ncols}{c}}
\includegraphics[width=\imgwidthSRD]{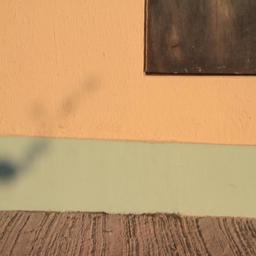} &
\includegraphics[width=\imgwidthSRD]{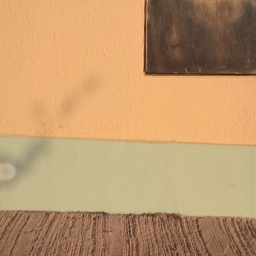} &
\includegraphics[width=\imgwidthSRD]{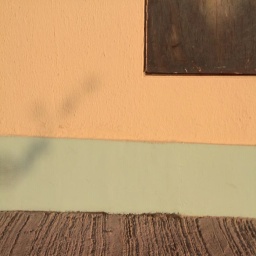} &
\includegraphics[width=\imgwidthSRD]{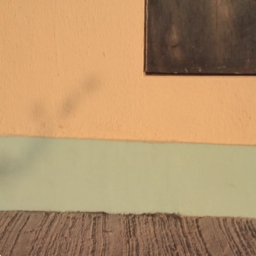} &
\includegraphics[width=\imgwidthSRD]{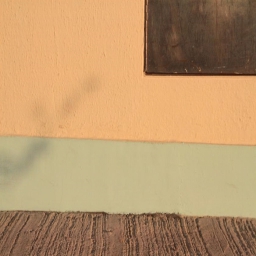} &
\includegraphics[width=\imgwidthSRD]{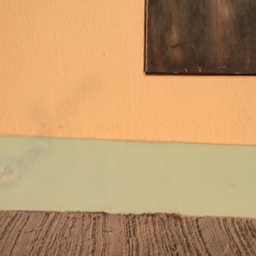} &
\includegraphics[width=\imgwidthSRD]{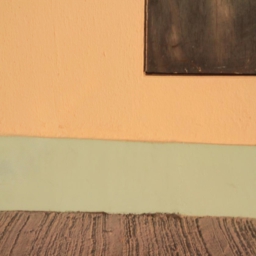} &
\includegraphics[width=\imgwidthSRD]{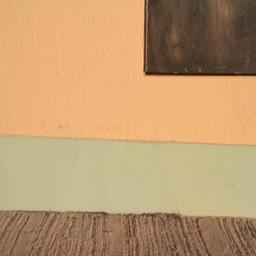} \\

\includegraphics[width=\imgwidthSRD]{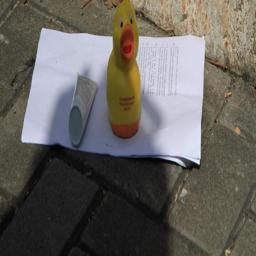} &
\includegraphics[width=\imgwidthSRD]{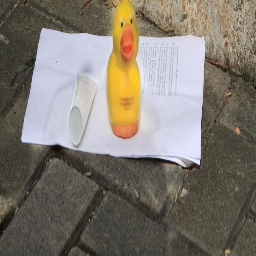} &
\includegraphics[width=\imgwidthSRD]{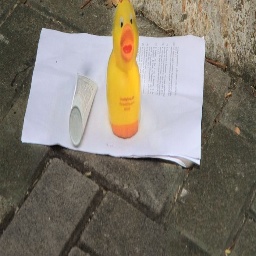} &
\includegraphics[width=\imgwidthSRD]{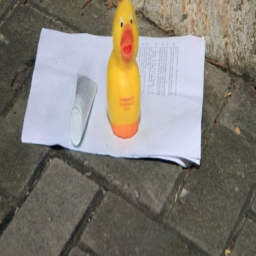} &
\includegraphics[width=\imgwidthSRD]{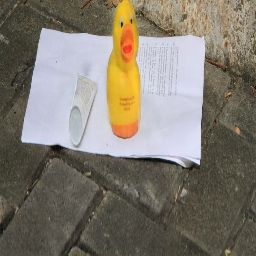} &
\includegraphics[width=\imgwidthSRD]{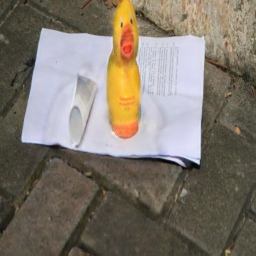} &
\includegraphics[width=\imgwidthSRD]{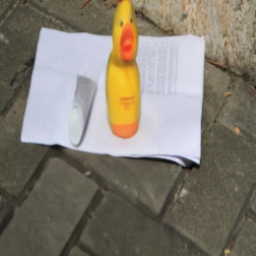} &
\includegraphics[width=\imgwidthSRD]{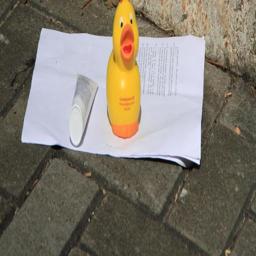}\\

\includegraphics[width=\imgwidthSRD]{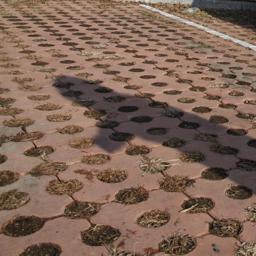} &
\includegraphics[width=\imgwidthSRD]{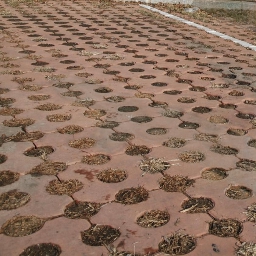} &
\includegraphics[width=\imgwidthSRD]{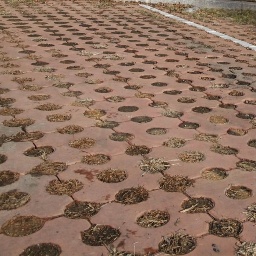} &
\includegraphics[width=\imgwidthSRD]{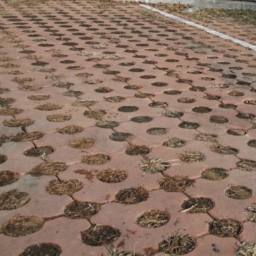} &
\includegraphics[width=\imgwidthSRD]{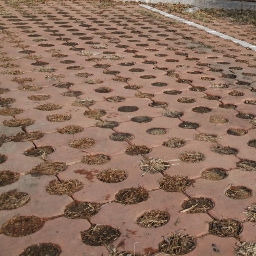} &
\includegraphics[width=\imgwidthSRD]{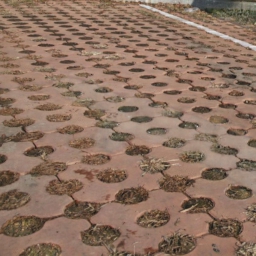} &
\includegraphics[width=\imgwidthSRD]{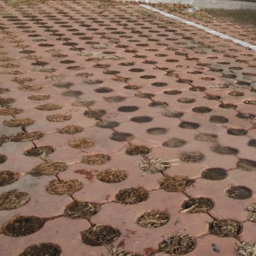} &
\includegraphics[width=\imgwidthSRD]{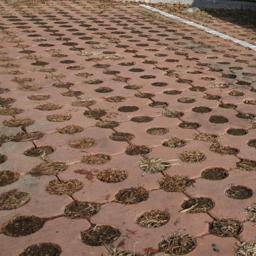} \\

\includegraphics[width=\imgwidthSRD]{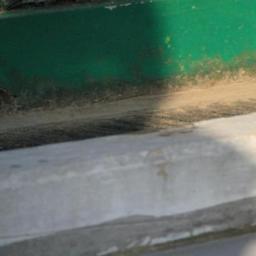} &
\includegraphics[width=\imgwidthSRD]{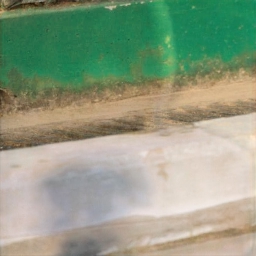} &
\includegraphics[width=\imgwidthSRD]{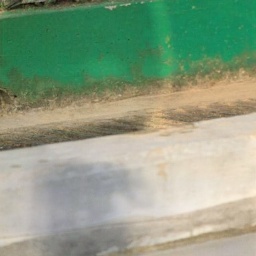} &
\includegraphics[width=\imgwidthSRD]{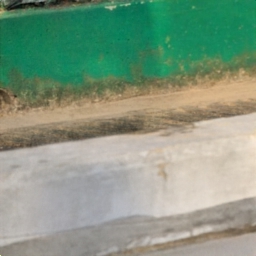} &
\includegraphics[width=\imgwidthSRD]{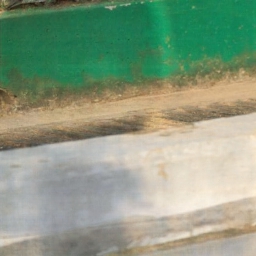} &
\includegraphics[width=\imgwidthSRD]{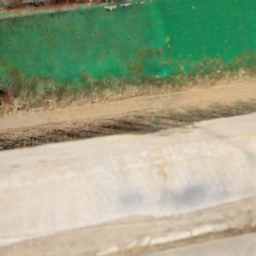} &
\includegraphics[width=\imgwidthSRD]{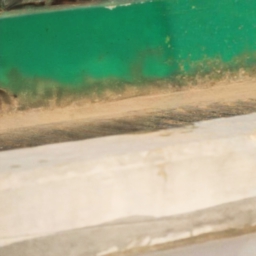} &
\includegraphics[width=\imgwidthSRD]{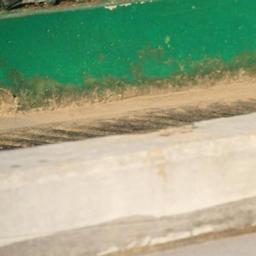}\\[-0.2em]

\tiny Input &
\tiny BMNet &
\resizebox{\imgwidthSRD}{!}{ShadowFormer} &
\resizebox{\imgwidthSRD}{!}{ShadowDiffusion} &
\tiny HomoFormer &
\resizebox{\imgwidthSRD}{!}{DC-ShadowNet} &
\tiny Ours &
\tiny GT \\
\end{tabular}

\vspace{-1.5ex}
\caption{Comparison on the SRD dataset \cite{qu2017deshadownet}. Compared to DC-ShadowNet, \acs{name} better removes shadows and preserves natural brightness and texture in non-shadow regions.}
\label{fig:demo_SRD}
\vspace{-3ex}
\end{figure}

%% file: tables/ablation.tex
\begin{table}[ht]
\caption{Ablation study of different components and parameter settings on the AISTD dataset. 
The best and second-best results are highlighted in \textbf{bold} and {\ul underlined}, respectively.}
\vspace{-1.2ex}
\label{tab:ablation}
\resizebox{\linewidth}{!}{
\begin{tabular}{l|ccc|ccc|cccc|c}
\hline
\multirow{2}{*}{Method} & \multicolumn{3}{c|}{Shadow Region} & \multicolumn{3}{c|}{Non-Shadow Region} & \multicolumn{4}{c|}{All Image} & \multirow{2}{*}{\begin{tabular}[c]{@{}c@{}}ARI abs.\\ gain\end{tabular}} \\
 & PSNR & SSIM & RMSE & PSNR & SSIM & RMSE & PSNR & SSIM & RMSE & BER &  \\ \hline
Input & 20.83 & 0.930 & 39.01 & 37.46 & 0.985 & 2.40 & 20.46 & 0.894 & 8.40 & - & - \\ \hline
Ours $\lambda_g=\lambda_p=0$ w/o group & 25.82 & 0.943 & 20.99 & 32.70 & 0.971 & 4.25 & 25.14 & 0.915 & 6.43 & 19.55 & - \\
Ours $\lambda_g=\lambda_p=0$ + DINOv3 group & 29.42 & 0.970 & 14.02 & 33.08 & 0.971 & 4.21 & 26.52 & 0.929 & 5.72 & 12.31 & 0.0 \\
Ours $\lambda_p=0$ & 32.18 & 0.981 & 10.08 & 34.02 & 0.975 & 3.72 & 28.96 & 0.943 & 4.58 & 8.02 & 0.30 \\
Ours w/o shadow generator & 37.06 & \textbf{0.990} & 5.89 & 37.08 & \textbf{0.985} & {\ul 2.52} & 32.98 & \textbf{0.965} & {\ul 3.14} & 2.17 & 0.33 \\
Ours confidence weight=1 & 36.12 & 0.987 & 6.02 & 35.42 & 0.980 & 3.04 & 31.89 & 0.960 & 3.42 & 2.25 & 0.22 \\
Ours $\lambda_p=0$ + LPIPS & 35.22 & 0.986 & 7.04 & 35.86 & 0.981 & 2.88 & 31.73 & 0.955 & 3.75 & 3.68 & 0.31 \\
Ours $\mathcal{L}_{\text{g}}$ on DINOv3 & 35.61 & 0.986 & 6.81 & 36.05 & 0.982 & 2.79 & 32.00 & 0.956 & 3.46 & 3.45 & 0.24 \\
Ours $\lambda_g=0.5$ & 36.62 & 0.988 & 5.92 & 36.31 & 0.983 & 2.96 & 32.72 & 0.961 & 3.38 & 1.83 & 0.30 \\
Ours $\lambda_p=0.5$ & 37.18 & \textbf{0.990} & {\ul 5.64} & 36.53 & 0.983 & 2.77 & 33.02 & {\ul 0.964} & 3.24 & {\ul 1.56} & 0.34 \\
Ours $\lambda_g=2.0$ & 37.43 & 0.988 & 5.91 & {\ul 37.09} & 0.983 & 2.56 & \textbf{33.39} & 0.963 & 3.16 & \textbf{1.50} & 0.34 \\
Ours $\lambda_p=2.0$ & 36.74 & 0.988 & 5.88 & 36.27 & 0.982 & 2.71 & 32.88 & 0.960 & 3.31 & 1.88 & 0.33 \\
Ours ($s=64$) & 36.28 & 0.988 & 6.18 & \textbf{37.38} & \textbf{0.985} & \textbf{2.43} & 33.22 & 0.963 & 3.15 & 1.73 & 0.33 \\
Ours ($s=256$) & \textbf{38.21} & \textbf{0.990} & \textbf{5.49} & 36.19 & 0.979 & 2.89 & 33.04 & 0.962 & 3.18 & 1.59 & 0.32 \\
Ours & {\ul 37.47} & {\ul 0.989} & 5.86 & 36.96 & {\ul 0.984} & 2.64 & {\ul 33.35} & {\ul 0.964} & \textbf{3.13} & 1.77 & 0.33 \\ \hline
Ours 1 image/GT group & 35.15 & 0.987 & 8.24 & 35.32 & 0.977 & 3.54 & 30.62 & 0.951 & 4.22 & 4.12 & - \\
Ours trained on SRD & 34.71 & 0.983 & 8.21 & 35.01 & 0.976 & 3.22 & 30.71 & 0.949 & 4.12 & 3.36 & - \\
Ours w GT group & 37.49 & {\ul 0.990} & 5.60 & \textbf{37.70} & \textbf{0.985} & {\ul 2.41} & 33.55 & {\ul 0.970} & {\ul 2.91} & 1.90 & - \\
Ours trained on SRD + AISTD & {\ul 38.28} & {\ul 0.990} & {\ul 5.52} & 36.70 & {\ul 0.984} & 2.64 & {\ul 34.05} & 0.965 & 2.92 & {\ul 1.68} & - \\
Ours iterative (\cref{sec:diffusion}) & \textbf{38.94} & \textbf{0.992} & \textbf{5.11} & {\ul 37.12} & \textbf{0.985} & \textbf{2.40} & \textbf{34.23} & \textbf{0.972} & \textbf{2.50} & \textbf{1.21} & \textbf{-} \\ \hline
\end{tabular}
}
\vspace{-2ex}
\end{table}

%% file: figures/segmentation.tex
\begin{figure}[ht]
\centering
\newcommand{\ncols}{6}
\setlength{\tabcolsep}{0.1em}

\newlength{\imgwidthSeg}
\setlength{\imgwidthSeg}{\dimexpr(0.98\linewidth - \ncols\tabcolsep)/\ncols\relax}

\begin{tabular}{*{\ncols}{c}}
\centering
\includegraphics[width=\imgwidthSeg]{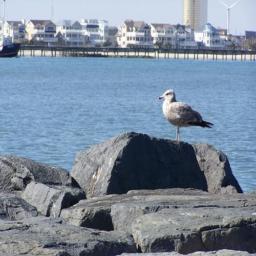} &
\includegraphics[width=\imgwidthSeg]{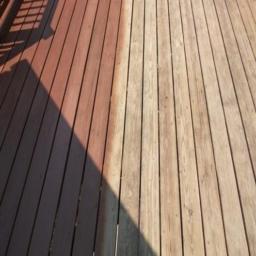} &
\includegraphics[width=\imgwidthSeg]{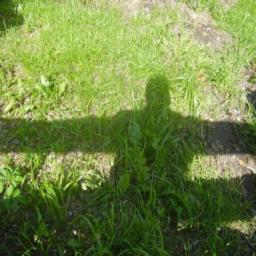} &
\includegraphics[width=\imgwidthSeg]{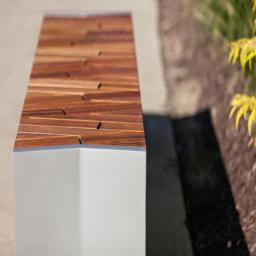} &
\includegraphics[width=\imgwidthSeg]{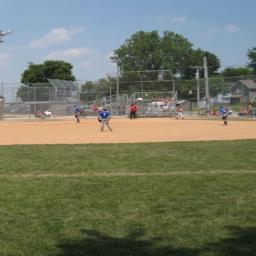} &
\includegraphics[width=\imgwidthSeg]{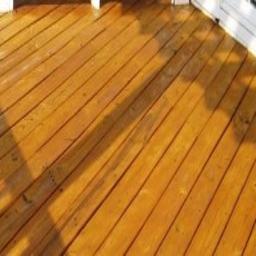} \\

\multicolumn{\ncols}{c}{Input images} \\

\includegraphics[width=\imgwidthSeg]{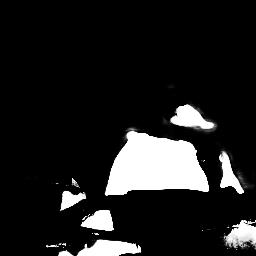} &
\includegraphics[width=\imgwidthSeg]{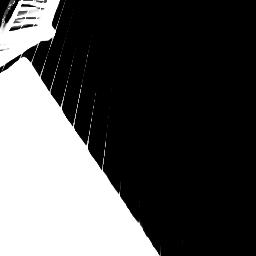} &
\includegraphics[width=\imgwidthSeg]{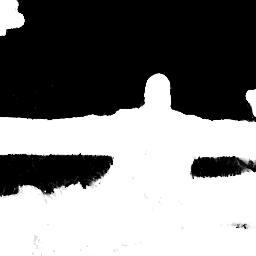} &
\includegraphics[width=\imgwidthSeg]{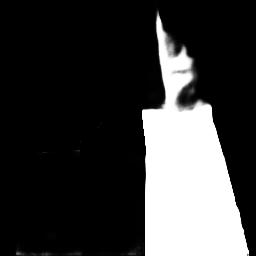} &
\includegraphics[width=\imgwidthSeg]{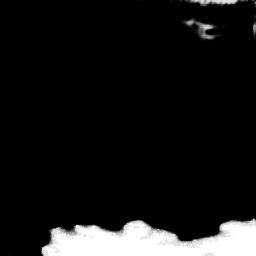} &
\includegraphics[width=\imgwidthSeg]{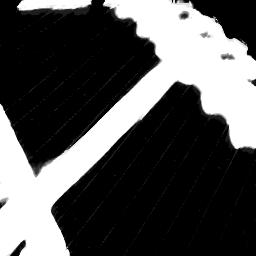} \\

\multicolumn{\ncols}{c}{Ours output matte} \\

\includegraphics[width=\imgwidthSeg]{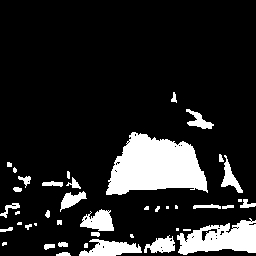} &
\includegraphics[width=\imgwidthSeg]{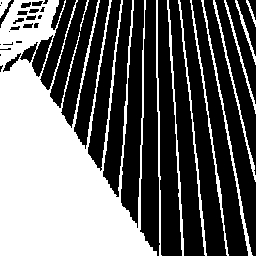} &
\includegraphics[width=\imgwidthSeg]{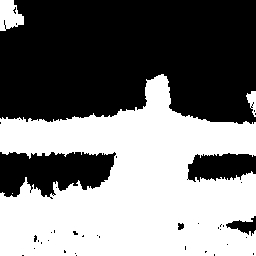} &
\includegraphics[width=\imgwidthSeg]{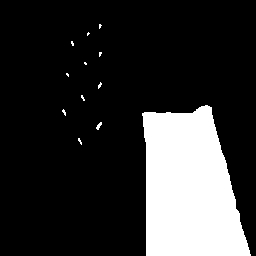} &
\includegraphics[width=\imgwidthSeg]{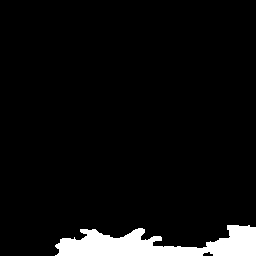} &
\includegraphics[width=\imgwidthSeg]{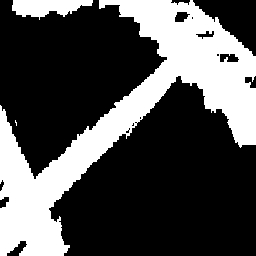} \\

\multicolumn{\ncols}{c}{Ground truth mask} \\

\includegraphics[width=\imgwidthSeg]{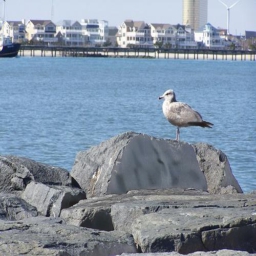} &
\includegraphics[width=\imgwidthSeg]{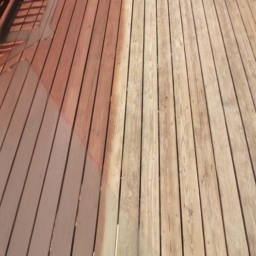} &
\includegraphics[width=\imgwidthSeg]{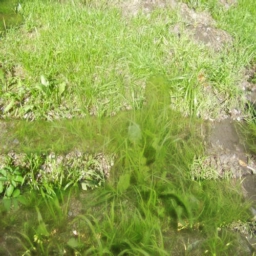} &
\includegraphics[width=\imgwidthSeg]{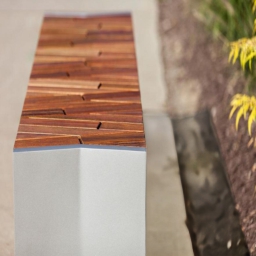} &
\includegraphics[width=\imgwidthSeg]{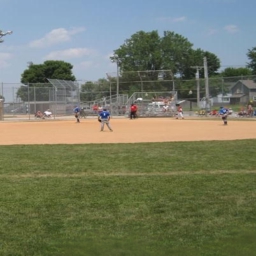} &
\includegraphics[width=\imgwidthSeg]{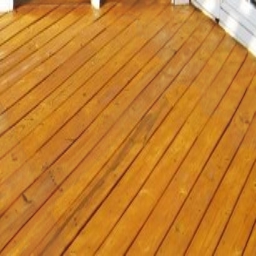} \\[-0.2em]

\multicolumn{\ncols}{c}{Ours output images} \\
\end{tabular}
\vspace{-1.2ex}
\caption{Unsupervised shadow segmentation on SBU dataset using the gating map $g$.}
\label{fig:segmentation}
\end{figure}

%% file: tables/segmentation.tex
\begin{table}[ht]
\centering
\caption{Comparison on shadow detection.}
\label{tab:segmentation}
\resizebox{\linewidth}{!}{
\begin{tabular}{l|l|ccccc|ccccc}
\hline
\multirow{2}{*}{Method} & \multirow{2}{*}{Setting} & \multicolumn{5}{c|}{SBU} & \multicolumn{5}{c}{AISTD} \\ \cline{3-12} 
 &  & BER & mIoU & Precision & Recall & F1 & BER & mIoU & Precision & Recall & F1 \\ \hline
MTMT \cite{chen2020multi} & \multirow{3}{*}{Supervised} & 6.32 & 75.61 & 87.53 & 84.55 & 86.01 & 3.15 & \textbf{87.97} & \textbf{90.75} & 96.06 & \textbf{92.28} \\
ECA \cite{fang2021robust} &  & 7.08 & 75.63 & \textbf{91.51} & 81.21 & 86.05 & 2.03 & \textbf{-} & \textbf{-} & - & - \\
SDDNet \cite{cong2023sddnet} &  & \textbf{5.39} & \textbf{76.05} & 83.96 & 88.47 & \textbf{86.16} & \textbf{1.27} & - & - & - & - \\ \hline
HSV-Shadow~\cite{kar2015moving} & \multirow{4}{*}{Unsupervised} & 16.51 & - & - & - & - & 18.91 & - & - & - & - \\
ASD~\cite{freitas2017automatic} &  & 15.69 & - & - & - & - & 12.45 & - & - & - & - \\
Zhang~\etal~\cite{zhang2024deep} &  & 7.13 & - & - & - & - & 5.57 & - & - & - & - \\
Ours &  & 6.84 & 71.46 & 77.92 & \textbf{91.84} & 84.29 & 1.68 & 82.41 & 83.87 & \textbf{97.56} & 90.20 \\ \hline
\end{tabular}
}
\end{table}

%% file: tables/video.tex
\begin{table}[ht]
\caption{Quantitative comparison results on video shadow removal. All metrics are computed in the moving-shadow region; RMSE* uses a threshold of 40, while others use 80. '+' denotes models fine-tuned for one epoch.}
\label{tab:video}
\centering
\resizebox{\linewidth}{!}{
\begin{tabular}{l|cccc}
\hline
Method & PSNR & SSIM & RMSE & RMSE* \\ \hline
SP+M-Net \cite{le2019shadow} & - & - & - & 22.2 \\
Le \etal \cite{le2020shadow} & - & - & - & 20.9 \\
Le \etal + \cite{le2020shadow} & - & - & - & 18.0 \\
Mask-ShadowGAN \cite{hu2019mask} & 20.38 & 0.887 & 22.7 & 19.6 \\
LG-ShadowNet \cite{liu2021shadow} & 20.68 & 0.880 & 22.0 & 18.3 \\
G2R \cite{liu2021from} & 21.07 & 0.882 & 21.8 & 18.8 \\
G2R+ \cite{liu2021from} & - & - & {\ul 18.7} & - \\
Guo \etal \cite{guo2023boundary} & 22.23 & 0.893 & - & 17.7 \\
\acs{name} (Ours) & {\ul 22.52} & {\ul 0.896} & {\ul 19.5} & {\ul 17.0} \\
\acs{name}+ (Ours) & \textbf{22.94} & \textbf{0.911} & \textbf{18.6} & \textbf{14.4} \\ \hline
\end{tabular}
}
\end{table}

%% file: figures/video.tex
\begin{figure}[ht]
\centering
\newcommand{\ncols}{4}
\setlength{\tabcolsep}{0.1em}

\newlength{\imgwidthVideo}
\setlength{\imgwidthVideo}{\dimexpr(0.96\linewidth - \ncols\tabcolsep)/\ncols\relax}

\begin{tabular}{*{\ncols}{c}}
\includegraphics[width=\imgwidthVideo]{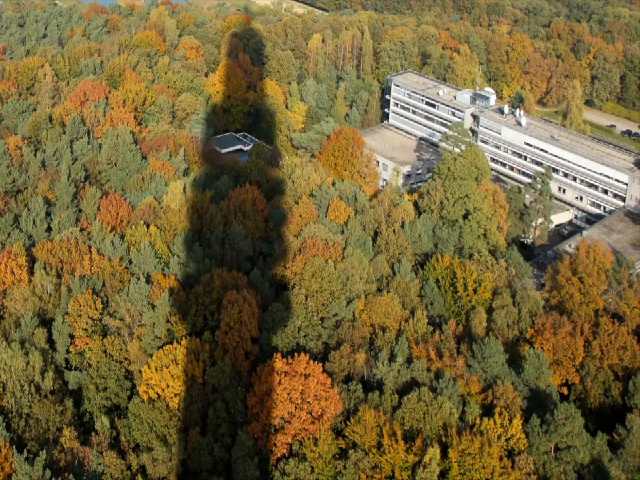} &
\includegraphics[width=\imgwidthVideo]{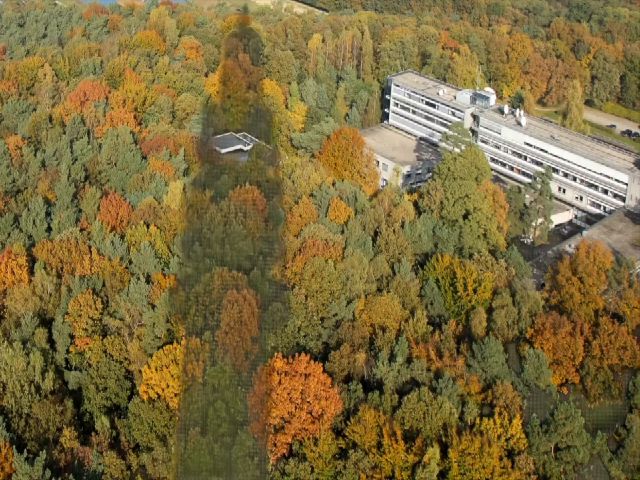} &
\includegraphics[width=\imgwidthVideo]{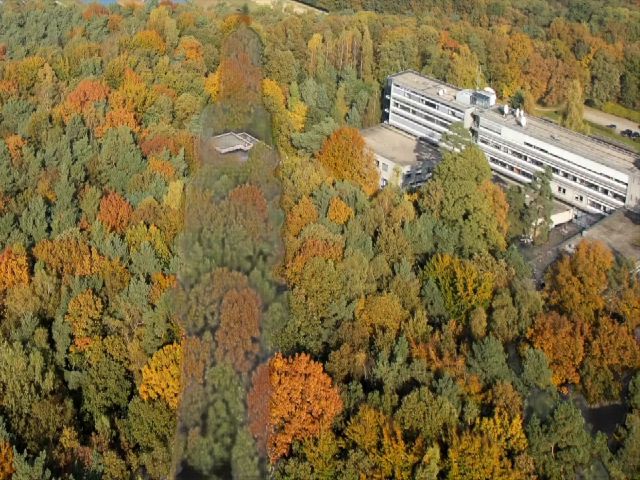} &
\includegraphics[width=\imgwidthVideo]{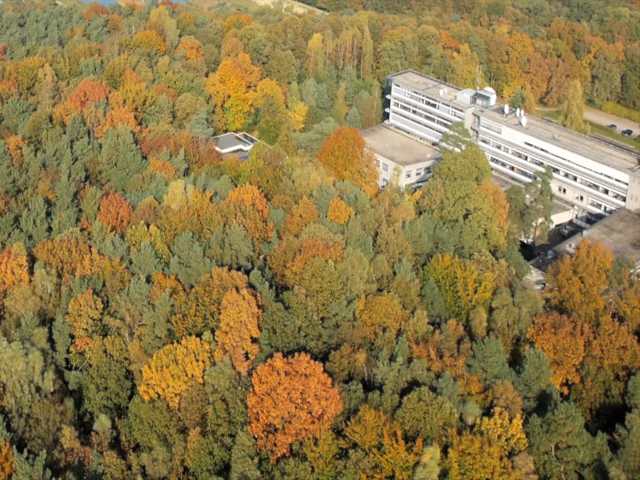} \\[-0.2em]

\scriptsize Input &
\scriptsize LG-ShadowNet &
\scriptsize G2R &
\scriptsize GT \\[0.1em]

\includegraphics[width=\imgwidthVideo]{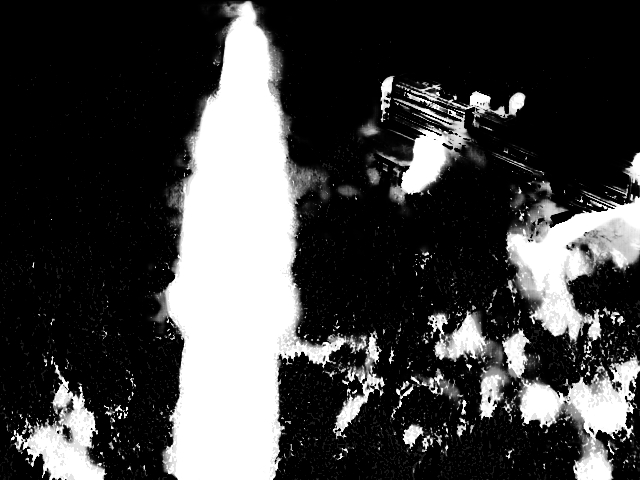} &
\includegraphics[width=\imgwidthVideo]{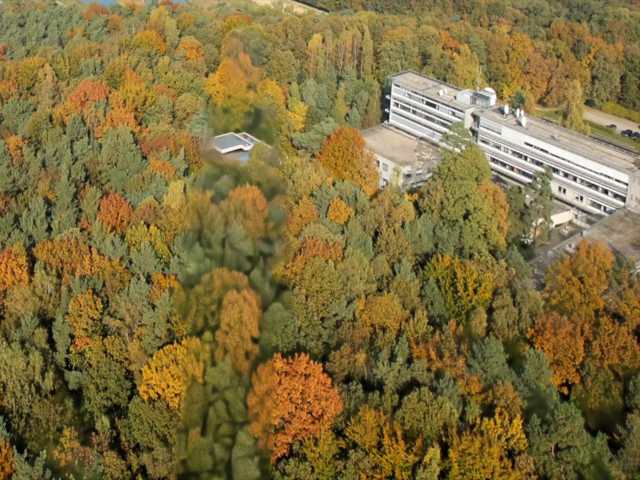} &
\includegraphics[width=\imgwidthVideo]{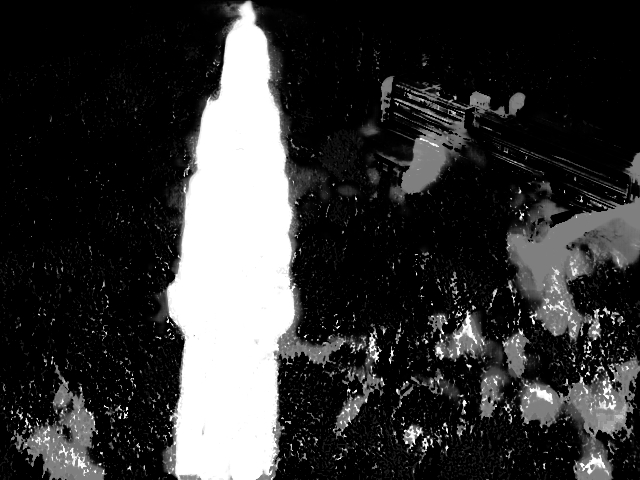} &
\includegraphics[width=\imgwidthVideo]{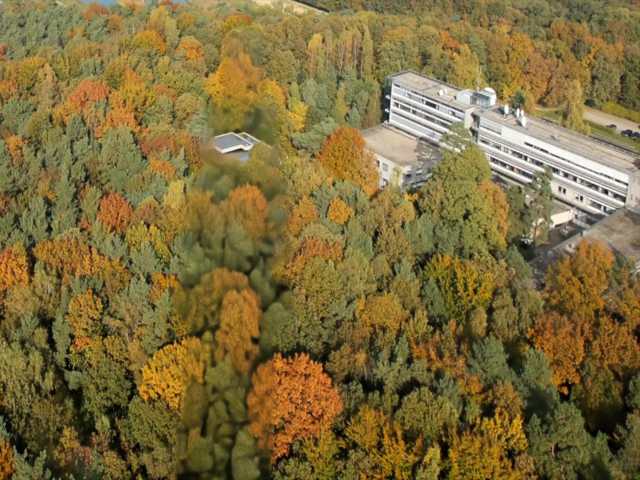} \\[-0.2em]

\scriptsize Ours matte &
\scriptsize Ours &
\scriptsize Ours+ matte &
\scriptsize Ours+ \\
\end{tabular}
\vspace{-1.2ex}
\caption{Qualitative results on the video shadow dataset~\cite{le2020shadow}.}
\label{fig:video}
\end{figure}

%% file: figures/failure.tex
\begin{figure}[ht]
\centering
\newcommand{\ncols}{6}
\setlength{\tabcolsep}{0.1em}

\newlength{\imgwidthFailure}
\setlength{\imgwidthFailure}{\dimexpr(0.9\linewidth - \ncols\tabcolsep)/\ncols\relax}

\begin{tabular}{*{\ncols}{c}}

\includegraphics[width=\imgwidthFailure]{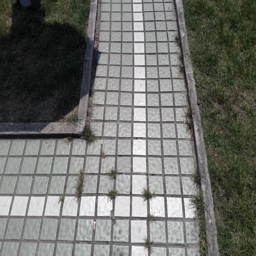} &
\imgpsnr{\imgwidthFailure}{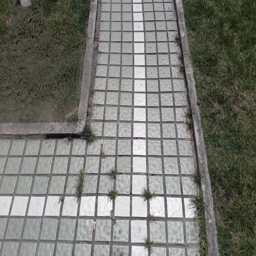}{31.15} &
\imgpsnr{\imgwidthFailure}{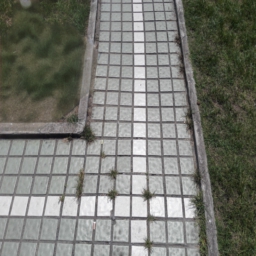}{31.78} & 
\imgpsnr{\imgwidthFailure}{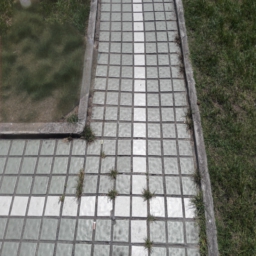}{31.69} & 
\imgpsnr{\imgwidthFailure}{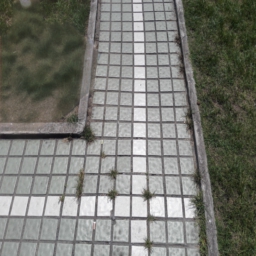}{31.24} & 
\includegraphics[width=\imgwidthFailure]{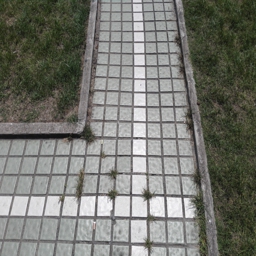} \\[-0.3ex]

\includegraphics[width=\imgwidthFailure]{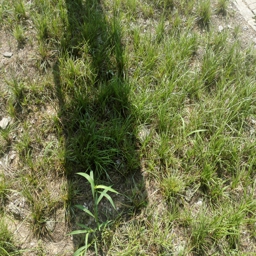} &
\imgpsnr{\imgwidthFailure}{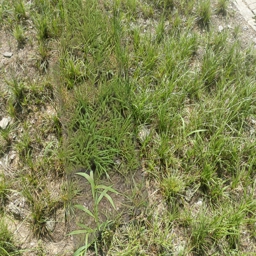}{25.22} &
\imgpsnr{\imgwidthFailure}{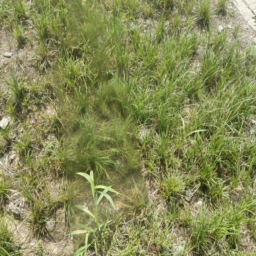}{25.81} &
\imgpsnr{\imgwidthFailure}{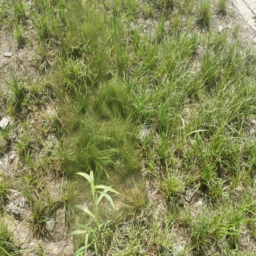}{25.68} &
\imgpsnr{\imgwidthFailure}{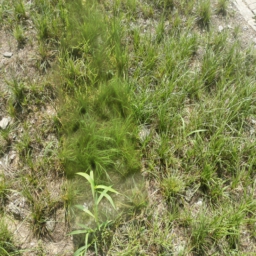}{25.34} &
\includegraphics[width=\imgwidthFailure]{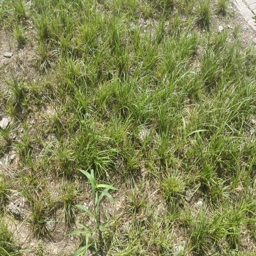} \\[-0.3ex]

\includegraphics[width=\imgwidthFailure]{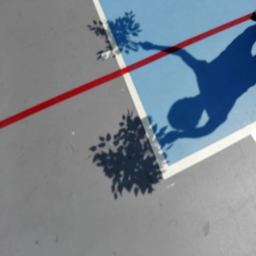} &
\imgpsnr{\imgwidthFailure}{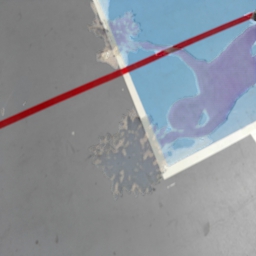}{29.50} &
\imgpsnr{\imgwidthFailure}{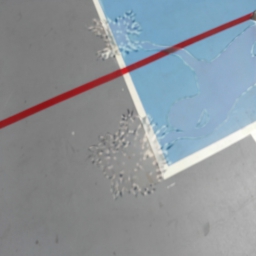}{31.21} & 
\imgpsnr{\imgwidthFailure}{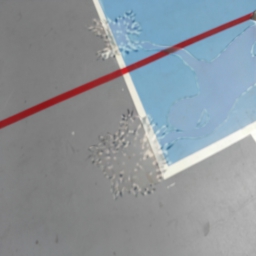}{31.21} & 
\imgpsnr{\imgwidthFailure}{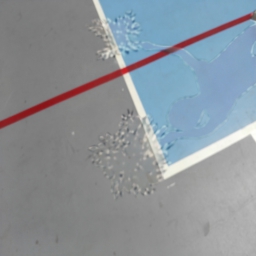}{31.60} & 
\includegraphics[width=\imgwidthFailure]{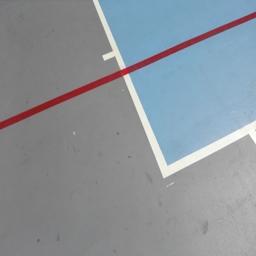} \\[-0.3ex]

\includegraphics[width=\imgwidthFailure]{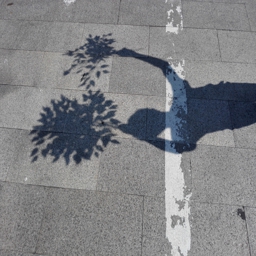} &
\imgpsnr{\imgwidthFailure}{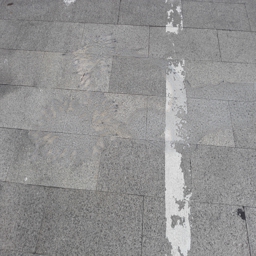}{31.33} &
\imgpsnr{\imgwidthFailure}{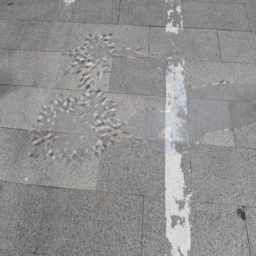}{32.03} & 
\imgpsnr{\imgwidthFailure}{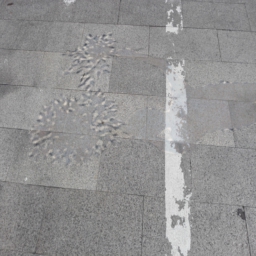}{33.14} & 
\imgpsnr{\imgwidthFailure}{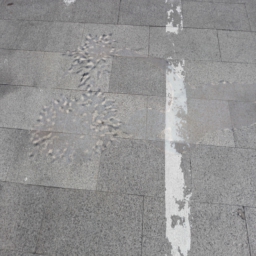}{31.39} & 
\includegraphics[width=\imgwidthFailure]{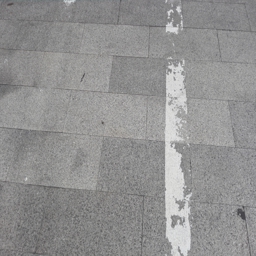} \\[-0.3ex]

\tiny Input &
\tiny G2R &
\tiny Ours $(s=64)$ & 
\tiny Ours $(s=128)$ & 
\tiny Ours $(s=256)$ & 
\tiny GT \\[-1.0ex]

\end{tabular}

\caption{Failure cases of our method. Regions with fine details, small shadow structures may appear slightly blurred due to the learning strategy. We include G2R for comparison, while our method still achieves higher PSNR.}
\label{fig:failure}
\vspace{-3ex}
\end{figure}

%% file: sec/5_conclusion.tex
\section{Conclusion}
\label{sec:conclusion}

We introduced \ac{name}, an unsupervised framework for shadow removal that learns without relying on shadow masks or shadow-free references. By enforcing consistency among images at pixel, global, and patch-wise levels; the model effectively learns to restore illumination and remove shadows in a self-supervised manner. Extensive experiments show that \ac{name} achieves competitive, and in some cases superior, results compared with unsupervised state-of-the-art approaches. Overall, our results indicate that consistency learning offers a practical and effective direction for unsupervised shadow removal.

%% file: sec/X_supp.tex
\clearpage
\setcounter{page}{1}
\maketitlesupplementary

\input{sec/6_diffusion}
\input{sec/7_proof}
\input{sec/8_more}

%% file: sec/6_diffusion.tex
\section{Iterative blending strategy}
\label{sec:diffusion}

Following Ho et al.~\cite{ho2020denoising}, the diffusion model defines a forward process that gradually perturbs a clean image $x^0$ by adding Gaussian noise over $T$ steps. The transition at each step $t$ is given by:
\begin{equation}
    q(x^t | x^{t-1}) = \mathcal{N}(x^t; \sqrt{1-\beta_t}\, x^{t-1}, \beta_t \mathbf{I}),
\end{equation}
where $\beta_t$ denotes the noise variance at step $t$.  
The cumulative effect of the forward process allows $x^t$ to be directly sampled from $x^0$ as
\begin{equation}
    q(x^t \mid x^0) = \mathcal{N}\!\left(x^t; \sqrt{\bar{\alpha}_t}\, x^0, (1-\bar{\alpha}_t)\mathbf{I}\right),
\end{equation}
where $\alpha_t = 1 - \beta_t$ and $\bar{\alpha}_t = \prod_{i=1}^{t}\alpha_i$.

The reverse process aims to iteratively recover $x^0$ from a pure Gaussian noise sample $x^T \!\sim\! \mathcal{N}(0,\mathbf{I})$.  
At each step, the posterior distribution of $x^{t-1}$ conditioned on $x^t$ is expressed as
\begin{equation}
    p_\theta(x^{t-1} \mid x^t) = 
    \mathcal{N}\!\left(x^{t-1}; \mu_\theta(x^t, t), \Sigma_\theta(x^t, t)\right),
\end{equation}
where $\mu_\theta(x^t, t)$ and $\Sigma_\theta(x^t, t)$ denote the mean and variance of the reverse Gaussian transition.  
The mean term can be reparameterized by the predicted noise $\epsilon_\theta(x^t, t)$ as
\begin{equation}
    \mu_\theta(x^t, t) =
    \frac{1}{\sqrt{\alpha_t}}
    \left(x^t - \frac{\beta_t}{\sqrt{1-\bar{\alpha}_t}}\, \epsilon_\theta(x^t, t)\right).
\end{equation}
This formulation provides the theoretical foundation for diffusion-based image generation and restoration models used in subsequent works.

Recent diffusion-based approaches have shown strong capability in refining illumination and texture restoration for shadow removal tasks~\cite{guo2023boundary}. Motivated by this, we extend our method with a diffusion-inspired formulation that interprets shadows as structured noise. We draw inspiration from the iterative denoising principle of diffusion models~\cite{ho2020denoising}, and formulate a deterministic refinement process that progressively removes structured shadow noise over $T$ inference steps. The key idea is to apply the model iteratively over multiple steps, where each step partially updates the image toward a shadow-free state while preserving its original structure.

Formally, given a shadow image $x^T$, and an intermediate estimate $x^t$, the update rule at step $t$ is defined as

\begin{equation}
    x^{t-1} = \left( 1-\frac{t-1}{T} \right)\hat{U}\left( x^{t} \right) + \frac{t-1}{T}x^T,
\end{equation}
where $\hat{U}\left( \cdot \right)$ denotes our pretrained shadow removal network and $T$ is the total number of inference steps.

This formulation divides the inference into $T$ gradual refinements. At each iteration, the network output $\hat{U}\left( x^{t} \right)$ replaces only a fraction of the original $x^T$, effectively diffusing the illumination correction across steps. As $t$ decreases, from $T$ down to $1$, the process progressively suppresses shadow artifacts while maintaining fine-grained texture and tone consistency.

\input{figures/diffusion.tex}
\input{tables/diffusion}

Following~\cite{jin2024des3}, a DINOv3~\cite{simeoni2025dinov3} similarity metric is employed between the intermediate keys of the input image and the output image during the reverse refinement process to select the most visually consistent result among the intermediate steps. This similarity-based selection ensures that the final output maintains both texture coherence and structural fidelity without introducing over-smoothing artifacts. In practice, we use $T = 100$ inference steps, corresponding to approximately $12$~TFLOPs including DINOv3. For reference, the state-of-the-art diffusion-based unsupervised shadow removal method of~\cite{guo2023boundary} requires about 61~TFLOPs. Despite being computationally lighter, inference with our iterative blending strategy achieves higher restoration quality and better illumination consistency.

As shown in \cref{fig:diffusion}, we provide a qualitative comparison between our iterative blending strategy and our base model without diffusion refinement. Since the only existing diffusion-based unsupervised shadow removal method~\cite{guo2023boundary} does not release source code or output images, we include DC-ShadowNet~\cite{jin2021dc} as an additional reference for visual comparison. The results clearly show that the diffusion-based extension yields cleaner illumination recovery and fewer residual artifacts, particularly in regions with complex shadow boundaries.

Quantitative results are summarized in \cref{tab:diffusion}, where the proposed diffusion-inspired extension achieves superior performance compared with prior diffusion-based approaches, confirming the effectiveness of iterative deterministic refinement for unsupervised shadow removal. Here, $k$ denotes the number of direct refinement steps applied to the input image before the final output. For TTA, we use five test-time augmentations, including the original image, flipped, and rotated variants, and aggregate the restored outputs by mean or median ensembling.

%% file: figures/diffusion.tex
\begin{figure}[h]
\centering
\newcommand{\ncols}{6}
\setlength{\tabcolsep}{0.1em}

\newlength{\imgwidthsixcolsoneside}
\setlength{\imgwidthsixcolsoneside}{\dimexpr(0.98\linewidth - \ncols\tabcolsep)/\ncols\relax}

\begin{tabular}{*{\ncols}{c}}
\centering
\includegraphics[width=\imgwidthsixcolsoneside]{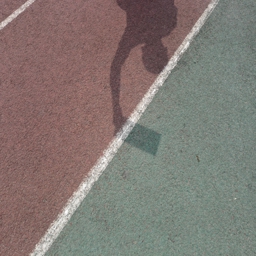} &
\includegraphics[width=\imgwidthsixcolsoneside]{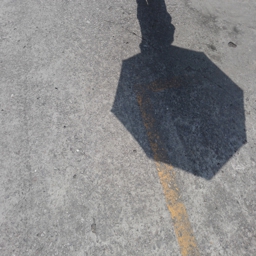} &
\includegraphics[width=\imgwidthsixcolsoneside]{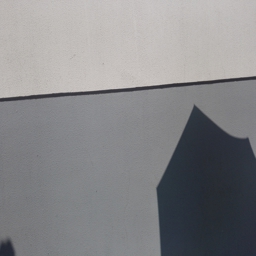} &
\includegraphics[width=\imgwidthsixcolsoneside]{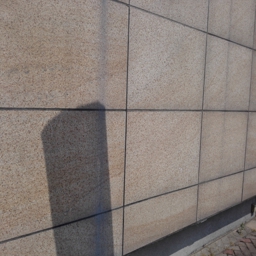} &
\includegraphics[width=\imgwidthsixcolsoneside]{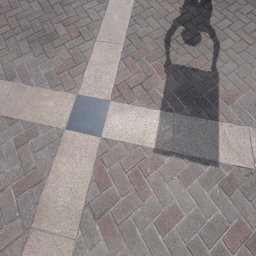} &
\includegraphics[width=\imgwidthsixcolsoneside]{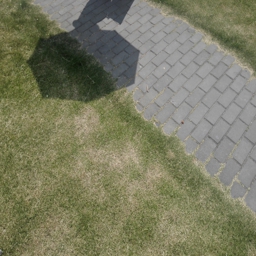} \\

\multicolumn{6}{c}{(a) Input shadow images} \\

\includegraphics[width=\imgwidthsixcolsoneside]{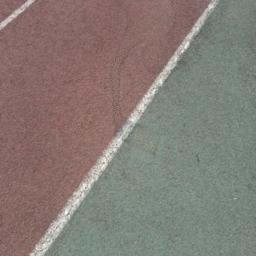} &
\includegraphics[width=\imgwidthsixcolsoneside]{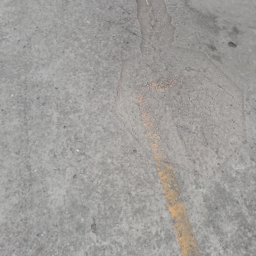} &
\includegraphics[width=\imgwidthsixcolsoneside]{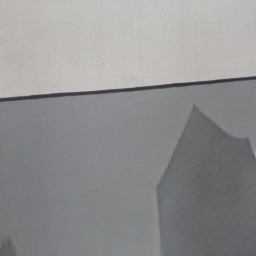} &
\includegraphics[width=\imgwidthsixcolsoneside]{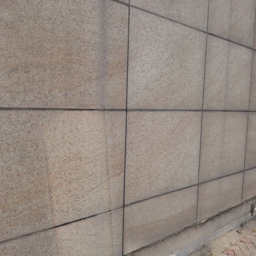} &
\includegraphics[width=\imgwidthsixcolsoneside]{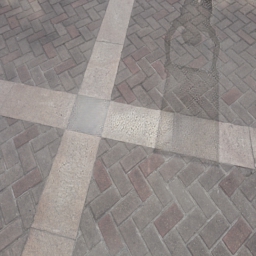} &
\includegraphics[width=\imgwidthsixcolsoneside]{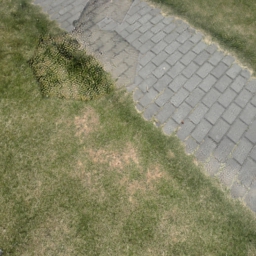} \\

\multicolumn{6}{c}{(b) DC-ShadowNet} \\

\includegraphics[width=\imgwidthsixcolsoneside]{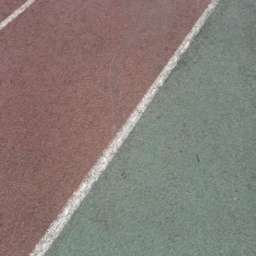} &
\includegraphics[width=\imgwidthsixcolsoneside]{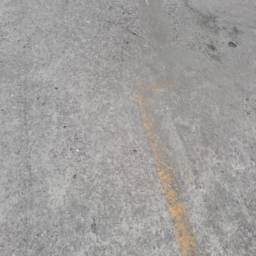} &
\includegraphics[width=\imgwidthsixcolsoneside]{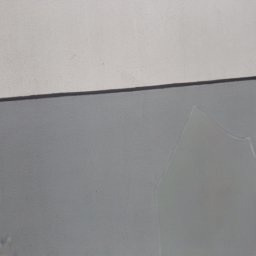} &
\includegraphics[width=\imgwidthsixcolsoneside]{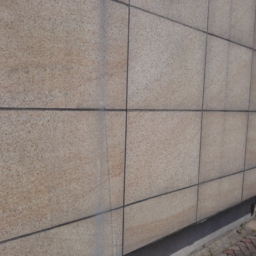} &
\includegraphics[width=\imgwidthsixcolsoneside]{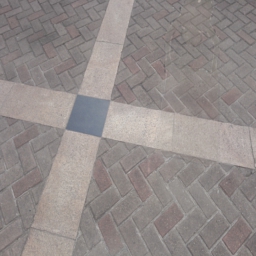} &
\includegraphics[width=\imgwidthsixcolsoneside]{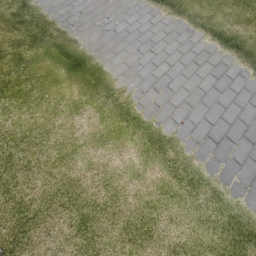} \\

\multicolumn{6}{c}{(c) Ours output images} \\

\includegraphics[width=\imgwidthsixcolsoneside]{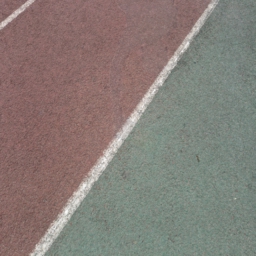} &
\includegraphics[width=\imgwidthsixcolsoneside]{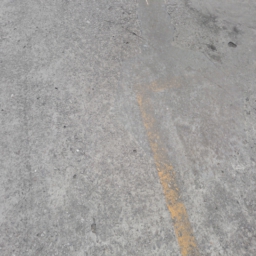} &
\includegraphics[width=\imgwidthsixcolsoneside]{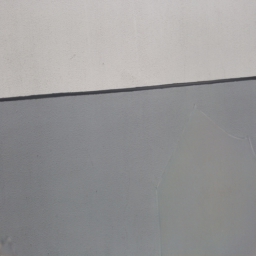} &
\includegraphics[width=\imgwidthsixcolsoneside]{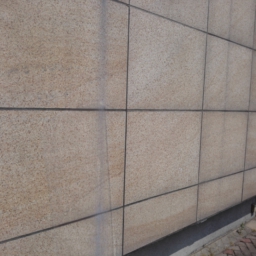} &
\includegraphics[width=\imgwidthsixcolsoneside]{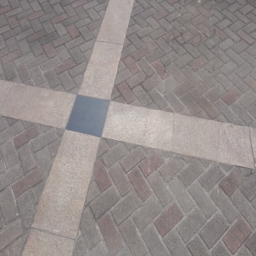} &
\includegraphics[width=\imgwidthsixcolsoneside]{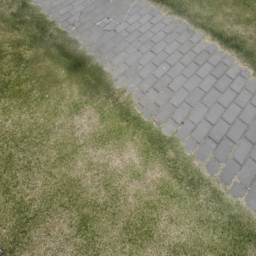} \\

\multicolumn{6}{c}{(d) Ours iterative output images} \\

\includegraphics[width=\imgwidthsixcolsoneside]{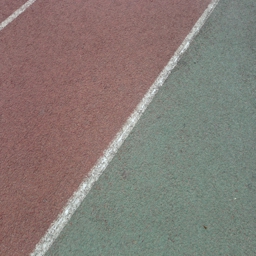} &
\includegraphics[width=\imgwidthsixcolsoneside]{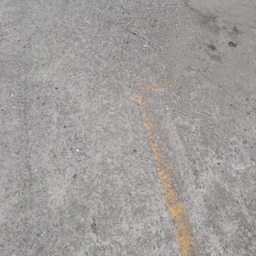} &
\includegraphics[width=\imgwidthsixcolsoneside]{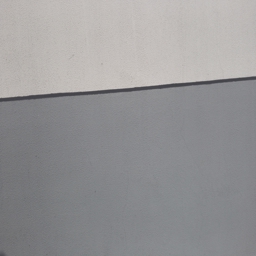} &
\includegraphics[width=\imgwidthsixcolsoneside]{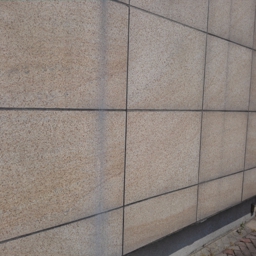} &
\includegraphics[width=\imgwidthsixcolsoneside]{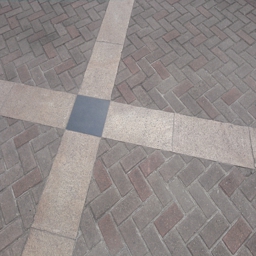} &
\includegraphics[width=\imgwidthsixcolsoneside]{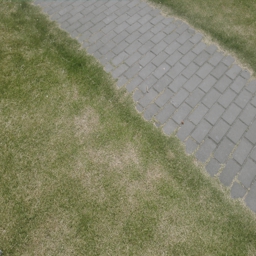} \\

\multicolumn{6}{c}{(e) Ground truth} \\

\end{tabular}

\caption{Qualitative comparison on the AISTD dataset~\cite{le2019shadow, wang2018stacked}. We compare DC-ShadowNet~\cite{jin2021dc}, our base model, and our iterative extension.}
\label{fig:diffusion}
\end{figure}

%% file: tables/diffusion.tex
\begin{table*}[ht]
\caption{Quantitative comparison of the proposed diffusion-inspired extension with the diffusion-based unsupervised shadow removal method~\cite{guo2023boundary} on the AISTD dataset. The proposed extension achieves higher performance on most metrics while requiring lower computational cost ($12$~TFLOPs vs.~61~TFLOPs).}
\label{tab:diffusion}
\centering
\begin{tabular}{l|ccc|ccc|ccc}
\hline
\multirow{2}{*}{Method} & \multicolumn{3}{c|}{Shadow} & \multicolumn{3}{c|}{Non-Shadow} & \multicolumn{3}{c}{All} \\
 & PSNR & SSIM & RMSE & PSNR & SSIM & RMSE & PSNR & SSIM & RMSE \\ \hline
Guo \etal \cite{guo2023boundary} (w/ detected mask) & 35.71 & 0.986 & 7.60 & 36.39 & 0.981 & 2.70 & 32.11 & 0.959 & 3.50 \\
Guo \etal \cite{guo2023boundary} (w/ GT mask) & 35.91 & 0.986 & 7.60 & \textbf{37.27} & {\ul 0.984} & \textbf{2.40} & 32.73 & 0.962 & 3.30 \\ \hline
Ours & 37.47 & 0.989 & 5.86 & 36.96 & {\ul 0.984} & 2.64 & 33.35 & 0.964 & 3.13 \\
Ours $k=1$ & 37.18 & 0.988 & 6.08 & 36.84 & 0.983 & 2.70 & 33.12 & 0.963 & 3.24 \\
Ours $k=2$ & 36.91 & 0.987 & 6.31 & 36.69 & 0.982 & 2.77 & 32.93 & 0.962 & 3.36 \\
Ours $k=3$ & 36.67 & 0.986 & 6.55 & 36.56 & 0.981 & 2.84 & 32.74 & 0.961 & 3.47 \\
Ours $k=4$ & 36.45 & 0.985 & 6.79 & 36.43 & 0.981 & 2.91 & 32.58 & 0.960 & 3.58 \\
Ours $x_{t-1}=U(0.5x_T+0.5x_t)$ & 36.98 & 0.987 & 6.22 & 36.73 & 0.982 & 2.75 & 33.00 & 0.962 & 3.31 \\
Ours TTA mean-ensemble & 37.69 & {\ul 0.990} & 5.62 & 37.04 & {\ul 0.984} & 2.57 & 33.52 & 0.966 & 3.01 \\
Ours TTA med-ensemble & {\ul 37.81} & {\ul 0.990} & {\ul 5.48} & 37.08 & {\ul 0.984} & {\ul 2.51} & {\ul 33.63} & {\ul 0.967} & {\ul 2.93} \\
Ours iterative & \textbf{38.94} & \textbf{0.992} & \textbf{5.11} & {\ul 37.12} & \textbf{0.985} & \textbf{2.40} & \textbf{34.23} & \textbf{0.972} & \textbf{2.50} \\ \hline
\end{tabular}
\end{table*}

%% file: sec/7_proof.tex
\section{Contrastive loss weight analysis}
\label{sec:proof}

In this section, we analyze the effect of the stop-gradient confidence weighting used in the global contrastive loss. For simplicity, we consider the loss associated with an anchor feature $z_i$. All features are $\ell_2$-normalized. Let
\begin{equation}
    c_i = z_i^\top \hat{z}_i \in [-1,1],
    \qquad
    w_i = \operatorname{sg}(1+c_i),
\end{equation}
where $(z_i,\hat{z}_i)$ is the positive pair and
$\operatorname{sg}(\cdot)$ denotes the stop-gradient operation.

The contrastive loss for sample $i$ is
\begin{equation}
\resizebox{\linewidth}{!}{$
    \mathcal{L}_{g}^{(i)}
    =
    -w_i
    \log
    \frac{
        \exp(c_i/\tau)
    }{
        \exp(c_i/\tau)
        +
        \sum_{j\neq i}\exp(z_i^\top z_j/\tau)
        +
        \sum_{j\neq i}\exp(z_i^\top \hat{z}_j/\tau)
    }.
$}
\end{equation}

For compactness, we denote the denominator by
\begin{equation}
\begin{aligned}
    D_i
    &=
    \exp(c_i/\tau)
    +
    \sum_{j\neq i}\exp(z_i^\top z_j/\tau) \\
    &\quad+
    \sum_{j\neq i}\exp(z_i^\top \hat{z}_j/\tau),
\end{aligned}
\end{equation}
and define the corresponding softmax probabilities as
\begin{equation}
    p_i^{+}
    =
    \frac{\exp(c_i/\tau)}{D_i},
\end{equation}
\begin{equation}
    p_{ij}
    =
    \frac{\exp(z_i^\top z_j/\tau)}{D_i},
    \qquad
    \hat{p}_{ij}
    =
    \frac{\exp(z_i^\top \hat{z}_j/\tau)}{D_i}.
\end{equation}

Since $w_i$ is detached from the computational graph, it is treated as a constant during backpropagation. The gradient with respect to the anchor feature is therefore
\begin{equation}
\begin{aligned}
    \frac{\partial \mathcal{L}_{g}^{(i)}}{\partial z_i}
    &=
    \frac{w_i}{\tau}
    \Bigg[
        (p_i^{+}-1)\hat{z}_i
        +
        \sum_{j\neq i}p_{ij}z_j
        +
        \sum_{j\neq i}\hat{p}_{ij}\hat{z}_j
    \Bigg].
\end{aligned}
\end{equation}

Let the gradient of the standard unweighted contrastive loss be
\begin{equation}
\begin{aligned}
    g_i
    =
    \frac{1}{\tau}
    \Bigg[
        (p_i^{+}-1)\hat{z}_i
        +
        \sum_{j\neq i}p_{ij}z_j
        +
        \sum_{j\neq i}\hat{p}_{ij}\hat{z}_j
    \Bigg].
\end{aligned}
\end{equation}
Then the proposed weighted loss satisfies
\begin{equation}
    \frac{\partial \mathcal{L}_{g}^{(i)}}{\partial z_i}
    =
    w_i g_i
    =
    (1+c_i)g_i.
\end{equation}

Consequently,
\begin{equation}
    \left\|
    \frac{\partial \mathcal{L}_{g}^{(i)}}{\partial z_i}
    \right\|_2
    =
    (1+c_i)\|g_i\|_2.
\end{equation}
Thus, compared with the standard contrastive objective, the proposed confidence term scales the gradient by exactly $1+c_i$. Since
\begin{equation}
    0 \leq 1+c_i \leq 2,
\end{equation}
pairs with low cosine similarity receive a smaller update, whereas more confident positive pairs receive a larger contribution to representation learning.

The same behavior can be seen directly from the positive-pair component. The gradient with respect to $\hat{z}_i$ contributed by the positive term is
\begin{equation}
    \frac{\partial \mathcal{L}_{g}^{(i)}}{\partial \hat{z}_i}
    =
    -\frac{w_i}{\tau}(1-p_i^{+})z_i.
\end{equation}
Hence, a gradient-descent update with learning rate $\eta$ gives
\begin{equation}
    \Delta \hat{z}_i^{+}
    =
    \frac{\eta}{\tau}
    (1+c_i)(1-p_i^{+})z_i.
\end{equation}
Therefore, $1+c_i$ acts as an explicit confidence multiplier on the attraction between the positive pair. Similarly, the negative terms are scaled by the same confidence factor:
\begin{equation}
    \Delta z_i^{-}
    =
    -\frac{\eta(1+c_i)}{\tau}
    \left(
        \sum_{j\neq i}p_{ij}z_j
        +
        \sum_{j\neq i}\hat{p}_{ij}\hat{z}_j
    \right).
\end{equation}

The stop-gradient weighting controls the gradient magnitude defined by the contrastive objective. For two samples $a$ and $b$ with
\begin{equation}
    -1 \leq c_b < c_a \leq 1,
\end{equation}
their confidence multipliers satisfy
\begin{equation}
    1+c_b < 1+c_a.
\end{equation}
Thus, relative to their corresponding unweighted contrastive gradients, sample $a$ receives a stronger update than sample $b$. In particular, when $c_i\rightarrow -1$, the contribution of an unreliable pair approaches zero, whereas when $c_i\rightarrow 1$, its contribution approaches twice that of the standard contrastive objective.

This behavior is desirable for self-refining grouping. During early training, potentially incorrect pseudo-group assignments are expected to produce less consistent positive features and therefore receive smaller weights. As the representation and pseudo groups improve, positive similarities increase and these reliable pairs contribute more strongly to contrastive learning. The weighting consequently reduces the influence of uncertain assignments while preserving the attraction-repulsion behavior of the original contrastive objective.

The same analysis directly applies to the patch-wise correspondence loss $\mathcal{L}_p$, where the confidence is defined from the cosine similarity between a patch and its selected positive correspondence. Therefore, uncertain local correspondences are similarly updated more conservatively, while reliable correspondences receive stronger supervision.

%% file: sec/8_more.tex
\section{Additional results}
\label{sec:more_results}
In this section, we provide additional qualitative results to further illustrate the behavior of our model.

\input{figures/supp/Guo}

\input{figures/ins}

In \cref{fig:demo_Guo} we illustrate the qualitative comparison of our method with Guo \etal~\cite{guo2023boundary} on the ISTD dataset using images extracted from their original paper. In \cref{fig:ins} we show additional qualitative results of \acs{name} on the INS dataset~\cite{xu2025omnisr}. The INS dataset is a synthetic benchmark designed for large-scale shadow removal evaluation, while WSRD+ originates from the NTIRE 2024 Shadow Removal Challenge and contains diverse real-world scenes with complex illumination conditions. 

\begin{figure}[!ht]
    \centering
    \includegraphics[width=0.9\linewidth]{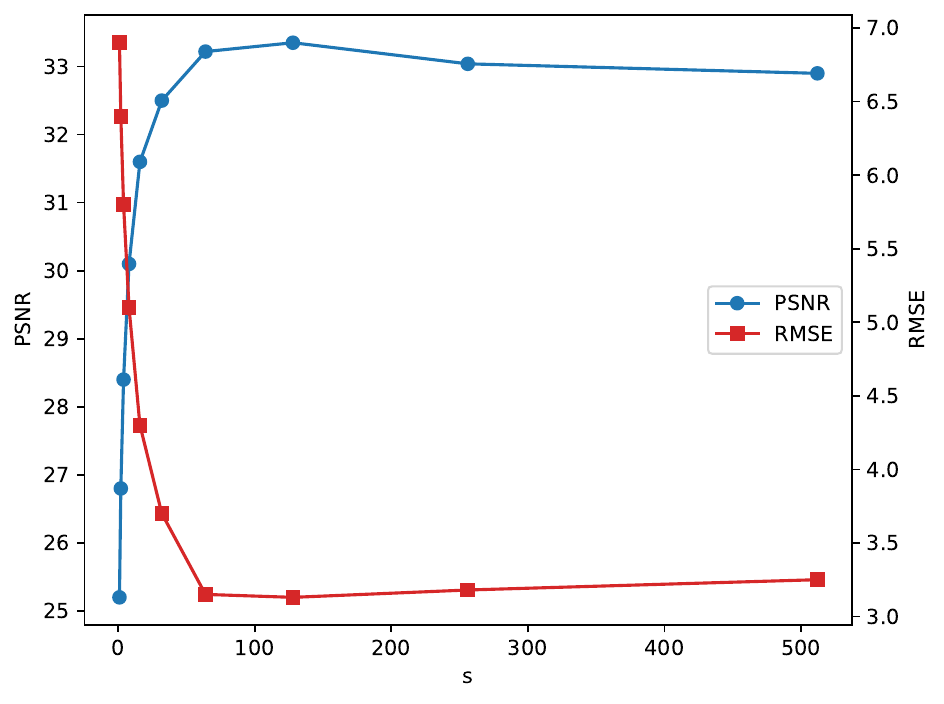}
    \caption{PSNR and RMSE of \acs{name} with varying $s$.}
    \label{fig:param_s}
\end{figure}

As discussed in \cref{sec:pairwise}, the hyperparameter $s$ controls the strength of the sigmoid gating and directly affects the degree of brightness adjustment. Since an overly small or large $s$ may lead to under-correction or over-enhancement, we analyze its impact in \cref{fig:param_s}. The results show that the performance varies moderately across a wide range of $s$, with stable PSNR and RMSE trends, indicating that the proposed framework is not overly sensitive to this parameter while still benefiting from proper calibration.

\input{tables/clustering}
Since our framework relies on image grouping, we further evaluate its sensitivity to the clustering strategy in \Cref{tab:clustering}. We replace Affinity Propagation with KMeans~\cite{lloyd1982least}, using settings from a minimum of $50$ to a maximum of $500$ clusters, and DBSCAN~\cite{ester1996density}, with the minimum number of samples ranging from $1$ to $8$. The results show that grouping quality affects the output image quality, but overall the method remains applicable. KMeans and DBSCAN also depend on prior knowledge of the group structure through their clustering parameters, whereas AP does not require such information and is therefore more reliable when the group structure is unknown. For large-scale datasets, computing all pairwise similarities may become expensive. Following~\cite{duong2026collision} for large-scale settings, clustering can instead be fitted on a small randomly sampled subset of the training set, and the resulting cluster model can then be used to assign the remaining training samples. To evaluate our method under this setting, we fit AP using only $1\%$, $5\%$, $10\%$, and $20\%$ of the training data and assign the remaining images using the learned exemplars. Using 10--20\% of the training data yields performance close to that obtained with full-data AP, whereas using smaller subsets degrades performance as expected due to the very limited number of exemplars.

\input{tables/complexity}
\input{tables/more_ablation}
\Cref{tab:ablation_ab} further analyzes the reconstruction loss $\mathcal{L}_{\text{r}} = \| U'(S(x_i)) - \hat{x}_i \|_1 + a\| U'(S(x_i)) - x_i \|_1 + b\| U'(S(x_i)) - U'(S(\hat{x}_i)) \|_1$. The results show that using only the pseudo-target reconstruction is insufficient. The effect of $a$ and $b$ also depends on the dataset. On AISTD, images within the same scene are more consistent, so a stronger group-consistency term is more helpful. On SRD, illumination and target appearances vary more across grouped images, making the self-reconstruction term relatively more important for preserving image-specific content. We also compare computational efficiency with prior methods in \cref{tab:complexity}. Our standard model requires $0.05$ TFLOPs per image, approximately 18 images/s on a single NVIDIA H100 GPU. The iterative refinement variant requires about $12$ TFLOPs.

To further analyze the effect of shadow generation, we revisit the ablation in \Cref{tab:ablation}. When similar images are available, using or removing the shadow generator has only a limited effect, and in some metrics removing it even gives slightly better results. This can be explained by the fact that shadows directly observed in other images of the same group are more realistic than synthetically generated ones. However, the role of the shadow generator becomes clearer when similar images are unavailable. We repeat the one-image-per-group setting without shadow generation and obtain only $26.94$ PSNR and $0.915$ SSIM, compared with $30.62$ PSNR and $0.951$ SSIM when shadow generation is used. This shows that the shadow generator is an important component when training data do not provide similar observations. Interestingly, the one-image-per-group result is also close to the cross-dataset evaluation where the model is trained on SRD and tested on AISTD ($30.71$ PSNR and $0.949$ SSIM). In both cases, the shadows used during training follow a distribution different from the shadows in AISTD, which may explain why their performance is lower than when the model is trained directly on AISTD.

\input{figures/example_SRD_group}
\Cref{fig:group_srd} shows randomly selected groups from SRD and INS. The SRD groups mostly contain visually related real images with overlapping scene regions, while INS is more challenging because it is synthetically generated, contains fewer images per scene, and exhibits larger inter-image variations with often small self-shadows.

As shown in \cref{fig:reflection}, our method removes shadows effectively while preserving fine reflections and illumination cues, in some cases even producing reflection consistency superior to the ground truth. This demonstrates that the model fulfills its intended objective by focusing solely on shadow removal.

In \cref{fig:fig1,fig:fig2,fig:fig3,fig:video2,fig:video3,fig:fig6}, we present additional qualitative results across all datasets used in this paper, covering both shadow removal and shadow segmentation. These examples show that our method performs robustly across diverse scenes, often outperforming several supervised approaches and, in certain cases, even producing results that are visually more consistent than the ground truth.

\clearpage

\input{figures/supp/reflection}

\input{figures/supp/fig1}
\input{figures/supp/fig2}
\input{figures/supp/fig3}
\input{figures/supp/fig4}
\input{figures/supp/fig5}
\input{figures/supp/fig6}

%% file: figures/supp/Guo.tex
\begin{figure}[ht]
\centering
\newcommand{\ncols}{5}
\setlength{\tabcolsep}{0.1em}

\newlength{\imgwidthREBone}
\setlength{\imgwidthREBone}{\dimexpr(0.999\linewidth - \ncols\tabcolsep)/\ncols\relax}

\begin{tabular}{*{\ncols}{c}}
\includegraphics[width=\imgwidthREBone]{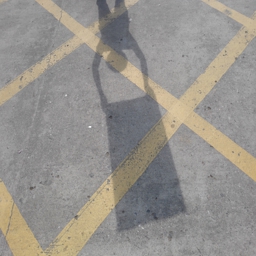} &
\includegraphics[width=\imgwidthREBone]{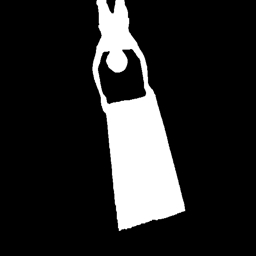} &
\includegraphics[width=\imgwidthREBone]{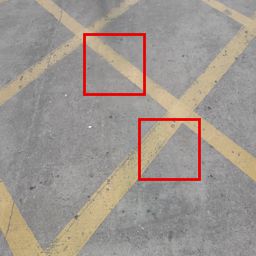} &
\includegraphics[width=\imgwidthREBone]{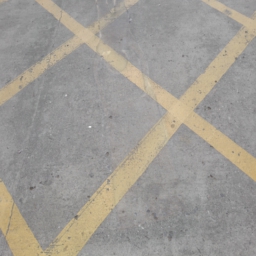} &
\includegraphics[width=\imgwidthREBone]{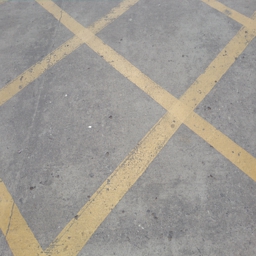} \\[-0.5ex]

\includegraphics[width=\imgwidthREBone]{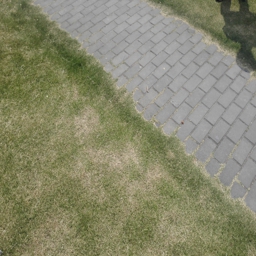} &
\includegraphics[width=\imgwidthREBone]{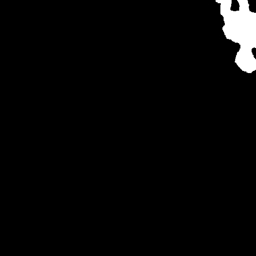} &
\includegraphics[width=\imgwidthREBone]{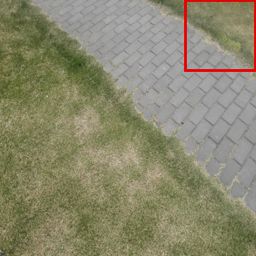} &
\includegraphics[width=\imgwidthREBone]{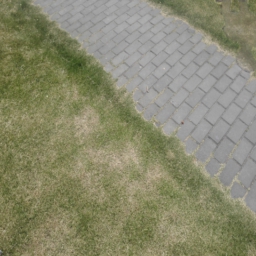} &
\includegraphics[width=\imgwidthREBone]{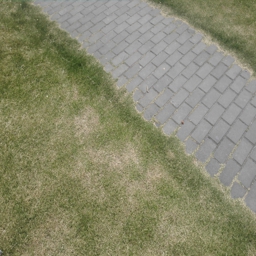}\\[-0.5ex]

\includegraphics[width=\imgwidthREBone]{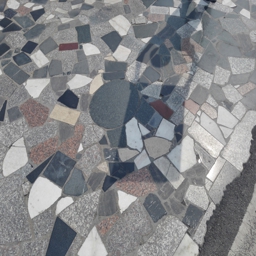} &
\includegraphics[width=\imgwidthREBone]{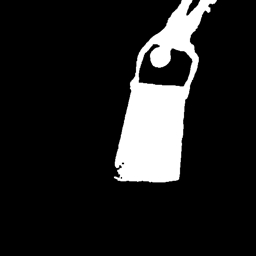} &
\includegraphics[width=\imgwidthREBone]{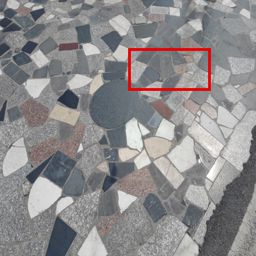} &
\includegraphics[width=\imgwidthREBone]{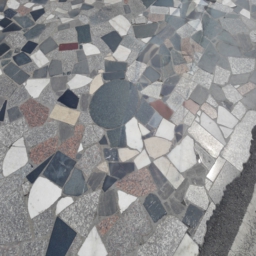} &
\includegraphics[width=\imgwidthREBone]{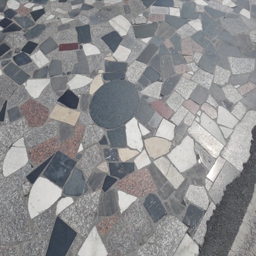} \\[-0.5ex]

\includegraphics[width=\imgwidthREBone]{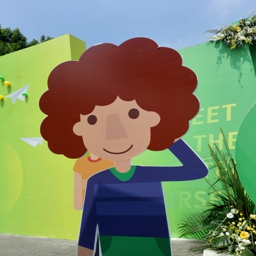} &
\includegraphics[width=\imgwidthREBone]{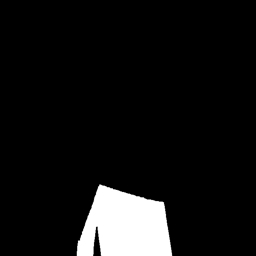} &
\includegraphics[width=\imgwidthREBone]{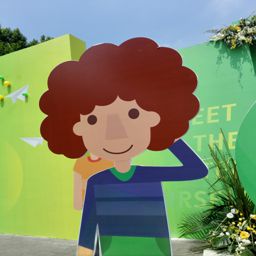} &
\includegraphics[width=\imgwidthREBone]{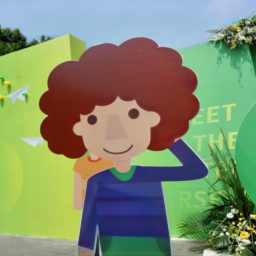} &
\includegraphics[width=\imgwidthREBone]{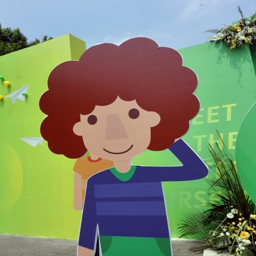}\\[-1.5ex]

\tiny (a) Input &
\tiny (b) Shadow mask &
\tiny (c) Guo \etal &
\tiny (d) Ours &
\tiny (e) GT \\
\end{tabular}

\caption{Comparison on the ISTD dataset. Images are extracted from the original paper by Guo~\etal~\cite{guo2023boundary}.}
\label{fig:demo_Guo}
\end{figure}

%% file: figures/ins.tex
\begin{figure}[h]
\centering
\newcommand{\ncols}{6}
\setlength{\tabcolsep}{0.1em}

\newlength{\imgwidthINS}
\setlength{\imgwidthINS}{\dimexpr(0.98\linewidth - \ncols\tabcolsep)/\ncols\relax}

\begin{tabular}{*{\ncols}{c}}
\centering
\includegraphics[width=\imgwidthINS]{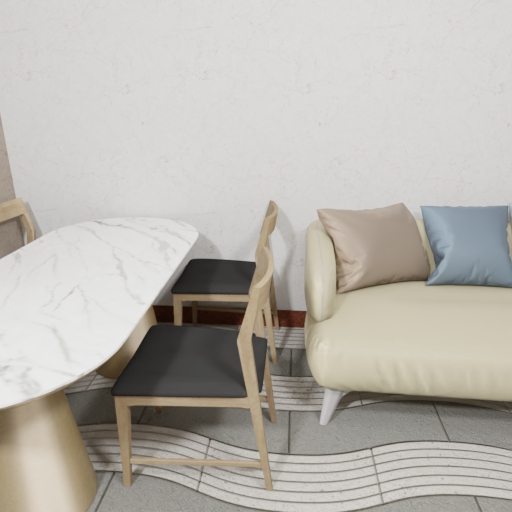} &
\includegraphics[width=\imgwidthINS]{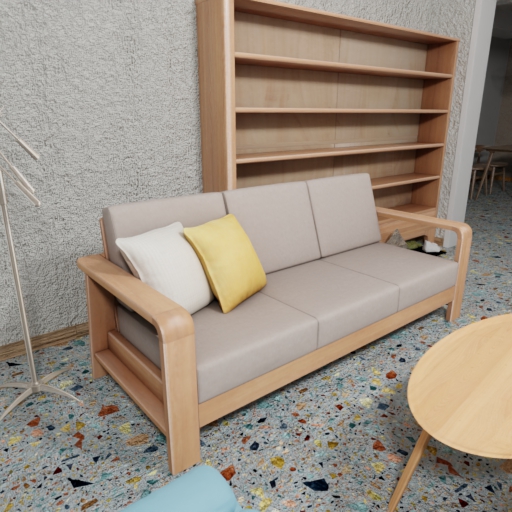} &
\includegraphics[width=\imgwidthINS]{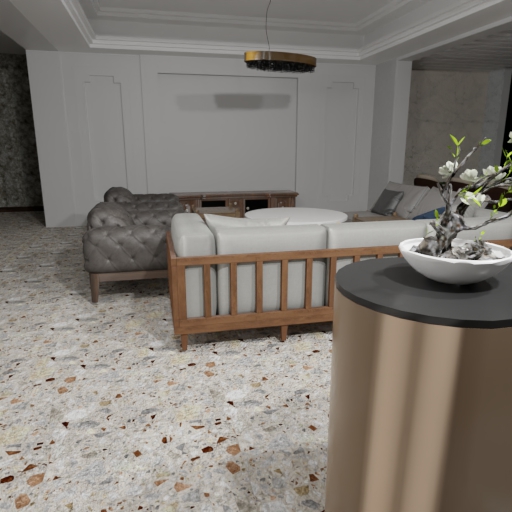} &
\includegraphics[width=\imgwidthINS]{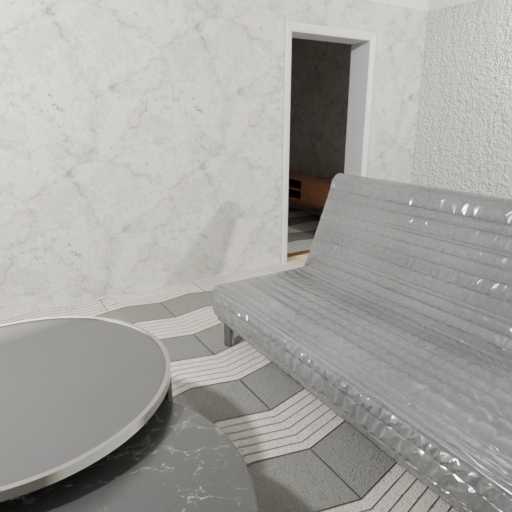} &
\includegraphics[width=\imgwidthINS]{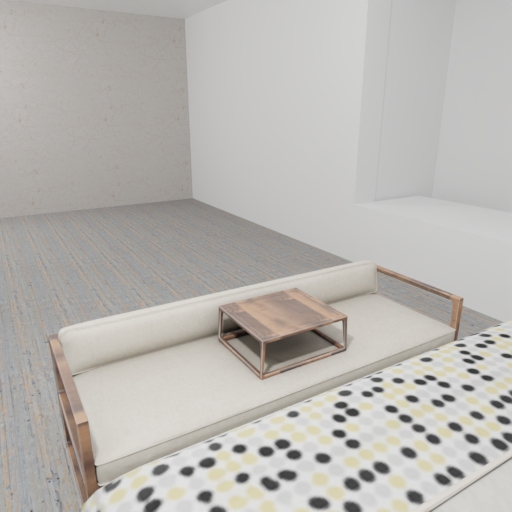} &
\includegraphics[width=\imgwidthINS]{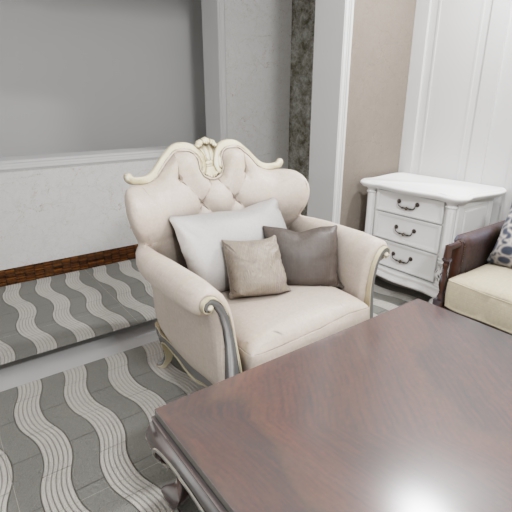} \\

\multicolumn{6}{c}{Input shadow images} \\

\includegraphics[width=\imgwidthINS]{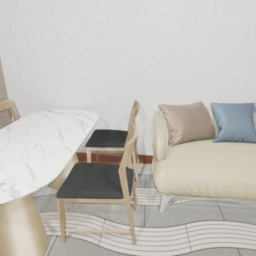} &
\includegraphics[width=\imgwidthINS]{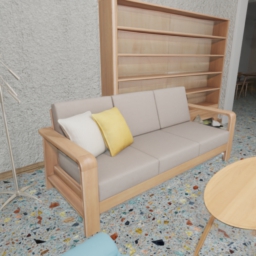} &
\includegraphics[width=\imgwidthINS]{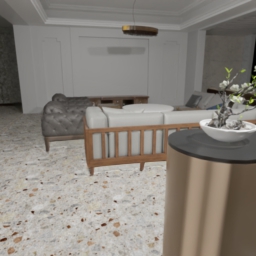} &
\includegraphics[width=\imgwidthINS]{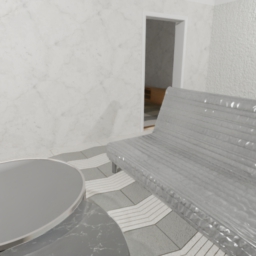} &
\includegraphics[width=\imgwidthINS]{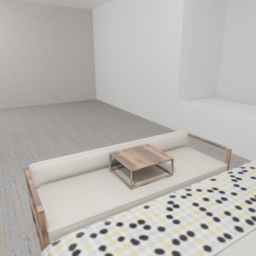} &
\includegraphics[width=\imgwidthINS]{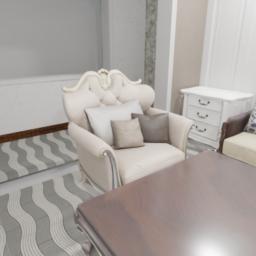} \\

\multicolumn{6}{c}{Ours output images} \\

\includegraphics[width=\imgwidthINS]{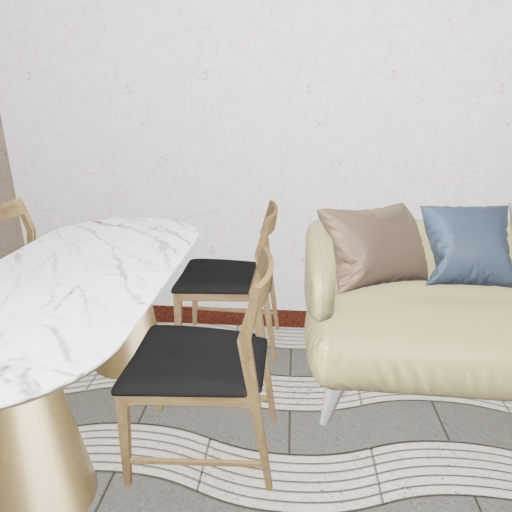} &
\includegraphics[width=\imgwidthINS]{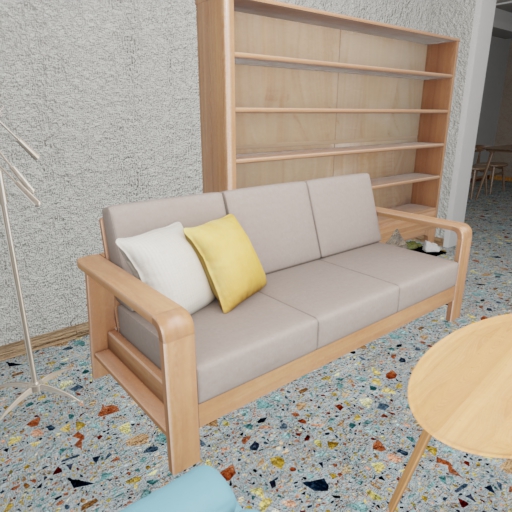} &
\includegraphics[width=\imgwidthINS]{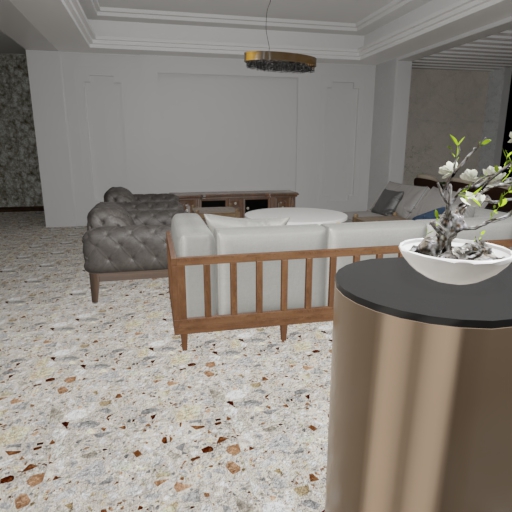} &
\includegraphics[width=\imgwidthINS]{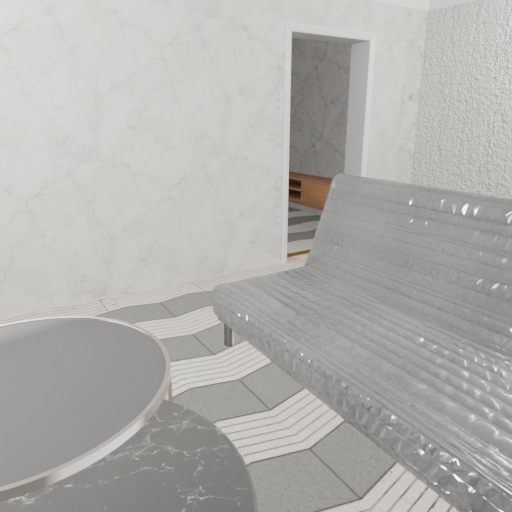} &
\includegraphics[width=\imgwidthINS]{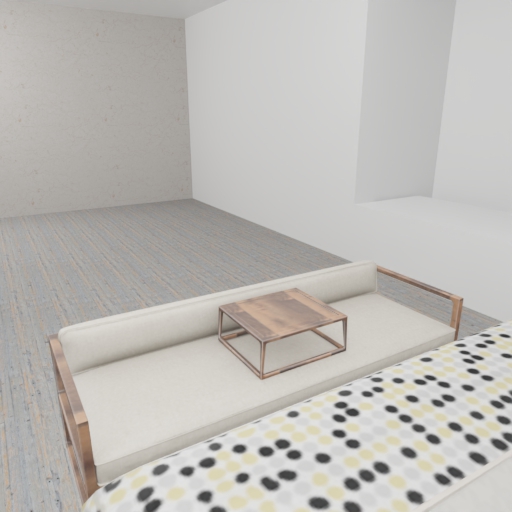} &
\includegraphics[width=\imgwidthINS]{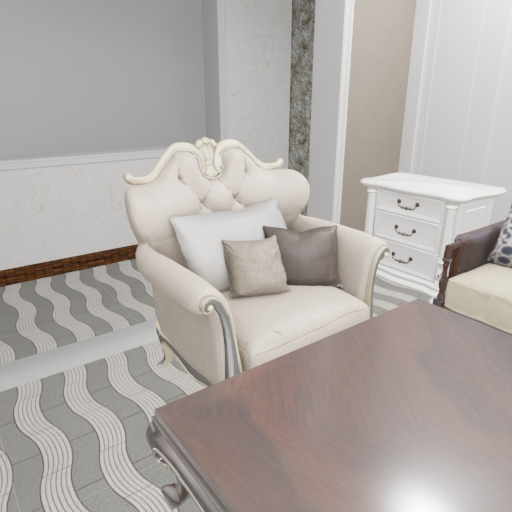} \\

\multicolumn{6}{c}{Ground truth} \\

\end{tabular}

\caption{Qualitative comparison on the INS dataset~\cite{xu2025omnisr}.}
\label{fig:ins}
\end{figure}

%% file: tables/clustering.tex
\begin{table}[ht]
\centering
\caption{Clustering ablation study on SRD and AISTD. We report PSNR and SSIM for shadow removal performance, and ARI for clustering quality. The best results are in bold, and the second best are underlined.}
\label{tab:clustering}
\resizebox{\linewidth}{!}{
\begin{tabular}{l|cc|cccc}
\hline
\multirow{2}{*}{Method} & \multicolumn{2}{c|}{SRD} & \multicolumn{4}{c}{AISTD} \\
 & PSNR & SSIM & PSNR & SSIM & ARI init & ARI end \\ \hline
Input & 18.25 & 0.837 & 20.46 & 0.894 & - & - \\ \hline
Ours KMeans (k=500) & 30.10 & 0.947 & 30.35 & 0.949 & 0.34 & 0.36 \\
Ours KMeans (k=200) & 31.38 & 0.954 & 32.42 & 0.957 & {\ul 0.64} & 0.70 \\
Ours KMeans (k=100) & 31.55 & 0.955 & 32.85 & 0.959 & \textbf{0.75} & 0.82 \\
Ours KMeans (k=50) & 29.20 & 0.946 & 32.05 & 0.956 & 0.48 & 0.65 \\
Ours DBSCAN (min\_samples=1) & \textbf{32.13} & \textbf{0.960} & 33.21 & \textbf{0.965} & 0.40 & \textbf{0.96} \\
Ours DBSCAN (min\_samples=4) & {\ul 32.07} & {\ul 0.959} & \textbf{34.24} & {\ul 0.964} & 0.40 & {\ul 0.94} \\
Ours DBSCAN (min\_samples=8) & 28.55 & 0.946 & 30.75 & 0.951 & 0.40 & 0.42 \\
Ours AP 20\% & 31.75 & {\ul 0.959} & 33.86 & \textbf{0.965} & 0.57 & 0.92 \\
Ours AP 10\% & 31.98 & {\ul 0.959} & {\ul 34.16} & {\ul 0.964} & 0.53 & 0.91 \\
Ours AP 5\% & 31.23 & 0.955 & 32.60 & 0.960 & 0.39 & 0.88 \\
Ours AP 1\% & 30.89 & 0.950 & 31.03 & 0.952 & 0.21 & 0.48 \\
Ours & 31.93 & 0.958 & 33.35 & {\ul 0.964} & 0.59 & 0.92 \\ \hline
\end{tabular}
}
\end{table}

%% file: tables/complexity.tex
\begin{table}[ht]
\centering
\caption{Computational efficiency comparison of the proposed method with prior unsupervised shadow removal methods.}
\label{tab:complexity}
\begin{tabular}{l|cc}
\hline
Methods & \multicolumn{1}{l}{TFLOPS} & \multicolumn{1}{l}{Params (M)} \\ \hline
Mask-ShadowGAN & 0.05 & 22.8 \\
LG-ShadowNet \cite{liu2021shadow} & 0.03 & 5.7 \\
DC-ShadowNet \cite{jin2021dc} & 0.05 & 10.6 \\
G2R & 0.11 & 22.8 \\
Guo \etal & 61 & 113.7 \\
\acs{name} (Ours) & 0.05 & 11.4 \\
\acs{name} (Ours) iterative & 12 & 11.4 \\ \hline
\end{tabular}
\end{table}

%% file: tables/more_ablation.tex
\begin{table}[ht]
\caption{Ablation study of the reconstruction loss $\mathcal{L}_{\text{r}} = \| U'(S(x_i)) - \hat{x}_i \|_1 + a\| U'(S(x_i)) - x_i \|_1 + b\| U'(S(x_i)) - U'(S(\hat{x}_i)) \|_1$ with different values of $a$ and $b$.} 
\label{tab:ablation_ab}
\resizebox{\linewidth}{!}{
\begin{tabular}{c|cc|cc}
\hline
\multirow{2}{*}{Method} & \multicolumn{2}{c|}{AISTD} & \multicolumn{2}{c}{SRD} \\
 & PSNR & SSIM & PSNR & SSIM \\ \hline
Ours ($a=0,b=0$) & 32.98 & 0.960 & 29.12 & 0.914 \\
Ours ($a=1,b=0$) & 31.15 & 0.954 & 29.88 & 0.933 \\
Ours ($a=0,b=1$) & 33.06 & 0.963 & 30.06 & 0.920 \\
Ours ($a=0,b=2$) & 32.82 & 0.964 & 30.23 & 0.915 \\
Ours ($a=0.5,b=1$) & 33.87 & 0.972 & 31.28 & 0.954 \\
Ours ($a=0.5,b=2$) & 34.18 & 0.971 & 30.78 & 0.950 \\
Ours ($a=1,b=2$) & 34.08 & 0.972 & 31.01 & 0.955 \\
Ours ($a=1,b=1$) & 33.35 & 0.964 & 31.93 & 0.958 \\ \hline
\end{tabular}
}
\end{table}

%% file: figures/example_SRD_group.tex
\begin{figure}[ht]
\centering
\newcommand{\ncols}{10}
\pgfmathtruncatemacro{\halfcols}{\ncols/2}
\setlength{\tabcolsep}{0.1em}

\newlength{\imgwidthGroupSRD}
\setlength{\imgwidthGroupSRD}{\dimexpr(0.99\linewidth - \ncols\tabcolsep)/\ncols\relax}

\begin{tabular}{*{\halfcols}{c}|*{\halfcols}{c}}

\multicolumn{\halfcols}{c|}{\tiny{SRD group 0}} &
\multicolumn{\halfcols}{c}{\tiny{INS group 0}} \\

\includegraphics[width=\imgwidthGroupSRD]{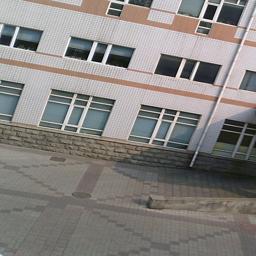} &
\includegraphics[width=\imgwidthGroupSRD]{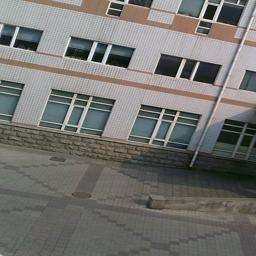} &
\includegraphics[width=\imgwidthGroupSRD]{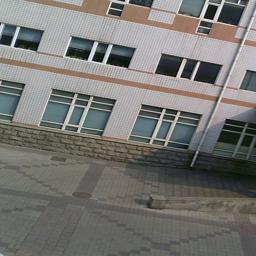} & 
& & 
\includegraphics[width=\imgwidthGroupSRD]{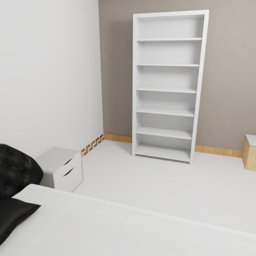} &
\includegraphics[width=\imgwidthGroupSRD]{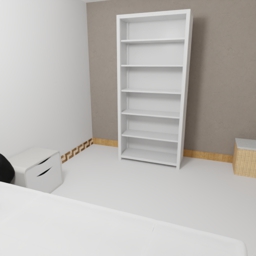} & & \\[-1.5ex]

\multicolumn{\halfcols}{c|}{\tiny{SRD group 1}} &
\multicolumn{\halfcols}{c}{\tiny{INS group 1}} \\

\includegraphics[width=\imgwidthGroupSRD]{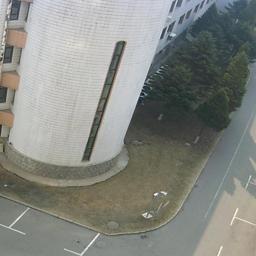} &
\includegraphics[width=\imgwidthGroupSRD]{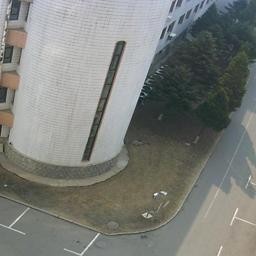} &
\includegraphics[width=\imgwidthGroupSRD]{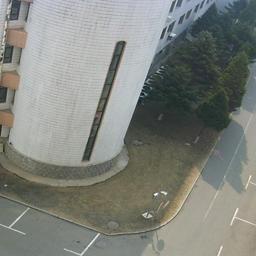} &
\includegraphics[width=\imgwidthGroupSRD]{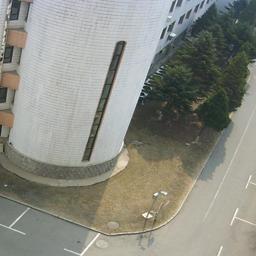} &
\includegraphics[width=\imgwidthGroupSRD]{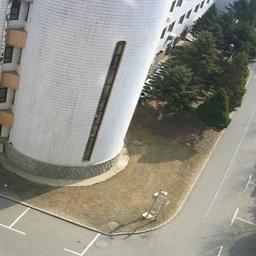} &
\includegraphics[width=\imgwidthGroupSRD]{figures/images/instrainresize/10009.jpg} &
\includegraphics[width=\imgwidthGroupSRD]{figures/images/instrainresize/9970.jpg} &
\includegraphics[width=\imgwidthGroupSRD]{figures/images/instrainresize/9977.jpg} &
\includegraphics[width=\imgwidthGroupSRD]{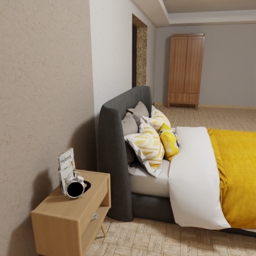} &
 \\[-1.5ex]

\multicolumn{\halfcols}{c|}{\tiny{SRD group 2}} &
\multicolumn{\halfcols}{c}{\tiny{INS group 2}} \\

\includegraphics[width=\imgwidthGroupSRD]{figures/images/srdtrainresize/DSCF0585.jpg} &
\includegraphics[width=\imgwidthGroupSRD]{figures/images/srdtrainresize/DSCF0647.jpg} &
\includegraphics[width=\imgwidthGroupSRD]{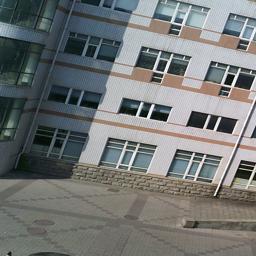} & 
\includegraphics[width=\imgwidthGroupSRD]{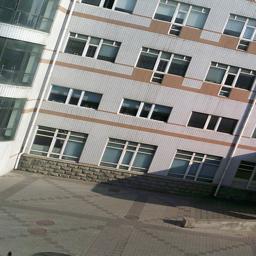} &
\includegraphics[width=\imgwidthGroupSRD]{figures/images/srdtrainresize/DSCF0573.jpg} & 
\includegraphics[width=\imgwidthGroupSRD]{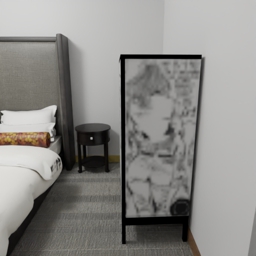} & 
\includegraphics[width=\imgwidthGroupSRD]{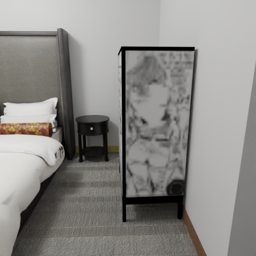} & & &  \\[-1.5ex]

\multicolumn{\halfcols}{c|}{\tiny{SRD group 3}} &
\multicolumn{\halfcols}{c}{\tiny{INS group 3}} \\

\includegraphics[width=\imgwidthGroupSRD]{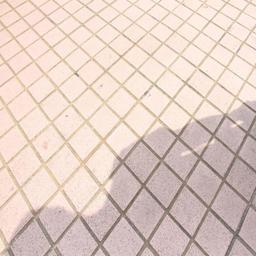} &
\includegraphics[width=\imgwidthGroupSRD]{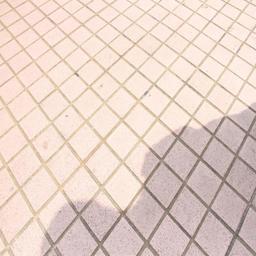} &
\includegraphics[width=\imgwidthGroupSRD]{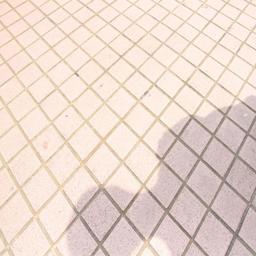} &
\includegraphics[width=\imgwidthGroupSRD]{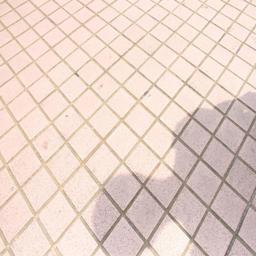} &
\includegraphics[width=\imgwidthGroupSRD]{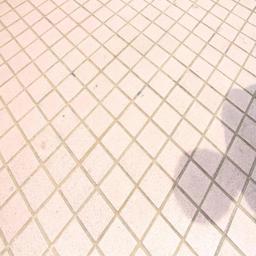} & 
\includegraphics[width=\imgwidthGroupSRD]{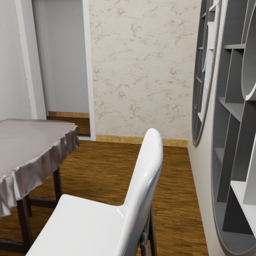} &  
\includegraphics[width=\imgwidthGroupSRD]{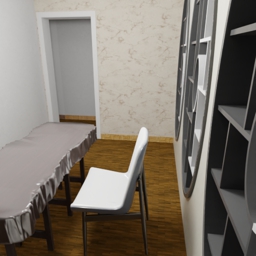} &  & \\[-1.5ex]

\end{tabular}

\caption{Random groups from SRD and INS, with up to five images per group where available.}
\label{fig:group_srd}
\end{figure}

%% file: figures/supp/reflection.tex
\begin{figure*}[ht]
\centering
\newcommand{\ncols}{6}
\setlength{\tabcolsep}{0.1em}

\newlength{\imgwidthreflection}
\setlength{\imgwidthreflection}{\dimexpr(0.95\linewidth - \ncols\tabcolsep)/\ncols\relax}

\begin{tabular}{*{\ncols}{c}}
\centering
\includegraphics[width=\imgwidthreflection]{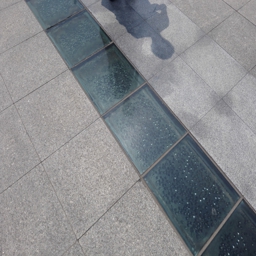} &
\includegraphics[width=\imgwidthreflection]{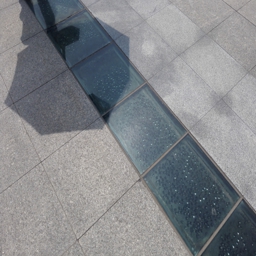} &
\includegraphics[width=\imgwidthreflection]{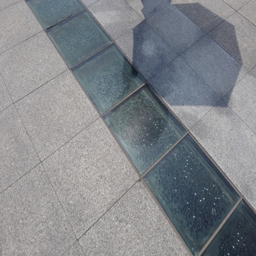} &
\includegraphics[width=\imgwidthreflection]{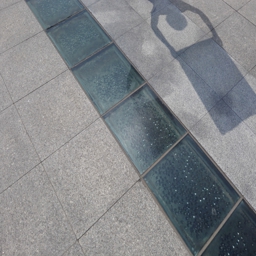} &
\includegraphics[width=\imgwidthreflection]{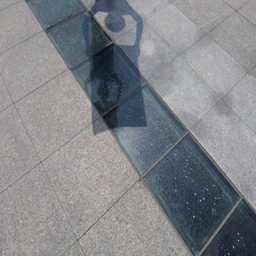} &
\includegraphics[width=\imgwidthreflection]{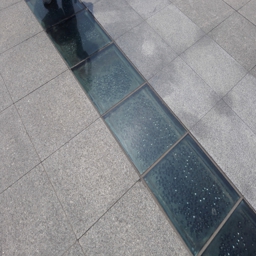} \\

\multicolumn{6}{c}{(a) Input shadow images} \\

\includegraphics[width=\imgwidthreflection]{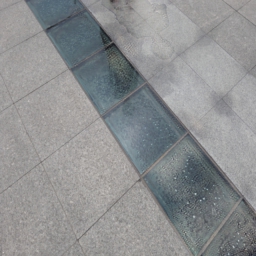} &
\includegraphics[width=\imgwidthreflection]{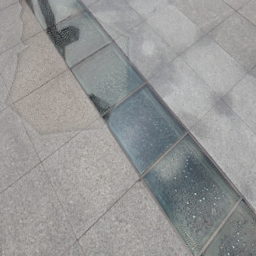} &
\includegraphics[width=\imgwidthreflection]{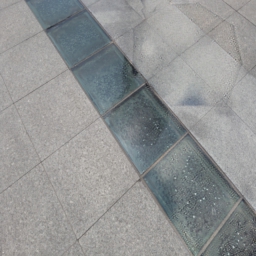} &
\includegraphics[width=\imgwidthreflection]{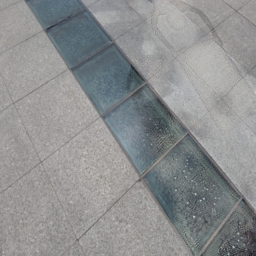} &
\includegraphics[width=\imgwidthreflection]{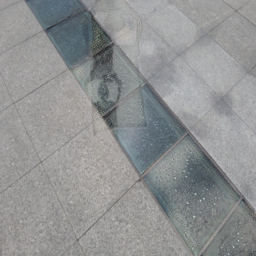} &
\includegraphics[width=\imgwidthreflection]{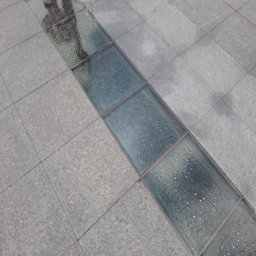} \\

\multicolumn{6}{c}{(b) DC-ShadowNet} \\

\includegraphics[width=\imgwidthreflection]{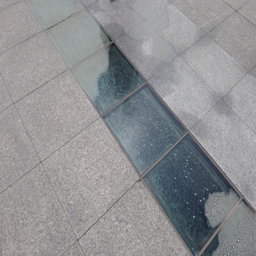} &
\includegraphics[width=\imgwidthreflection]{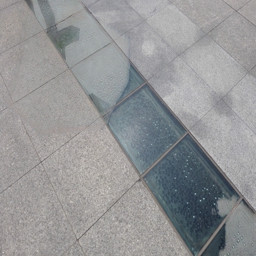} &
\includegraphics[width=\imgwidthreflection]{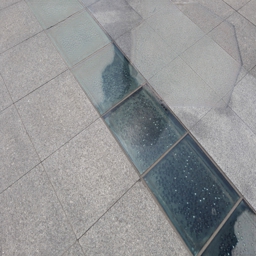} &
\includegraphics[width=\imgwidthreflection]{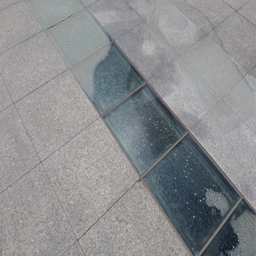} &
\includegraphics[width=\imgwidthreflection]{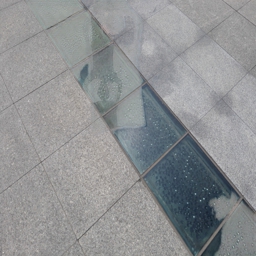} &
\includegraphics[width=\imgwidthreflection]{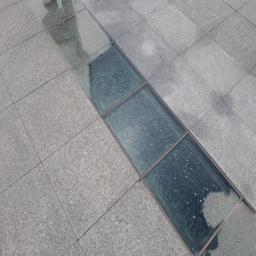} \\

\multicolumn{6}{c}{(c) G2R} \\

\includegraphics[width=\imgwidthreflection]{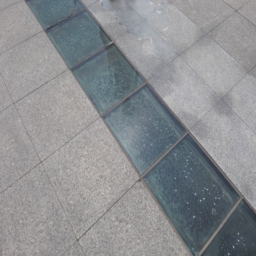} &
\includegraphics[width=\imgwidthreflection]{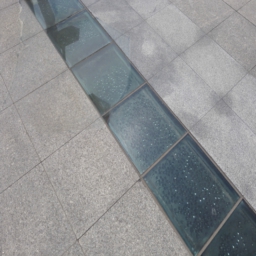} &
\includegraphics[width=\imgwidthreflection]{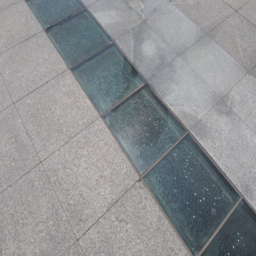} &
\includegraphics[width=\imgwidthreflection]{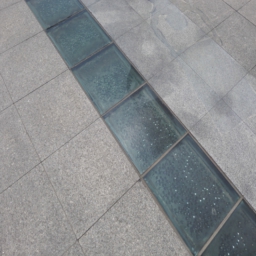} &
\includegraphics[width=\imgwidthreflection]{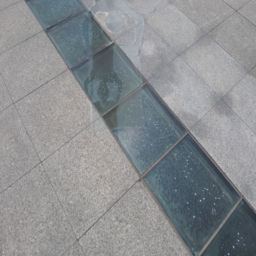} &
\includegraphics[width=\imgwidthreflection]{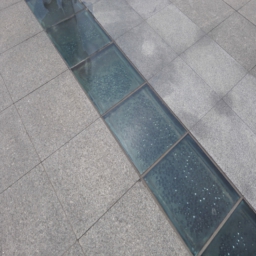} \\

\multicolumn{6}{c}{(d) Ours output images} \\

\includegraphics[width=\imgwidthreflection]{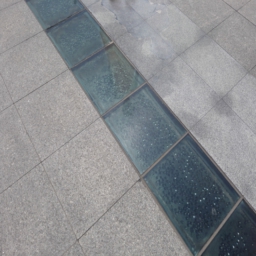} &
\includegraphics[width=\imgwidthreflection]{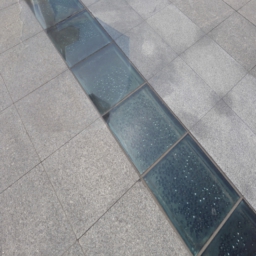} &
\includegraphics[width=\imgwidthreflection]{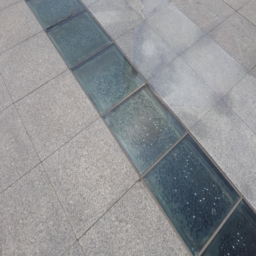} &
\includegraphics[width=\imgwidthreflection]{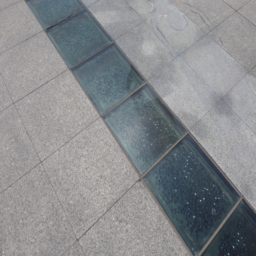} &
\includegraphics[width=\imgwidthreflection]{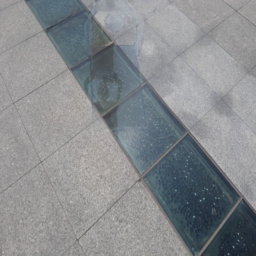} &
\includegraphics[width=\imgwidthreflection]{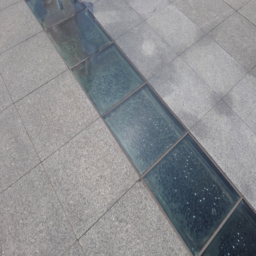} \\

\multicolumn{6}{c}{(e) Ours iterative output images} \\

\includegraphics[width=\imgwidthreflection]{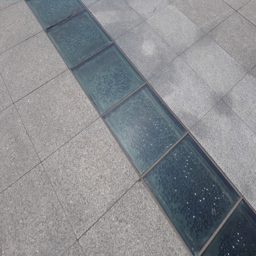} &
\includegraphics[width=\imgwidthreflection]{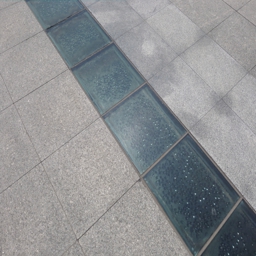} &
\includegraphics[width=\imgwidthreflection]{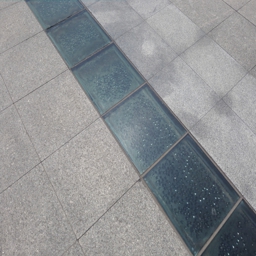} &
\includegraphics[width=\imgwidthreflection]{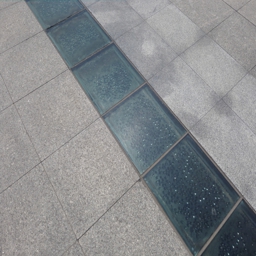} &
\includegraphics[width=\imgwidthreflection]{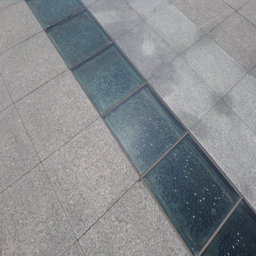} &
\includegraphics[width=\imgwidthreflection]{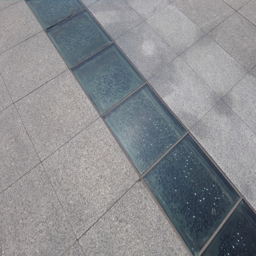} \\

\multicolumn{6}{c}{(f) Ground truth} \\

\end{tabular}

\caption{Qualitative results on the AISTD dataset \cite{le2019shadow, wang2018stacked}. Our model removes shadows while preserving fine reflections and illumination details, and in several cases achieves better reflection fidelity than the ground truth.}
\label{fig:reflection}
\end{figure*}

%% file: figures/supp/fig1.tex
\begin{figure*}[ht]
\centering
\newcommand{\ncols}{6}
\setlength{\tabcolsep}{0.1em}

\newlength{\imgwidthsixcols}
\setlength{\imgwidthsixcols}{\dimexpr(\textwidth - \ncols\tabcolsep)/\ncols\relax}

\begin{tabular}{*{\ncols}{c}}
\centering
\includegraphics[width=\imgwidthsixcols]{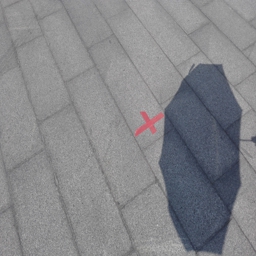} &
\includegraphics[width=\imgwidthsixcols]{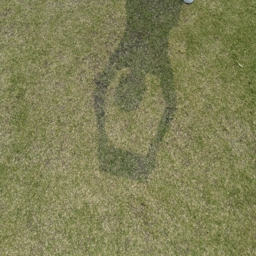} &
\includegraphics[width=\imgwidthsixcols]{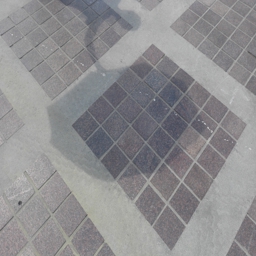} &
\includegraphics[width=\imgwidthsixcols]{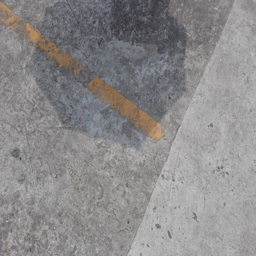} &
\includegraphics[width=\imgwidthsixcols]{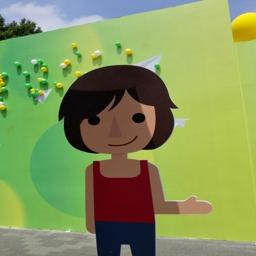} &
\includegraphics[width=\imgwidthsixcols]{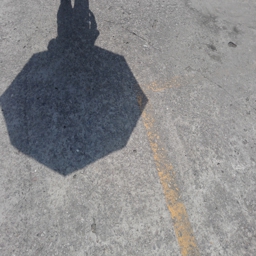} \\

\multicolumn{6}{c}{(a) Input shadow images} \\

\includegraphics[width=\imgwidthsixcols]{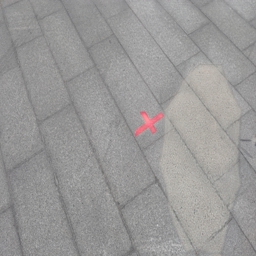} &
\includegraphics[width=\imgwidthsixcols]{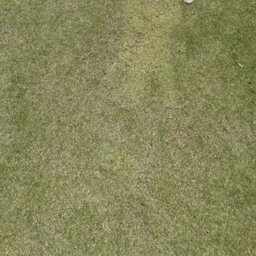} &
\includegraphics[width=\imgwidthsixcols]{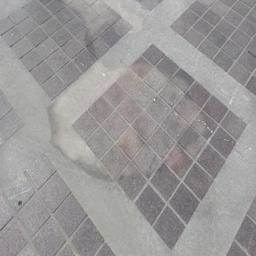} &
\includegraphics[width=\imgwidthsixcols]{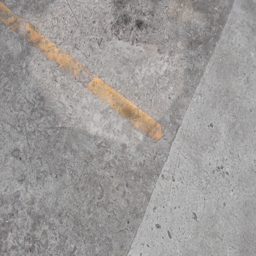} &
\includegraphics[width=\imgwidthsixcols]{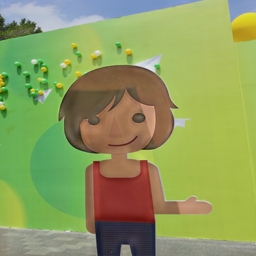} &
\includegraphics[width=\imgwidthsixcols]{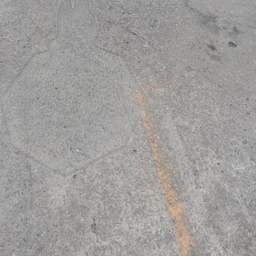} \\

\multicolumn{6}{c}{(b) LG-ShadowNet} \\

\includegraphics[width=\imgwidthsixcols]{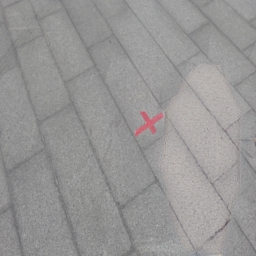} &
\includegraphics[width=\imgwidthsixcols]{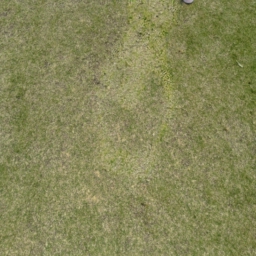} &
\includegraphics[width=\imgwidthsixcols]{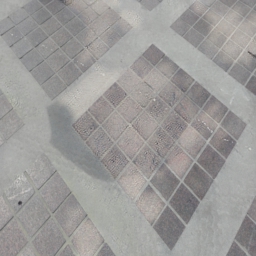} &
\includegraphics[width=\imgwidthsixcols]{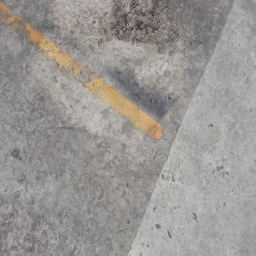} &
\includegraphics[width=\imgwidthsixcols]{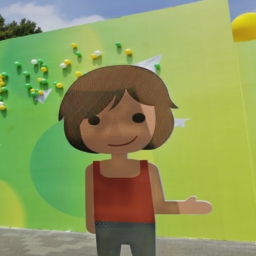} &
\includegraphics[width=\imgwidthsixcols]{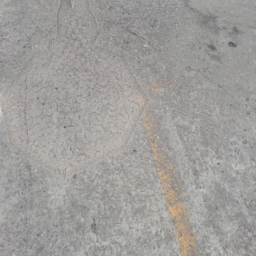} \\

\multicolumn{6}{c}{(c) DC-ShadowNet} \\

\includegraphics[width=\imgwidthsixcols]{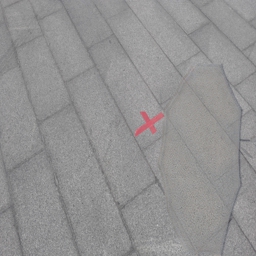} &
\includegraphics[width=\imgwidthsixcols]{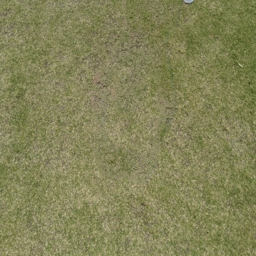} &
\includegraphics[width=\imgwidthsixcols]{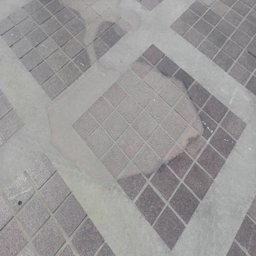} &
\includegraphics[width=\imgwidthsixcols]{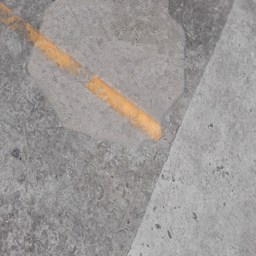} &
\includegraphics[width=\imgwidthsixcols]{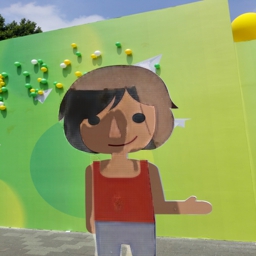} &
\includegraphics[width=\imgwidthsixcols]{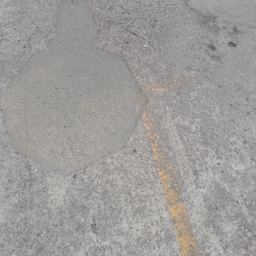} \\

\multicolumn{6}{c}{(d) G2R} \\

\includegraphics[width=\imgwidthsixcols]{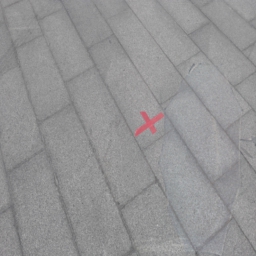} &
\includegraphics[width=\imgwidthsixcols]{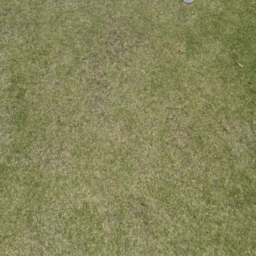} &
\includegraphics[width=\imgwidthsixcols]{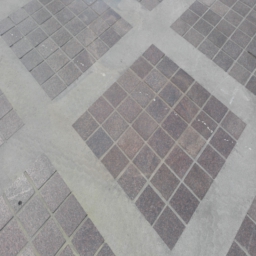} &
\includegraphics[width=\imgwidthsixcols]{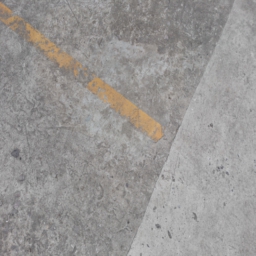} &
\includegraphics[width=\imgwidthsixcols]{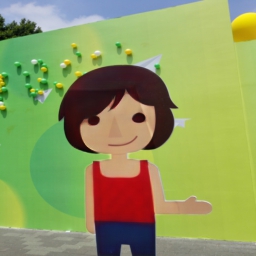} &
\includegraphics[width=\imgwidthsixcols]{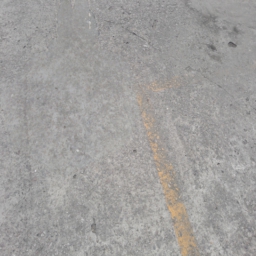} \\

\multicolumn{6}{c}{(e) Ours output images} \\

\includegraphics[width=\imgwidthsixcols]{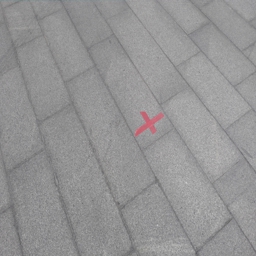} &
\includegraphics[width=\imgwidthsixcols]{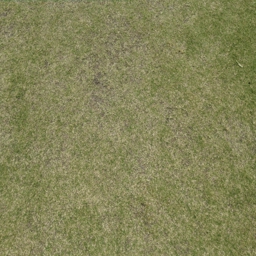} &
\includegraphics[width=\imgwidthsixcols]{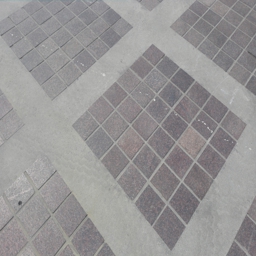} &
\includegraphics[width=\imgwidthsixcols]{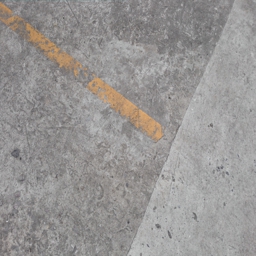} &
\includegraphics[width=\imgwidthsixcols]{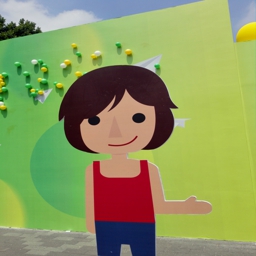} &
\includegraphics[width=\imgwidthsixcols]{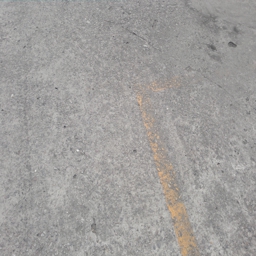} \\

\multicolumn{6}{c}{(f) Ground truth} \\

\end{tabular}

\caption{Qualitative comparison on the AISTD dataset \cite{le2019shadow, wang2018stacked}.}
\label{fig:fig1}
\end{figure*}

%% file: figures/supp/fig2.tex
\begin{figure*}[ht]
\centering
\newcommand{\ncols}{6}
\setlength{\tabcolsep}{0.1em}

\newlength{\imgwidthfigtwo}
\setlength{\imgwidthfigtwo}{\dimexpr(\textwidth - \ncols\tabcolsep)/\ncols\relax}

\begin{tabular}{*{\ncols}{c}}
\centering
\includegraphics[width=\imgwidthfigtwo]{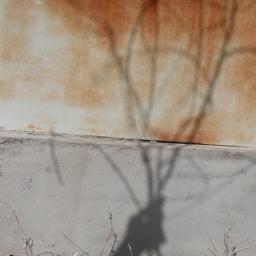} &
\includegraphics[width=\imgwidthfigtwo]{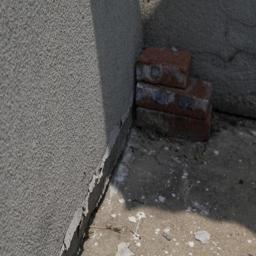} &
\includegraphics[width=\imgwidthfigtwo]{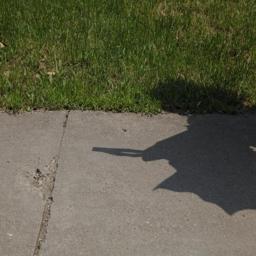} &
\includegraphics[width=\imgwidthfigtwo]{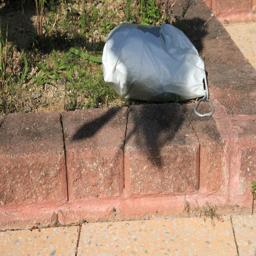} &
\includegraphics[width=\imgwidthfigtwo]{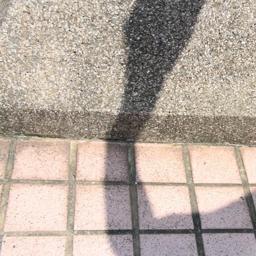} &
\includegraphics[width=\imgwidthfigtwo]{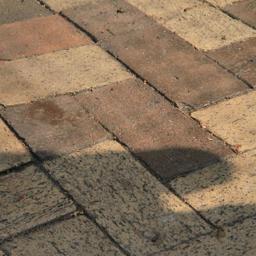} \\

\multicolumn{6}{c}{(a) Input shadow images} \\

\includegraphics[width=\imgwidthfigtwo]{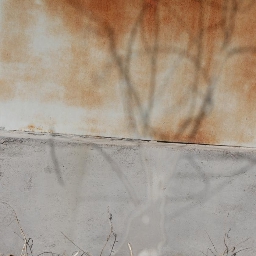} &
\includegraphics[width=\imgwidthfigtwo]{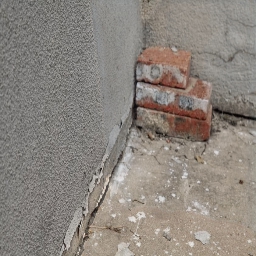} &
\includegraphics[width=\imgwidthfigtwo]{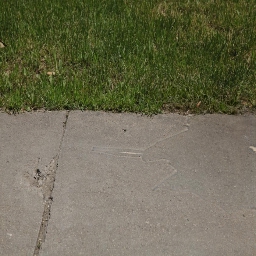} &
\includegraphics[width=\imgwidthfigtwo]{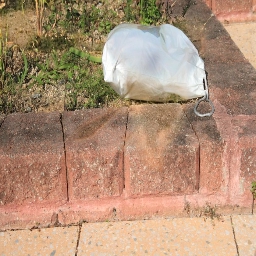} &
\includegraphics[width=\imgwidthfigtwo]{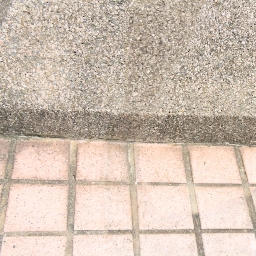} &
\includegraphics[width=\imgwidthfigtwo]{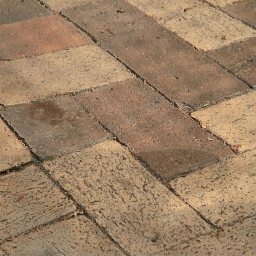}  \\

\multicolumn{6}{c}{(b) BMNet} \\

\includegraphics[width=\imgwidthfigtwo]{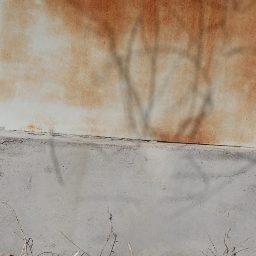} &
\includegraphics[width=\imgwidthfigtwo]{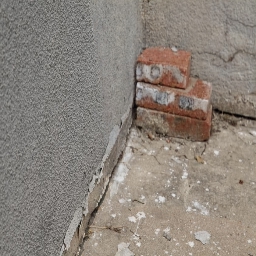} &
\includegraphics[width=\imgwidthfigtwo]{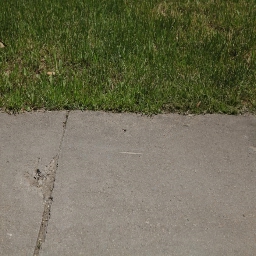} &
\includegraphics[width=\imgwidthfigtwo]{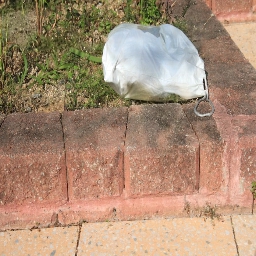} &
\includegraphics[width=\imgwidthfigtwo]{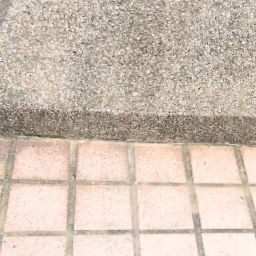} &
\includegraphics[width=\imgwidthfigtwo]{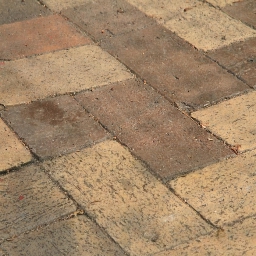}  \\

\multicolumn{6}{c}{(c) HomoFormer} \\

\includegraphics[width=\imgwidthfigtwo]{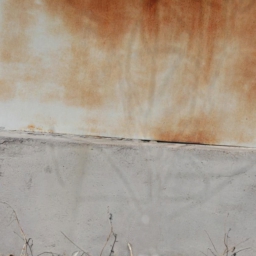} &
\includegraphics[width=\imgwidthfigtwo]{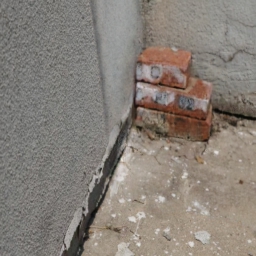} &
\includegraphics[width=\imgwidthfigtwo]{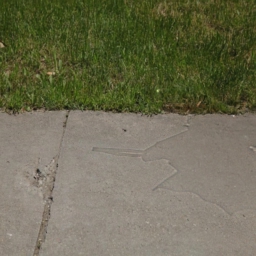} &
\includegraphics[width=\imgwidthfigtwo]{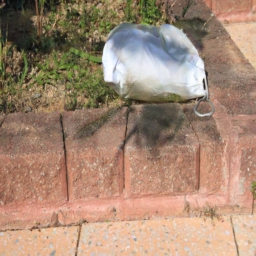} &
\includegraphics[width=\imgwidthfigtwo]{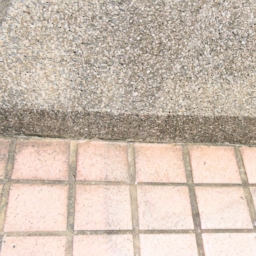} &
\includegraphics[width=\imgwidthfigtwo]{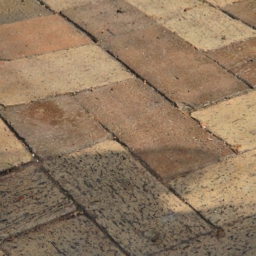} \\

\multicolumn{6}{c}{(d) DC-ShadowNet} \\

\includegraphics[width=\imgwidthfigtwo]{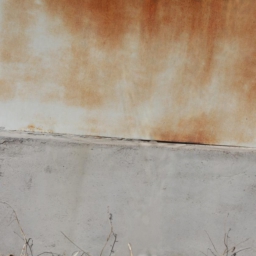} &
\includegraphics[width=\imgwidthfigtwo]{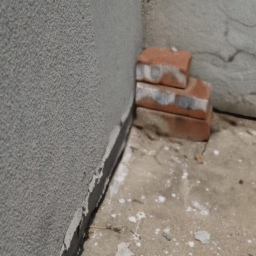} &
\includegraphics[width=\imgwidthfigtwo]{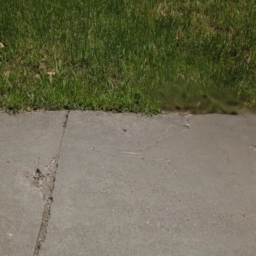} &
\includegraphics[width=\imgwidthfigtwo]{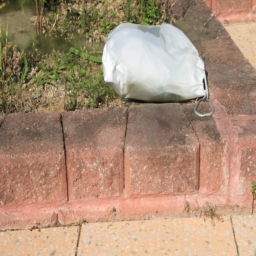} &
\includegraphics[width=\imgwidthfigtwo]{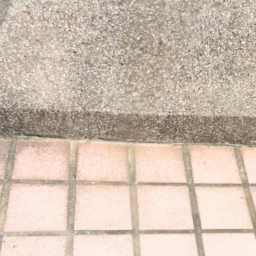} &
\includegraphics[width=\imgwidthfigtwo]{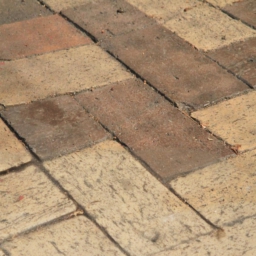} \\

\multicolumn{6}{c}{(e) Ours output images} \\

\includegraphics[width=\imgwidthfigtwo]{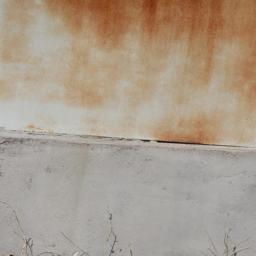} &
\includegraphics[width=\imgwidthfigtwo]{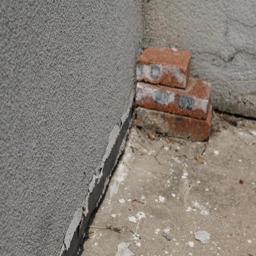} &
\includegraphics[width=\imgwidthfigtwo]{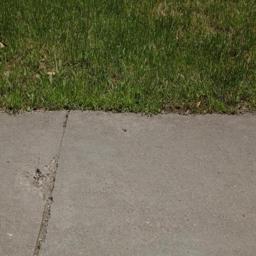} &
\includegraphics[width=\imgwidthfigtwo]{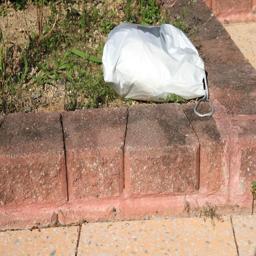} &
\includegraphics[width=\imgwidthfigtwo]{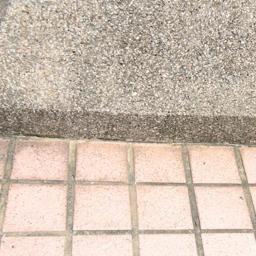} &
\includegraphics[width=\imgwidthfigtwo]{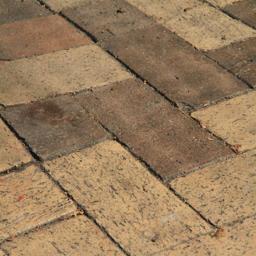} \\

\multicolumn{6}{c}{(f) Ground truth} \\

\end{tabular}

\caption{Qualitative comparison on the SRD dataset~\cite{qu2017deshadownet}.}
\label{fig:fig2}
\end{figure*}

%% file: figures/supp/fig3.tex
\begin{figure*}[ht]
\centering
\newcommand{\ncols}{6}
\setlength{\tabcolsep}{0.1em}

\newlength{\imgwidthfigthree}
\setlength{\imgwidthfigthree}{\dimexpr(\textwidth - \ncols\tabcolsep)/\ncols\relax}

\begin{tabular}{*{\ncols}{c}}
\centering

\includegraphics[width=\imgwidthfigthree]{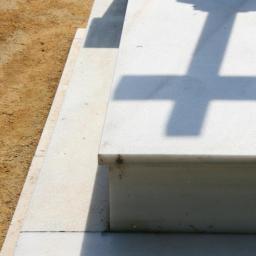} &
\includegraphics[width=\imgwidthfigthree]{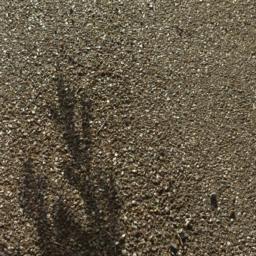} &
\includegraphics[width=\imgwidthfigthree]{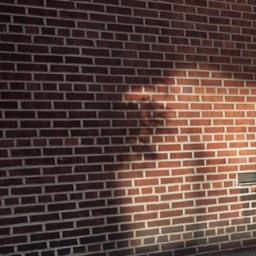} &
\includegraphics[width=\imgwidthfigthree]{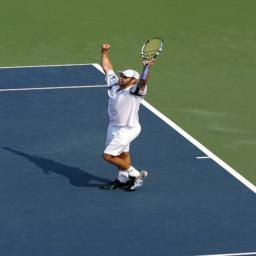} &
\includegraphics[width=\imgwidthfigthree]{figures/images/SBU/input/219.jpg} &
\includegraphics[width=\imgwidthfigthree]{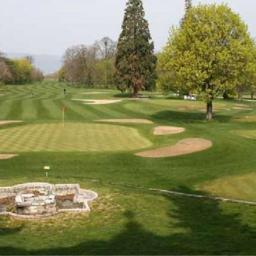}  \\

\multicolumn{6}{c}{(a) Input shadow images} \\

\includegraphics[width=\imgwidthfigthree]{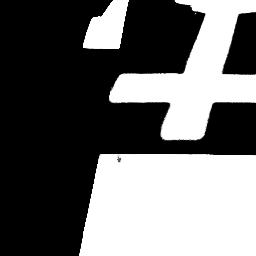} &
\includegraphics[width=\imgwidthfigthree]{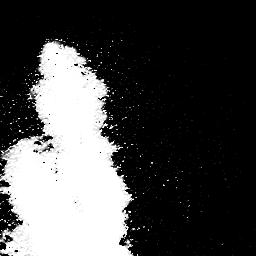} &
\includegraphics[width=\imgwidthfigthree]{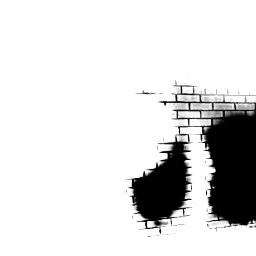} &
\includegraphics[width=\imgwidthfigthree]{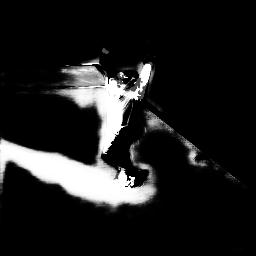} &
\includegraphics[width=\imgwidthfigthree]{figures/images/SBU/Ours_mask/219.jpg} &
\includegraphics[width=\imgwidthfigthree]{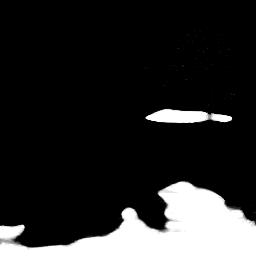}  \\

\multicolumn{6}{c}{(b) Ours output mattes} \\

\includegraphics[width=\imgwidthfigthree]{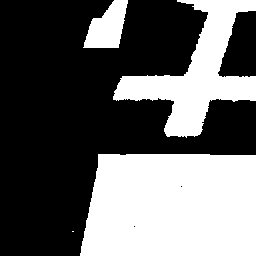} &
\includegraphics[width=\imgwidthfigthree]{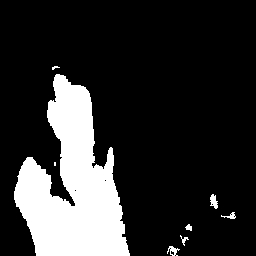} &
\includegraphics[width=\imgwidthfigthree]{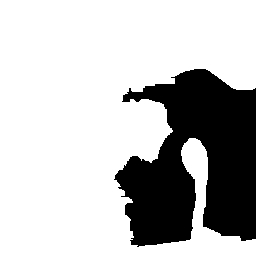} &
\includegraphics[width=\imgwidthfigthree]{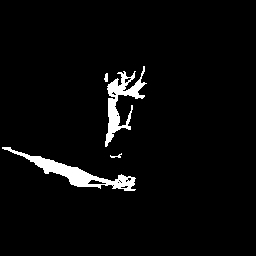} &
\includegraphics[width=\imgwidthfigthree]{figures/images/SBU/GT/219.jpg} &
\includegraphics[width=\imgwidthfigthree]{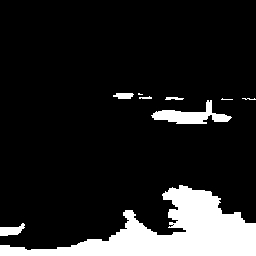}  \\

\multicolumn{6}{c}{(c) Ground truth masks} \\

\includegraphics[width=\imgwidthfigthree]{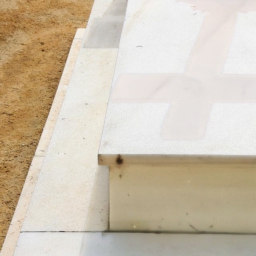} &
\includegraphics[width=\imgwidthfigthree]{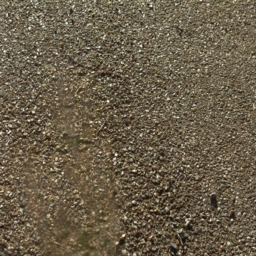} &
\includegraphics[width=\imgwidthfigthree]{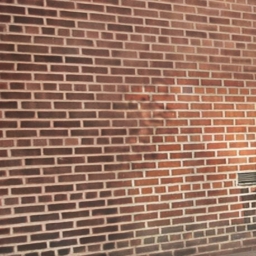} &
\includegraphics[width=\imgwidthfigthree]{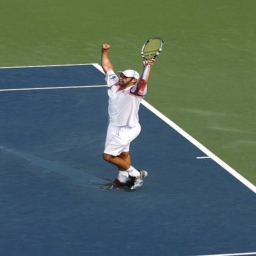} &
\includegraphics[width=\imgwidthfigthree]{figures/images/SBU/Ours/219.jpg} &
\includegraphics[width=\imgwidthfigthree]{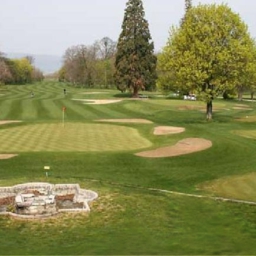}  \\

\multicolumn{6}{c}{(d) Ours output images} \\

\end{tabular}

\caption{Qualitative results of shadow segmentation on the SBU dataset \cite{vicente2016large}.}
\label{fig:fig3}
\end{figure*}

%% file: figures/supp/fig4.tex
\begin{figure*}[ht]
\centering
\newcommand{\ncols}{3}
\setlength{\tabcolsep}{0.1em}

\newlength{\imgwidthvideotwo}
\setlength{\imgwidthvideotwo}{\dimexpr(\textwidth - \ncols\tabcolsep)/\ncols\relax}

\begin{tabular}{*{\ncols}{c}}
\includegraphics[width=\imgwidthvideotwo]{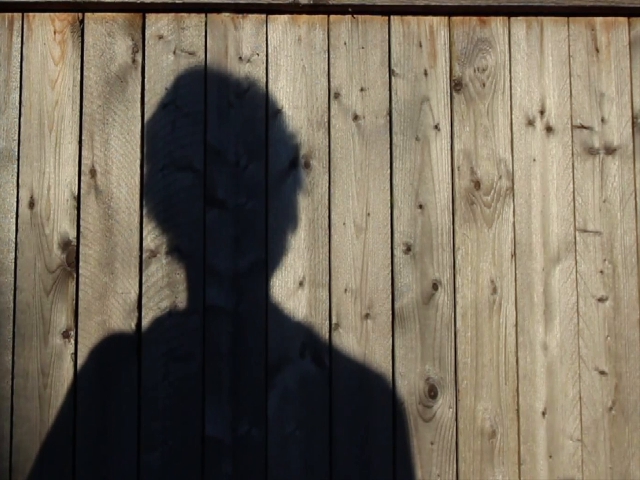} &
\includegraphics[width=\imgwidthvideotwo]{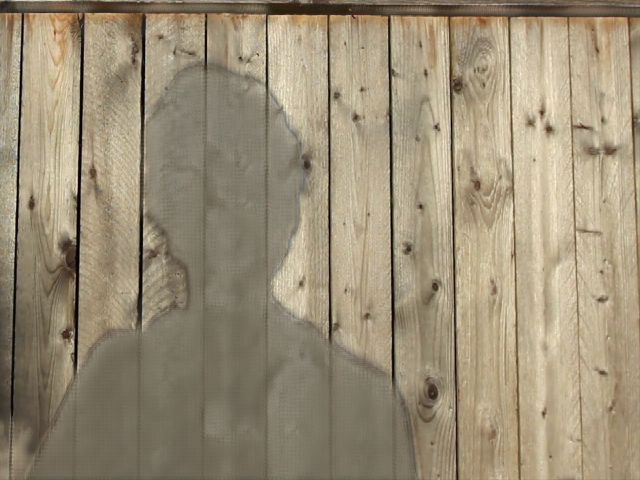} &
\includegraphics[width=\imgwidthvideotwo]{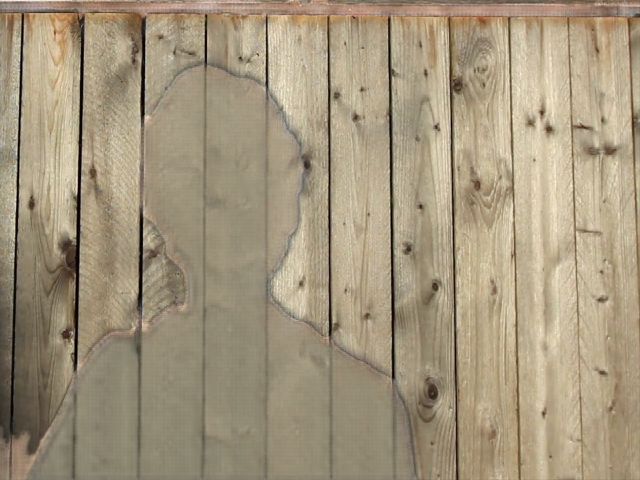} \\

(a) Input &
(b) LG-ShadowNet &
(c) G2R \\

\includegraphics[width=\imgwidthvideotwo]{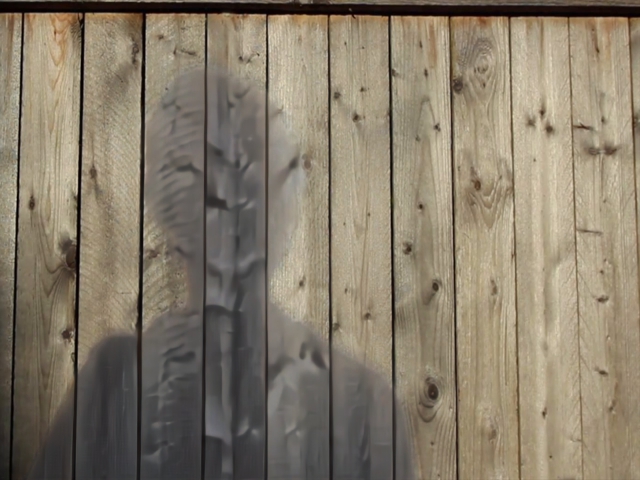} &
\includegraphics[width=\imgwidthvideotwo]{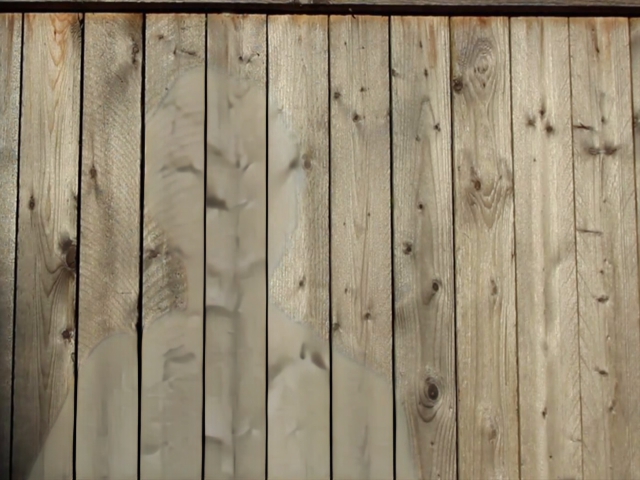} &
\includegraphics[width=\imgwidthvideotwo]{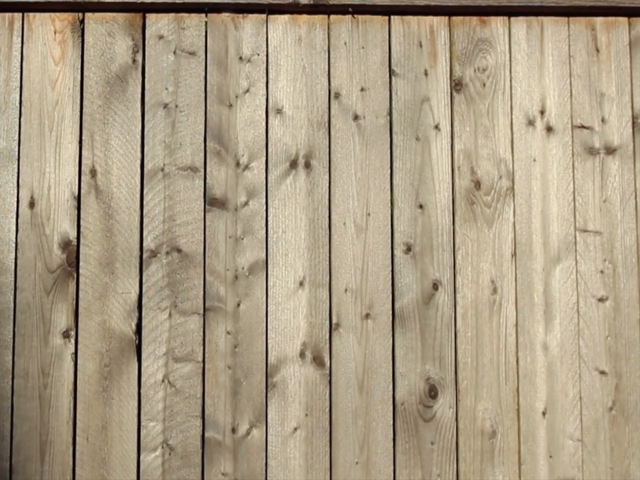} \\

(d) Ours &
(e) Ours+ &
(f) GT \\
\end{tabular}

\caption{Qualitative comparison on the video shadow dataset~\cite{le2020shadow}.}
\label{fig:video2}
\end{figure*}

%% file: figures/supp/fig5.tex
\begin{figure*}[ht]
\centering
\newcommand{\ncols}{6}
\setlength{\tabcolsep}{0.1em}

\newlength{\imgwidthvideothree}
\setlength{\imgwidthvideothree}{\dimexpr(\textwidth - \ncols\tabcolsep)/\ncols\relax}

\begin{tabular}{*{\ncols}{c}}
\includegraphics[width=\imgwidthvideothree]{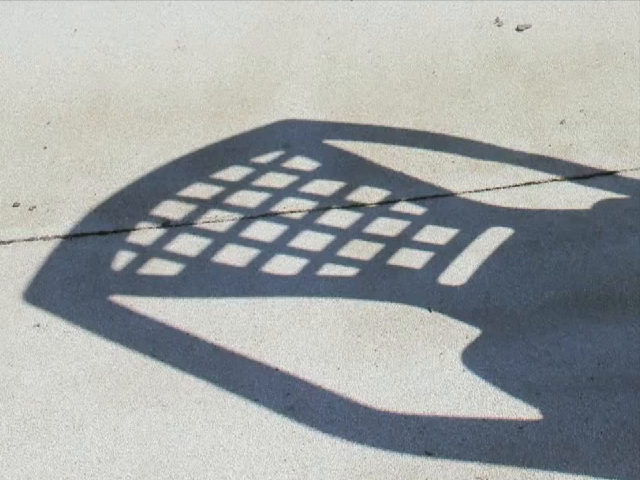} &
\includegraphics[width=\imgwidthvideothree]{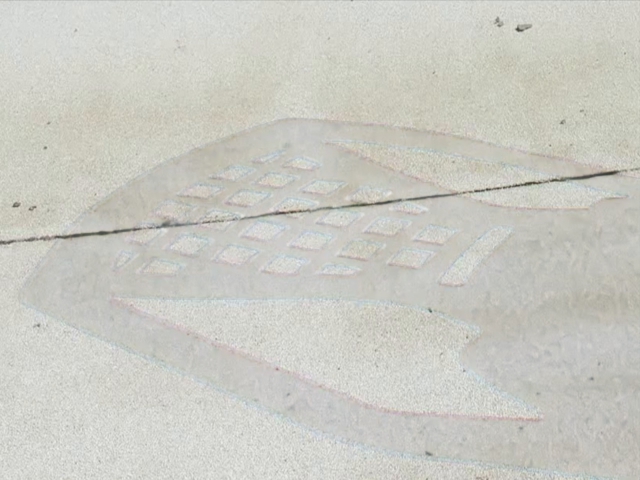} &
\includegraphics[width=\imgwidthvideothree]{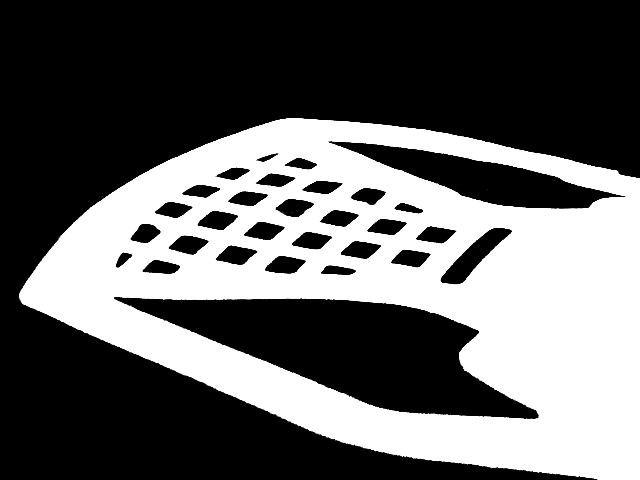} &
\includegraphics[width=\imgwidthvideothree]{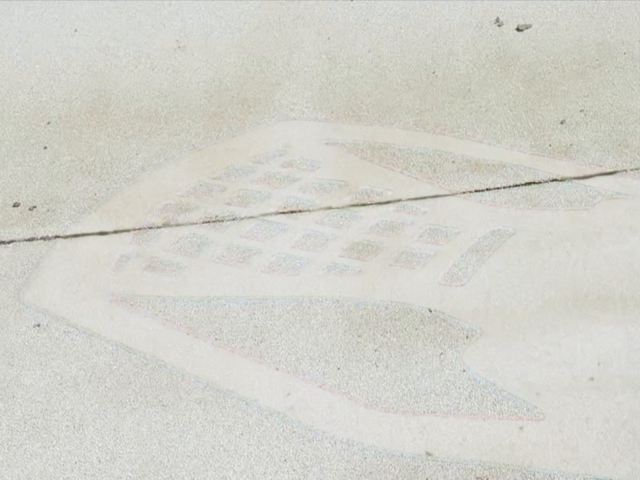} &
\includegraphics[width=\imgwidthvideothree]{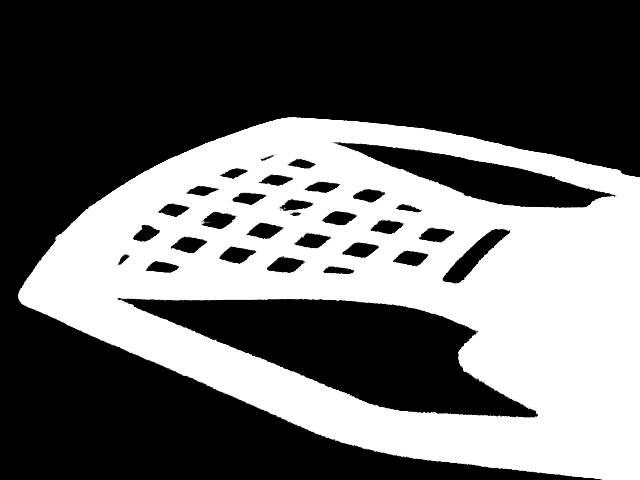} &
\includegraphics[width=\imgwidthvideothree]{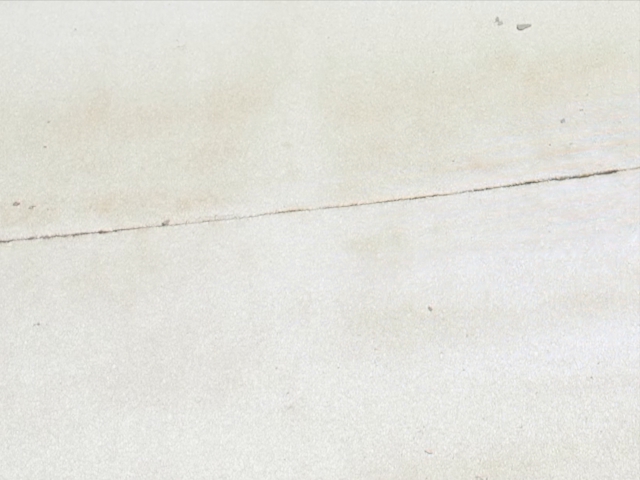} \\

\includegraphics[width=\imgwidthvideothree]{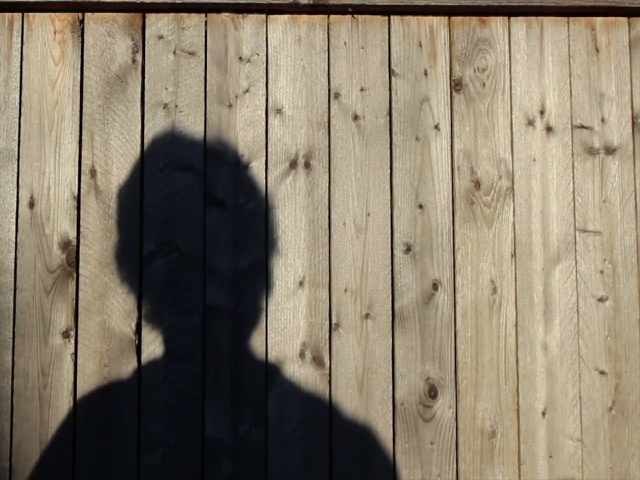} &
\includegraphics[width=\imgwidthvideothree]{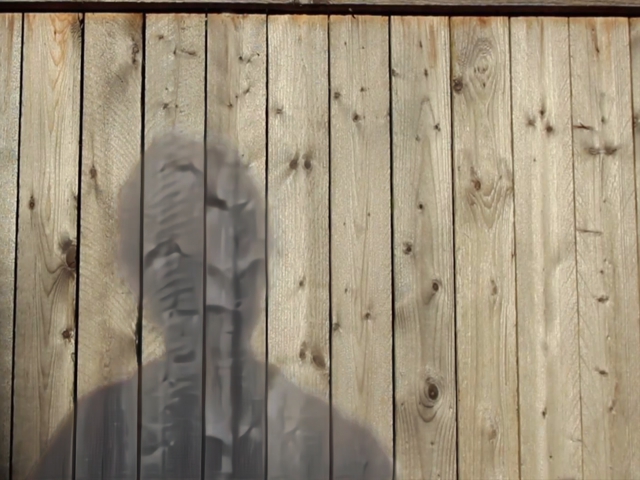} &
\includegraphics[width=\imgwidthvideothree]{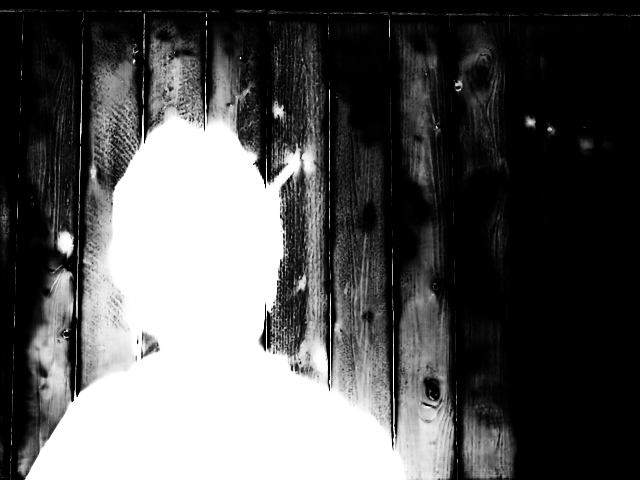} &
\includegraphics[width=\imgwidthvideothree]{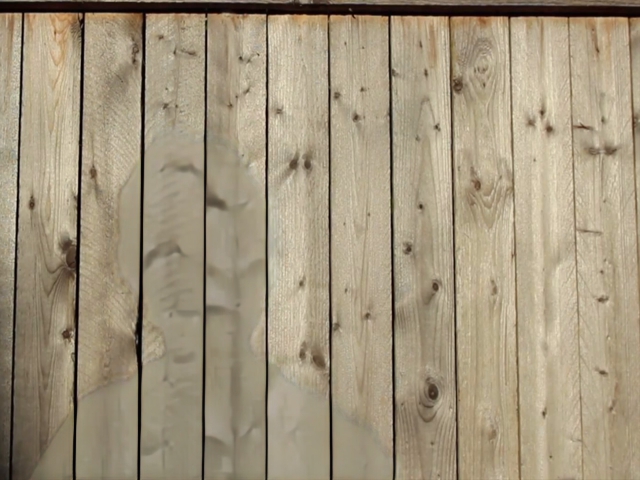} &
\includegraphics[width=\imgwidthvideothree]{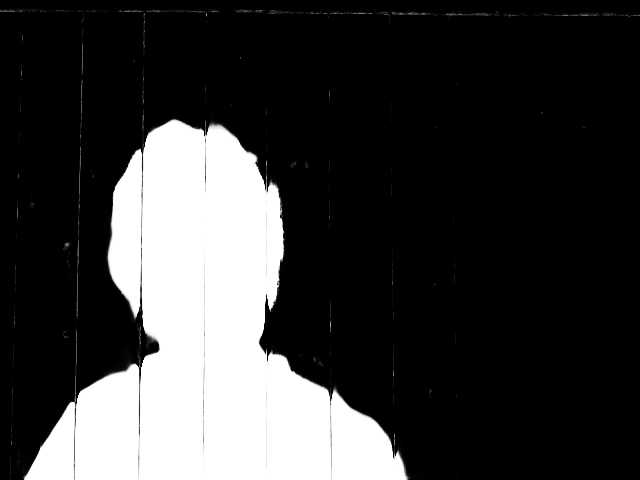} &
\includegraphics[width=\imgwidthvideothree]{figures/images/video/test_C/guy_max.jpg} \\

\includegraphics[width=\imgwidthvideothree]{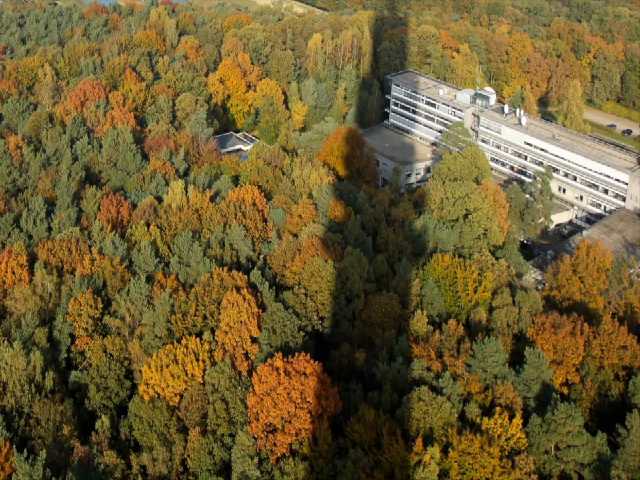} &
\includegraphics[width=\imgwidthvideothree]{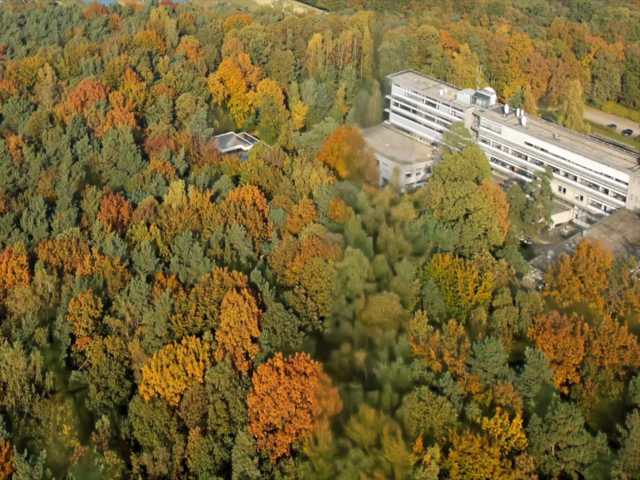} &
\includegraphics[width=\imgwidthvideothree]{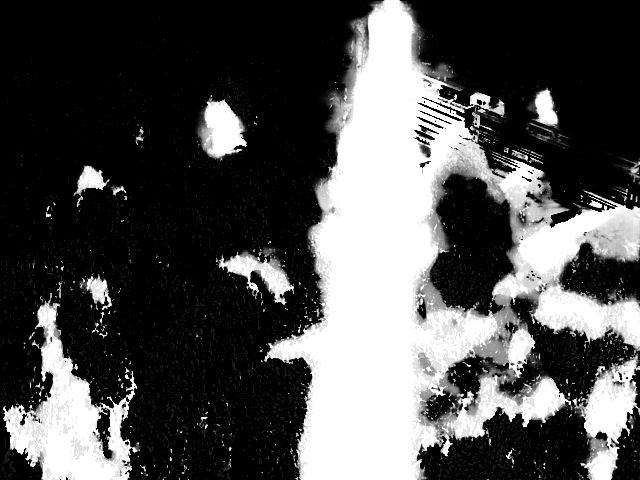} &
\includegraphics[width=\imgwidthvideothree]{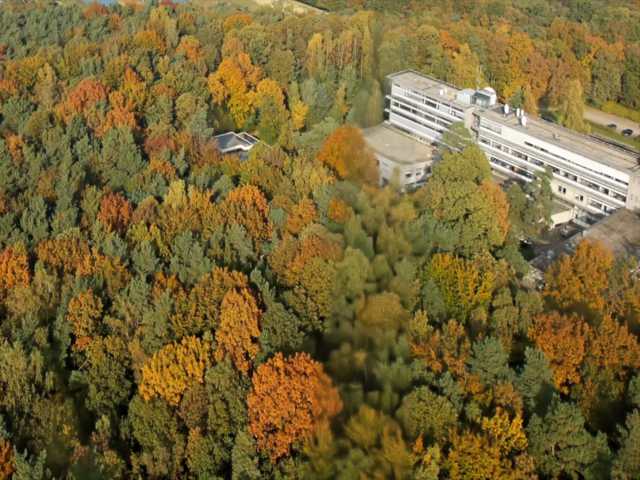} &
\includegraphics[width=\imgwidthvideothree]{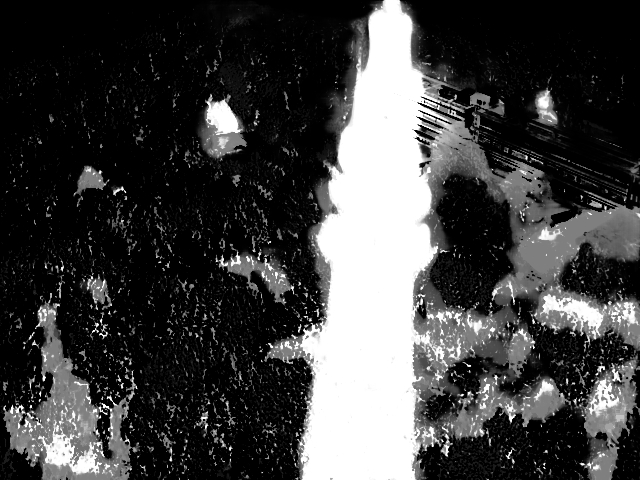} &
\includegraphics[width=\imgwidthvideothree]{figures/images/video/test_C/tower_max.jpg} \\

(a) Input &
(b) Ours &
(c) Ours mask &
(d) Ours+ &
(e) Ours+ mask &
(f) GT \\
\end{tabular}

\caption{Qualitative results on the video shadow dataset~\cite{le2020shadow}.}
\label{fig:video3}
\end{figure*}

%% file: figures/supp/fig6.tex
\begin{figure*}[ht]
\centering
\newcommand{\ncols}{6}
\setlength{\tabcolsep}{0.1em}

\newlength{\imgwidthfigfour}
\setlength{\imgwidthfigfour}{\dimexpr(\textwidth - \ncols\tabcolsep)/\ncols\relax}

\begin{tabular}{*{\ncols}{c}}
\centering

\includegraphics[width=\imgwidthfigfour]{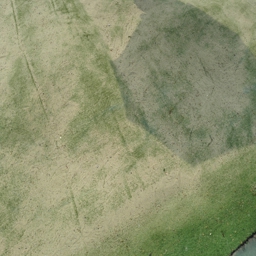} &
\includegraphics[width=\imgwidthfigfour]{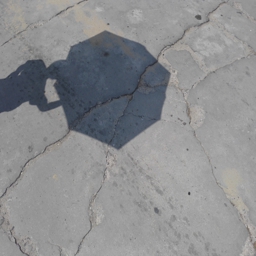} &
\includegraphics[width=\imgwidthfigfour]{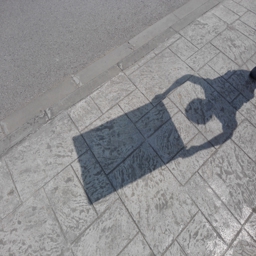} &
\includegraphics[width=\imgwidthfigfour]{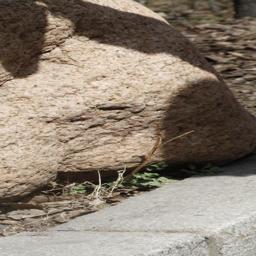} &
\includegraphics[width=\imgwidthfigfour]{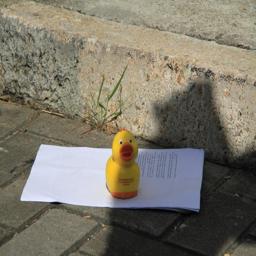} &
\includegraphics[width=\imgwidthfigfour]{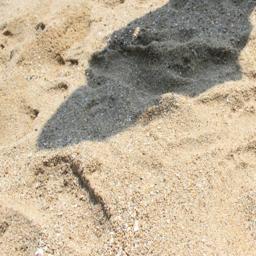} \\

\multicolumn{6}{c}{(a) Input shadow images} \\

\includegraphics[width=\imgwidthfigfour]{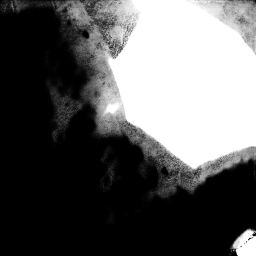} &
\includegraphics[width=\imgwidthfigfour]{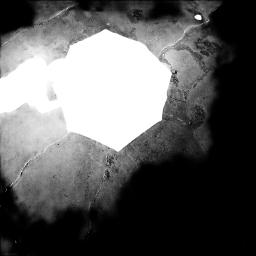} &
\includegraphics[width=\imgwidthfigfour]{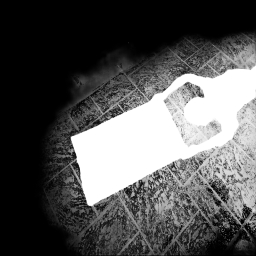} &
\includegraphics[width=\imgwidthfigfour]{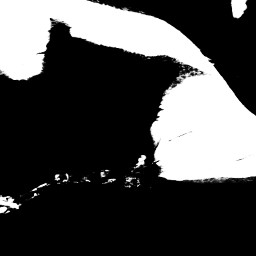} &
\includegraphics[width=\imgwidthfigfour]{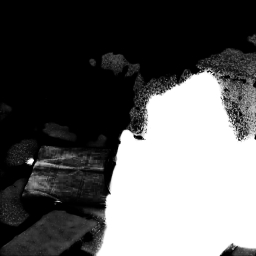} &
\includegraphics[width=\imgwidthfigfour]{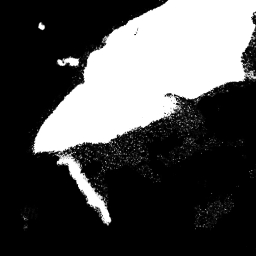} \\

\multicolumn{6}{c}{(b) Ours output mattes} \\

\includegraphics[width=\imgwidthfigfour]{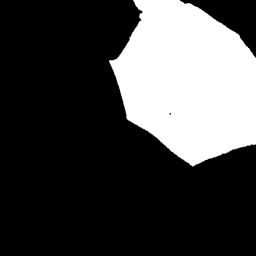} &
\includegraphics[width=\imgwidthfigfour]{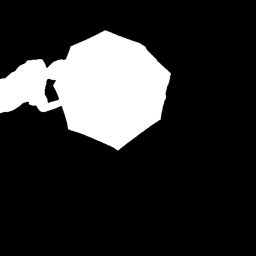} &
\includegraphics[width=\imgwidthfigfour]{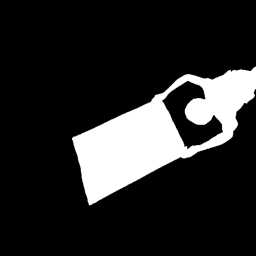} &
\includegraphics[width=\imgwidthfigfour]{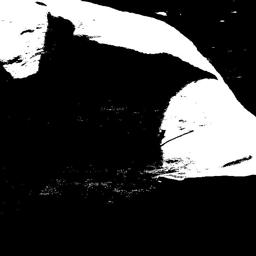} &
\includegraphics[width=\imgwidthfigfour]{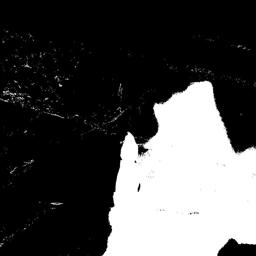} &
\includegraphics[width=\imgwidthfigfour]{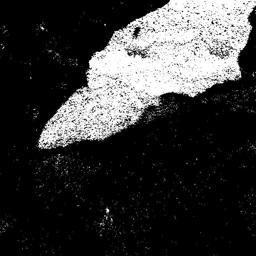} \\

\multicolumn{6}{c}{(c) Ground truth masks} \\

\includegraphics[width=\imgwidthfigfour]{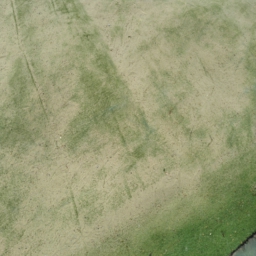} &
\includegraphics[width=\imgwidthfigfour]{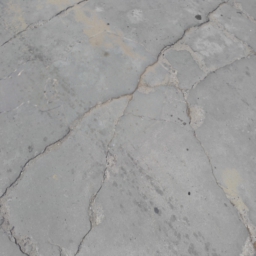} &
\includegraphics[width=\imgwidthfigfour]{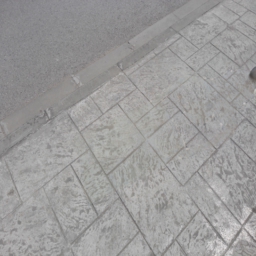} &
\includegraphics[width=\imgwidthfigfour]{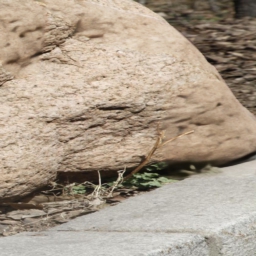} &
\includegraphics[width=\imgwidthfigfour]{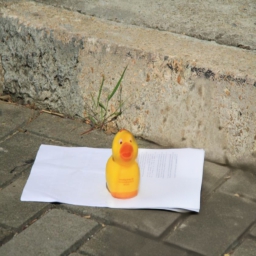} &
\includegraphics[width=\imgwidthfigfour]{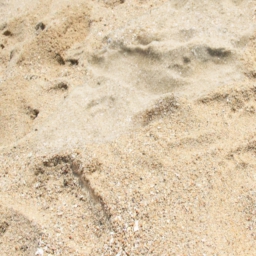} \\

\multicolumn{6}{c}{(d) Ours output images} \\

\includegraphics[width=\imgwidthfigfour]{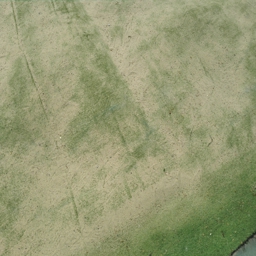} &
\includegraphics[width=\imgwidthfigfour]{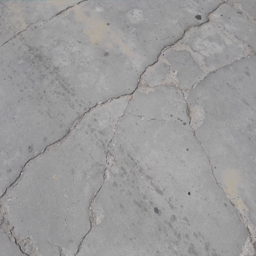} &
\includegraphics[width=\imgwidthfigfour]{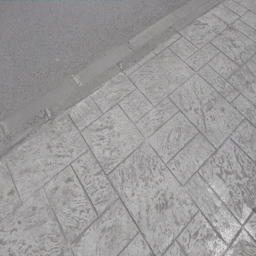} &
\includegraphics[width=\imgwidthfigfour]{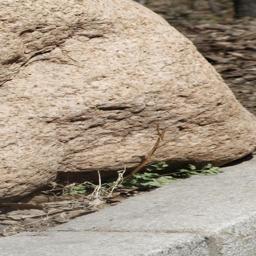} &
\includegraphics[width=\imgwidthfigfour]{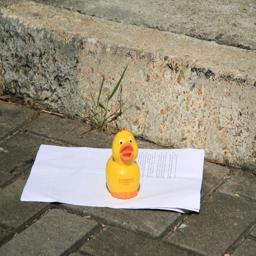} &
\includegraphics[width=\imgwidthfigfour]{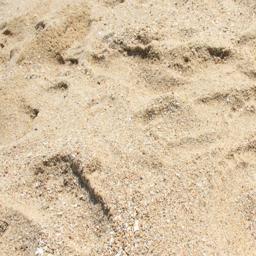} \\

\multicolumn{6}{c}{(e) Ground truth images} \\

\end{tabular}

\caption{Qualitative results of shadow segmentation on the AISTD~\cite{le2019shadow, wang2018stacked} and SRD~\cite{qu2017deshadownet} datasets.}
\label{fig:fig6}
\end{figure*}